\documentclass[11pt]{article}

\usepackage[final]{acl}

\usepackage[table]{xcolor} 
\usepackage{tabularx}
\newcolumntype{R}{>{\raggedleft\arraybackslash}X}

\definecolor{mybrown}{HTML}{EA580C}
\definecolor{mygreen}{HTML}{008744}
\definecolor{myorange}{HTML}{ffa700}
\definecolor{darkorange}{HTML}{E9570C}
\definecolor{myred}{HTML}{d62d20}
\definecolor{myblue}{HTML}{0668E1}
\definecolor{darkblue}{rgb}{0, 0, 0.5}

\usepackage{booktabs}
\usepackage{multirow}

\usepackage{amsmath}
\usepackage{amssymb}
\usepackage{mathtools}

\newcommand{\mc}[1]{\mathcal{#1}}
\newcommand{\tsc}[1]{\textsc{#1}}  
\newcommand{\ttt}[1]{\texttt{#1}}  

\usepackage{graphicx}
\usepackage{subcaption}
\usepackage[hypcap=false]{caption}

\usepackage{stfloats}
\usepackage{xurl}

\usepackage[ruled,vlined,algo2e]{algorithm2e}
\SetKwInOut{Input}{Input}
\SetKwInOut{Output}{Output}

\usepackage{microtype}
\usepackage{fontawesome5}
\usepackage{enumitem}

\usepackage{hyperref}

\definecolor{darkblue}{rgb}{0, 0, 0.5}
\hypersetup{
    colorlinks = true,
    citecolor  = darkblue,      
    linkcolor  = [HTML]{8B0000}, 
    filecolor  = [HTML]{00008B},
    urlcolor   = [HTML]{00008B}
}

\usepackage{times}

\usepackage[T1]{fontenc}

\usepackage[utf8]{inputenc}

\usepackage{inconsolata}

\title{\tsc{mSFT}: Addressing Dataset Mixtures Overfitting\\Heterogeneously in Multi-task SFT}

\author{
  \textbf{Woosung Koh\textsuperscript{\scalebox{0.7}{$\spadesuit$}$\star$}},
  \textbf{Jeyoung Jeon\textsuperscript{\scalebox{0.7}{$\diamondsuit$}$\star$}},
  \textbf{Youngjin Song\textsuperscript{\scalebox{0.7}{$\diamondsuit$}}},
  \textbf{Yujin Cheon},
\\
  \textbf{Soowon Oh\textsuperscript{\scalebox{0.7}{$\spadesuit\heartsuit$}}},
  \textbf{Jaehyeong Choi\textsuperscript{\scalebox{0.7}{$\diamondsuit$}}},
  \textbf{Se-Young Yun\textsuperscript{\scalebox{0.7}{$\spadesuit$}$\dagger$}}
\\
\\
  \textsuperscript{\scalebox{0.7}{$\spadesuit$}}KAIST AI \quad
  \textsuperscript{\scalebox{0.7}{$\diamondsuit$}}Yonsei University \quad
  \textsuperscript{\scalebox{0.7}{$\heartsuit$}}Samsung Advanced Institute of Technology
\\
  \textsuperscript{$\star$}Equal contribution \quad
  \textsuperscript{$\dagger$}Corresponding author
\\
  \texttt{\{reiss.koh, yunseyoung\}@kaist.ac.kr}
}

\begin{document}

\maketitle

\begin{abstract}
Current language model training commonly applies multi-task Supervised Fine-Tuning (SFT) using a homogeneous compute budget across all sub-datasets. This approach is fundamentally sub-optimal: heterogeneous learning dynamics cause faster-learning tasks to overfit early while slower ones remain under-fitted. To address this, we introduce \tsc{mSFT}, an iterative, overfitting-aware search algorithm for multi-task data mixtures. \tsc{mSFT} trains the model on an active mixture, identifies and excludes the earliest overfitting sub-dataset, and reverts to that specific optimal checkpoint before continuing. Extensive evaluations demonstrate that \tsc{mSFT} \textit{consistently} outperforms \textbf{4} baselines across \textbf{10} benchmarks and \textbf{6} base models. Further analysis confirms \tsc{mSFT}  maintains robust gains across diverse dataset sizes, task granularities, and is insensitive to its single new hyperparameter (compute budget). Notably, at low compute budget, \tsc{mSFT} can improve performance \textit{while} lowering training FLOPs. Ultimately, \tsc{mSFT} establishes a \textit{practical} overfitting-aware algorithm for multi-task SFT that maximizes the potential of models across diverse data mixtures.

\end{abstract}

\begin{center}
\href{https://github.com/reiss-koh/msft}{\faGithub \textbf{ Code}}
\end{center}

\section{Introduction} \label{sec:intro}

Since the introduction of transformers \citep{NIPS2017_3f5ee243} and scaling laws \citep{kaplan2020scalinglawsneurallanguage}, \textit{general} foundation models trained on \textit{diverse} data have overtaken specialized models \citep{maslej2025artificial}. These foundation models undertake a multi-task Supervised Fine-tuning (SFT) stage, where diverse sub-datasets are commonly randomly mixed together \citep{adler2024nemotron, hui2024qwen2, grattafiori2024llama}; primarily to avoid forgetting from sequential training \citep{wang2025nemotron, 11151751}. Within this paradigm, practitioners follow a well-known approach, identifying the pre-overfitting optimal training compute (epoch) given a fixed data size \citep{NIPS1991_ff4d5fbb}. This optimal compute level is determined empirically by allocating a large amount of compute while saving intermediate checkpoints in memory, then identifying the checkpoint with the best generalization benchmark scores \citep{PRECHELT1998761,JMLR:v23:21-0983}.

Within this framework, frontier open-weight models inherently assume that the \textit{global} optimal compute budget aligns with the optimal compute of each underlying sub-dataset. Consider Tab. \ref{tab:homo}, where \ttt{Magistral} \citep{rastogi2025magistral}, \ttt{OLMo} \citep{groeneveld-etal-2024-olmo,walsh2025,olmo2025olmo}, \ttt{DeepSeek} \citep{liu2024deepseek,guo2025deepseek}, and \ttt{Qwen} \citep{qwen2025qwen25technicalreport} family of models identify the final compute-level homogeneously (i.e., same compute for all sub-datasets). 

\begin{table}[t]
    \centering
    \setlength{\tabcolsep}{0.5pt}
    \small 
    \begin{tabular}{@{}l l l@{}}
        \toprule
        \textbf{Method} & \textbf{Type} & \textbf{Epochs} \\
        \midrule
        \ttt{Magistral} {\small \citep{rastogi2025magistral}} & \textcolor{myred}{Homogenous} & 4 \\
        \ttt{OLMo} {\small \citep{groeneveld-etal-2024-olmo}}& \textcolor{myred}{Homogenous} & 3 \\
        \ttt{OLMo 2} {\small \citep{walsh2025}}& \textcolor{myred}{Homogenous} & 2 \\
        \ttt{OLMo 3} {\small \citep{olmo2025olmo}}& \textcolor{myred}{Homogenous} & 2 \\
        \ttt{DeepSeek-V3} {\small \citep{liu2024deepseek}}& \textcolor{myred}{Homogenous} & 2 \\
        \ttt{DeepSeek-R1} {\small \citep{guo2025deepseek}}& \textcolor{myred}{Homogenous} & 2-3 \\
        \ttt{Qwen2.5} {\small \citep{qwen2025qwen25technicalreport}}& \textcolor{myred}{Homogenous} & 2 \\

        \multirow{2}{*}{\ttt{Nemotron-4} {\small \citep{adler2024nemotron}}} & \textcolor{darkorange}{Multi-stage} & 1 (Code) + \\ 
         & \textcolor{darkorange}{Homogenous} & 3 (General) \\
        \rowcolor{gray!15} \tsc{mSFT} (\textbf{ours}) & \textcolor{mygreen}{Heterogeneous} & Dynamic \\
        \bottomrule
    \end{tabular}
    \caption{\textbf{Status quo.} Frontier open-weight models continue to employ homogeneous SFT, where all sub-datasets are trained on the \textit{same} amount of compute.}
    \label{tab:homo}
\end{table}
    
We hypothesize that this \textit{de facto} approach is sub-optimal as each sub-dataset embody distinct distributions that lead to different learning and generalization dynamics. Nemotron \citep{adler2024nemotron} demonstrated that their code sub-dataset required less compute than every other sub-dataset. Nevertheless, their compute allocation remains coarse, which we term as "Multi-stage Homogenous" in Tab. \ref{tab:homo}.

\begin{figure}[htbp]
    \centering
    \includegraphics[width=\linewidth]{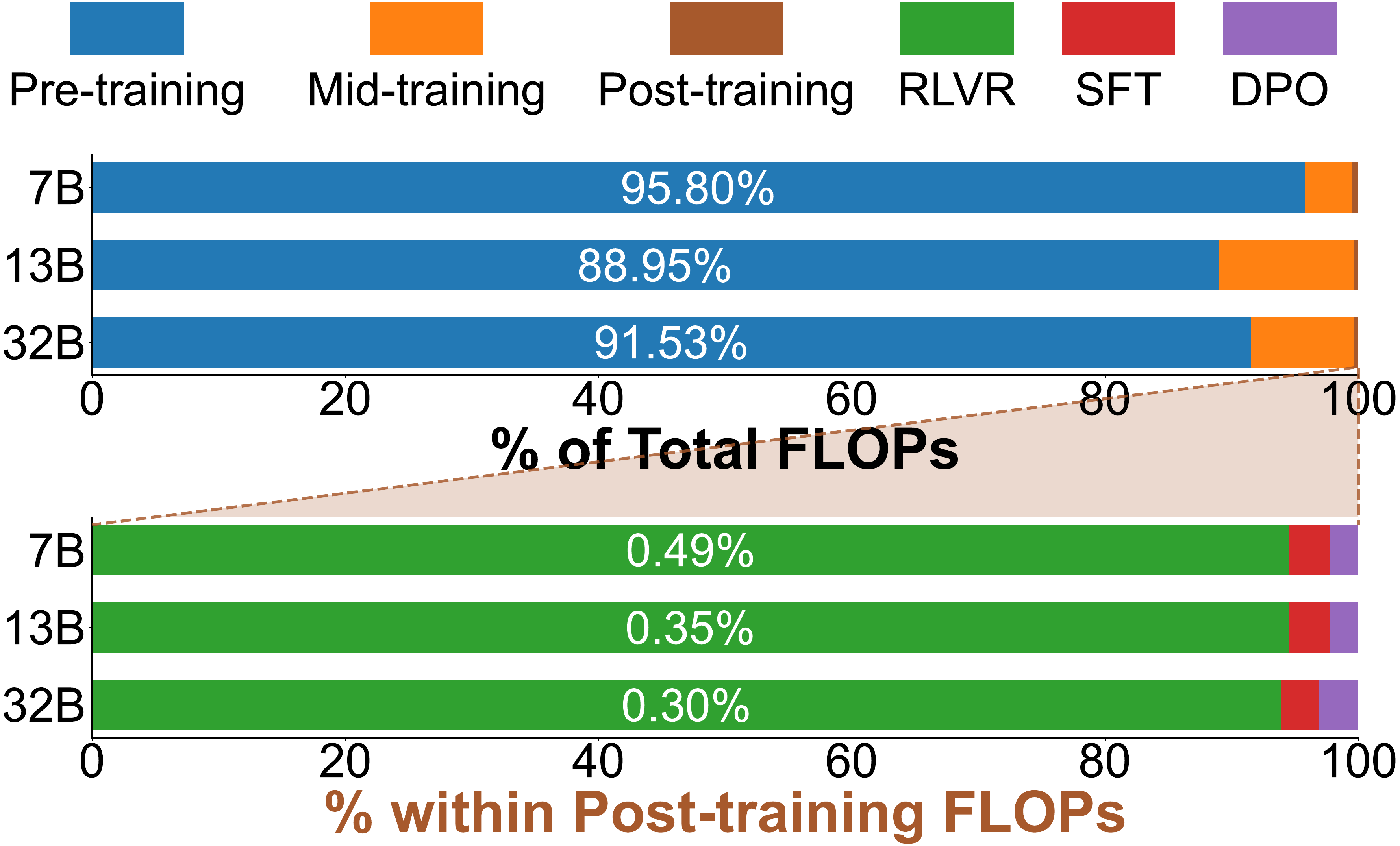}
    \caption{\textbf{SFT is compute-light.} Using \ttt{OLMo 2} as an example, SFT is relatively compute-light, and therefore additional compute usage at this stage is negligible.}
    \label{fig:train_proportion}
\end{figure}

Although empirically searching for the optimal compute per sub-dataset incurs additional costs, we argue these increases are negligible since SFT is one of the computationally lightest training stage. Consider Fig. \ref{fig:train_proportion}, where we visualize the proportion of training compute allocated to the SFT stage considering the end-to-end training pipeline. We detail how this was derived based on open-source information in Appendix \ref{app:train_proportion}. We observe that the SFT stage takes approximately 0.01\% of total training compute. Moreover, consistent performance gains with additional compute usage has been an influential philosophy guiding modern training \citep{chen-etal-2025-revisiting, tan2025scalingbehaviorsllmreinforcement,koh2026generativevisualcodemobile}.

\paragraph{Contribution.} Given this backdrop, we first empirically demonstrate that dataset mixtures composed of sub-datasets overfit heterogenously, confirming our hypothesis that the \textit{status quo} is sub-optimal (\textbf{\S\ \ref{sec:motivation}}, Fig. \ref{fig:motivation}). In response, we propose \tsc{mSFT} (m representing \underline{m}ulti-task \underline{m}ixture), an overfitting search algorithm for multi-task SFT (\textbf{\S\ \ref{sec:msft}}). Prior to introducing our approach, we discuss the limitations of a naïve approach (\textbf{\S\ \ref{sec:naive}}). Then, we introduce our search method which dynamically excludes sub-datasets by iteratively rolling back to the checkpoint where a sub-dataset over-fitted the quickest (\textbf{\S\ \ref{sec:ours}}, Alg. \ref{alg:msft})\footnote{Code is released with Apache-2.0 License}. Finally, we empirically demonstrate that \tsc{mSFT} is useful for practitioners, including extensive further analyses (\textbf{\S\ \ref{sec:exp}}):

\begin{itemize}
    \item \tsc{mSFT}'s average performance across \textbf{10} benchmarks outperform \textbf{4} baselines (and \textbf{2} ablative baselines) across \textbf{6} base models (\textbf{\S\ \ref{sec:exp_main_result}}, Tab. \ref{tab:main}, \ref{tab:ablation}).

    \begin{itemize}
        \item We observe that performance gains are not from disproportionate gains on a few outlier tasks, as seen by a decrease in standard deviation across benchmarks (Fig. \ref{fig:further}).
    \end{itemize}

    \item \tsc{mSFT} performance gains are robust across diverse dataset sizes (9K, 18K, 27K) and task counts (5, 10, 15) (\textbf{\S\ \ref{sec:further}}, Fig. \ref{fig:diverse_N}).

    \item Reducing \tsc{mSFT}'s only hyperparameter, compute budget $C$ does not lead to performance degradation; with low $C$ enabling FLOPs savings against SFT \textit{while} improving performance (\textbf{\S\ \ref{sec:further}}, Fig. \ref{fig:granular_flops}).

    \item We demonstrate that \tsc{mSFT} works on diverse levels of task granularity by experimenting \tsc{mSFT} on a single dataset with sub-categories (\textbf{\S\ \ref{sec:further}}, Fig. \ref{fig:med}).

    \item We decompose the performance difference of SFT and \tsc{mSFT} through the lense of overfitting avoidance and catastrophic forgetting; and also show that \tsc{mSFT} commonly achieves a lower train loss (\textbf{\S\ \ref{sec:further}}, Fig. \ref{fig:decomp}).
\end{itemize}

\section{Motivation: Dataset Mixtures Overfit Heterogeneously} \label{sec:motivation}

Multi-task SFT suffers from a fundamental misalignment between the diverse learning dynamics of individual tasks and the rigid nature of standard training paradigms. To formalize this, consider SFT of Language Models (LMs) parameterized by $\theta$ on a multi-task dataset mixture $\mc{D} = \bigcup_{i=1}^N \mc{D}_i$, which consists of $N$ distinct tasks. We measure training progress using a continuous compute variable $c$, generalizing training epochs into finer-grained units (e.g., fractional epochs). For any given task $i$, there exists an optimal compute $c^*_i$, defined as the stopping point where the model achieves maximum generalization on the task's held-out validation set:
\begin{equation}
    c^*_i = \operatorname*{arg max}_{c} \text{Metric}(\theta_{c}; \mc{D}_i^{\text{val}})
\end{equation}

Under the standard \textit{homogeneous} training paradigm, this inherent diversity in optimal stopping points is ignored. The model is trained on the dataset mixture $\mc{D}$ for a fixed global compute budget $c_{\text{global}}$. This imposes a rigid constraint where every task $i$ is forced to adhere to the exact same training compute, meaning $c_i := c_{\text{global}}, \forall i \in \{1, \dots, N\}$.

\begin{figure*}[tbp]
    \centering
    \begin{subfigure}{0.495\linewidth}
        \centering
        \includegraphics[width=\linewidth]{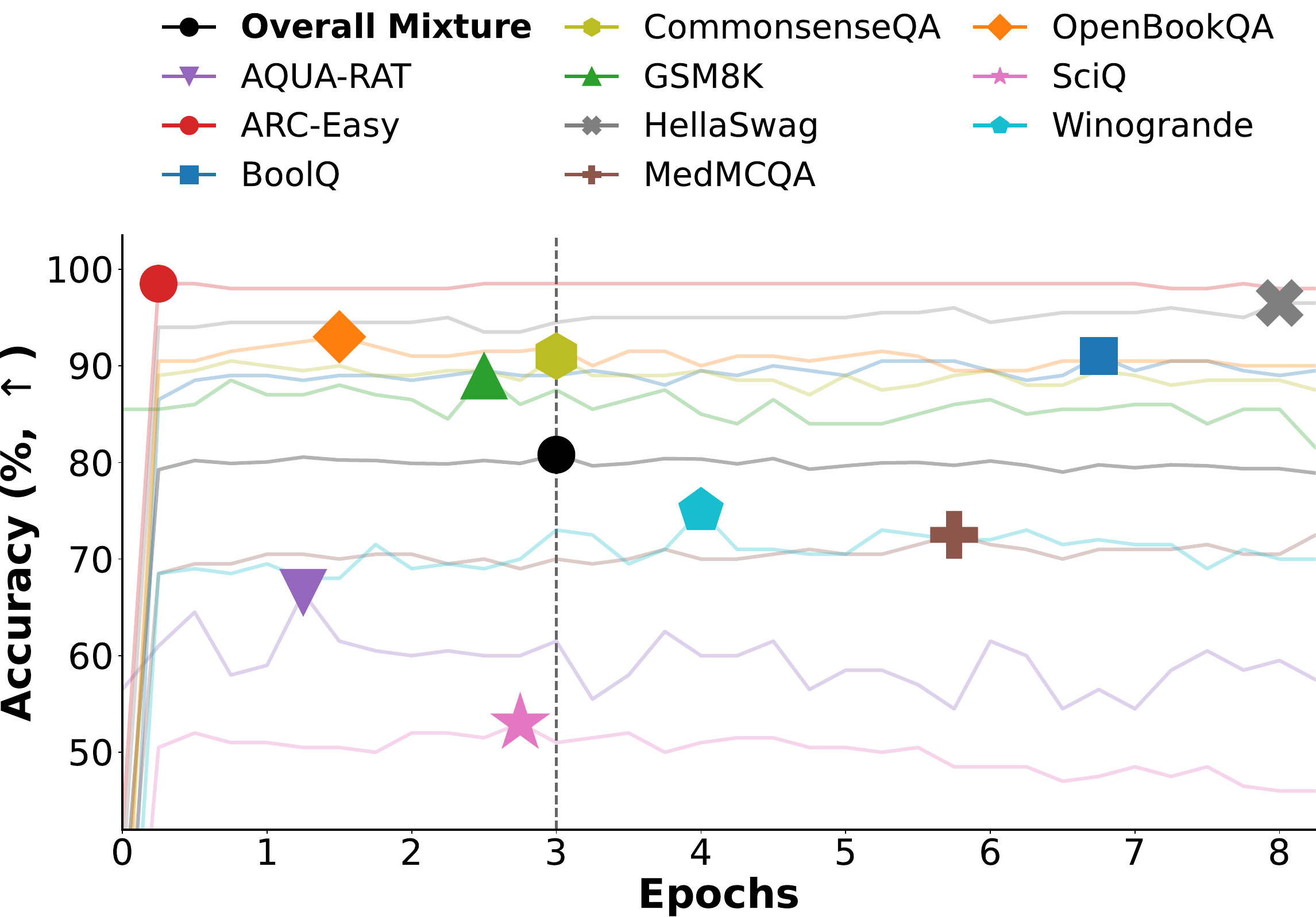}
        \caption{Validation set accuracy curves across sub-tasks with annotation at peak performance.}
        \label{fig:hetero}
    \end{subfigure}\hfill
    \begin{subfigure}{0.495\linewidth}
        \centering
        \includegraphics[width=\linewidth]{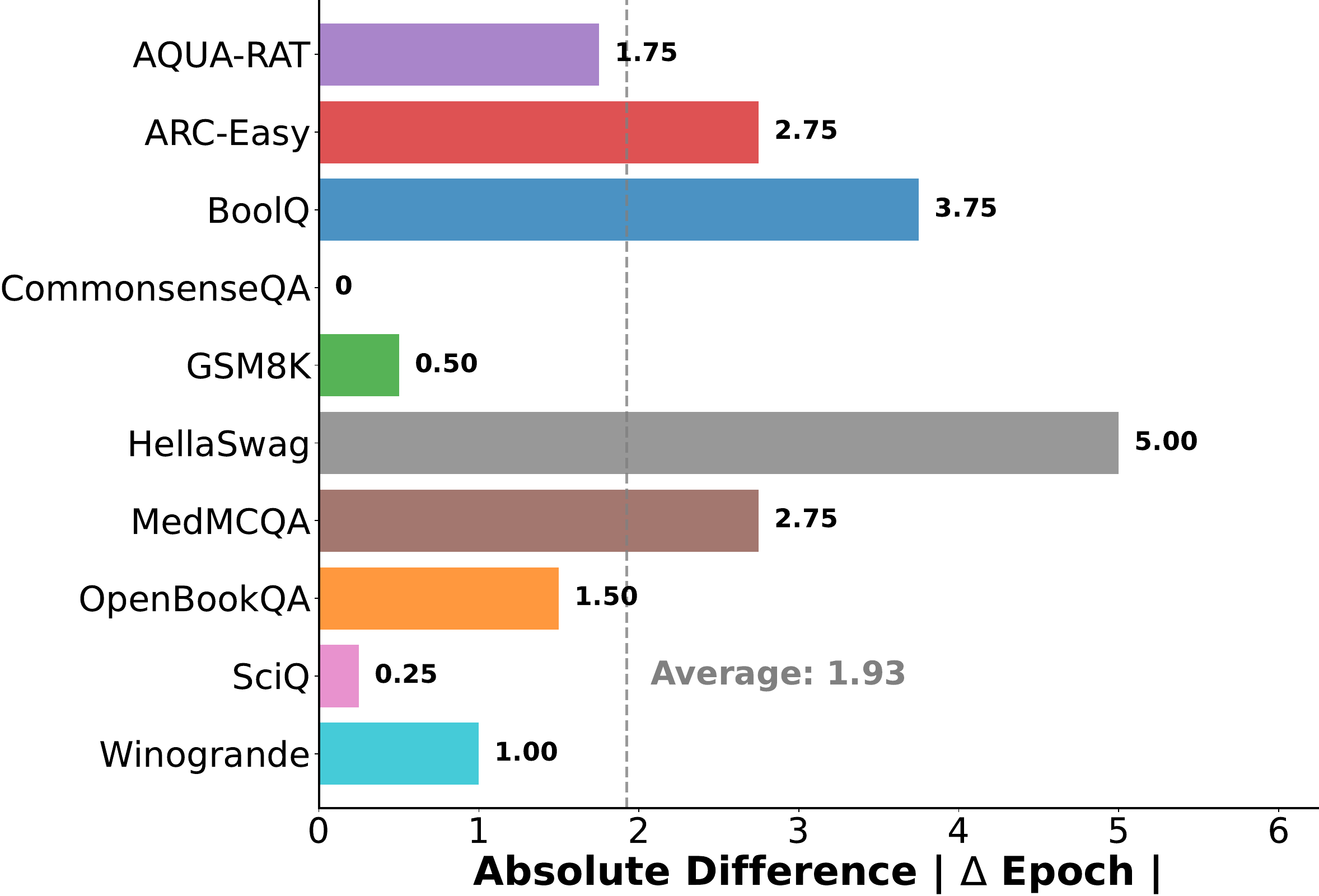}
        \caption{Absolute peak epoch difference of overall mixture and individual sub-datasets.}
        \label{fig:delta_epoch}
    \end{subfigure}
    
    \caption{\textbf{Heterogeneous learning dynamics.} Multi-task SFT on \ttt{Qwen3 8B} demonstrates that underlying sub-datasets overfitting dynamics vary greatly. This observation is consistent across all other models; visualized in Appendix \ref{app:motivation_app}.}
    \label{fig:motivation}
\end{figure*}

Consequently, enforcing a single global compute budget inevitably produces sub-optimal outcomes across the mixture due to \textit{heterogeneous learning dynamics}. Because distinct tasks differ significantly in data distribution and complexity, their convergence rates and optimal compute levels vary widely ($c^*_i \neq c^*_j$). Empirically, individual sub-datasets reach peak generalization performance at substantially different compute levels (see Fig. \ref{fig:motivation}). Thus, applying $c_{\text{global}}$ creates an inherent optimization conflict: rapidly converging tasks begin to overfit when $c_{\text{global}} > c^*_i$, while slower-learning tasks remain under-fitted when $c_{\text{global}} < c^*_i$.

\section{\tsc{mSFT}: Heterogeneous Early-stopping for Multi-task Data Mixtures}\label{sec:msft}

\subsection{Limitation of a Naïve Solution}\label{sec:naive}

A straightforward solution to heterogenous overfitting (as visualized in Fig. \ref{fig:motivation}) is leveraging the optimal compute found for each sub-dataset in Fig. \ref{fig:hetero} and exclude these sub-datasets at these points during a new training run. We name this method single roll-out search SFT (SRO SFT), and embodies two stages: \textbf{(i)} single roll-out search (Fig. \ref{fig:hetero}), and \textbf{(ii)} train from scratch with heterogeneous exclusion. For instance, in the example in Fig. \ref{fig:hetero}, in stage \textbf{(ii)}, AQUA-RAT would be excluded in epoch 1.25, while SciQ would be excluded in epoch 2.75. Pseudocode is available in Appendix \ref{app:baselines}.

However, the key limitation of SRO search is that the optimal compute found during the search stage is an approximation after the first sub-dataset is excluded. Formally, let the model parameter update at step $t$ be driven by the aggregate gradient of the active dataset mixture. In the search stage \textbf{(i)}, the exclusion set is empty ($\mc{E} = \emptyset$), so the update is a summation over all tasks $i$ in $\mc{D}$:
\begin{equation}
    \Delta \theta_t \propto \sum_{\mc{D}_i \in \mc{D}} w_i \nabla \mc{L}(\theta_t; \mc{D}_i),
    \label{eq:theta}
\end{equation}
where $w_i$ is the weight of the sub-dataset $i$. Consequently, the optimal compute budget $c^*_i$ for any specific task $i$ is conditional on the gradient interactions from the complete mixture.

However, in the SRO training stage \textbf{(ii)}, once a sub-dataset $\mc{D}_{\text{exclude}}$ is added to the exclusion set $\mc{E}$, the update rule shifts to:
\begin{equation}
    \Delta \theta'_t \propto \sum_{\mc{D}_i \in \mc{D} \setminus \mc{E}} w_i \nabla \mc{L}(\theta'_t; \mc{D}_i)
     \label{eq:theta_prime}
\end{equation}
The removal of $\nabla \mc{L}(\cdot; \mc{D}_{\text{exclude}})$ causes the optimization trajectory to diverge ($\theta'_t \neq \theta_t$). Crucially, this drift exacerbates as $|\mc{E}|$ increases: as more tasks are dropped over time, the active gradient sum deviates further from the original search dynamics, rendering the pre-computed $c^*_i$ increasingly inaccurate for late-stage tasks.

\begin{figure*}[tbp]
    \centering
    \begin{subfigure}[b]{0.495\textwidth}
        \centering
        \includegraphics[width=\linewidth]{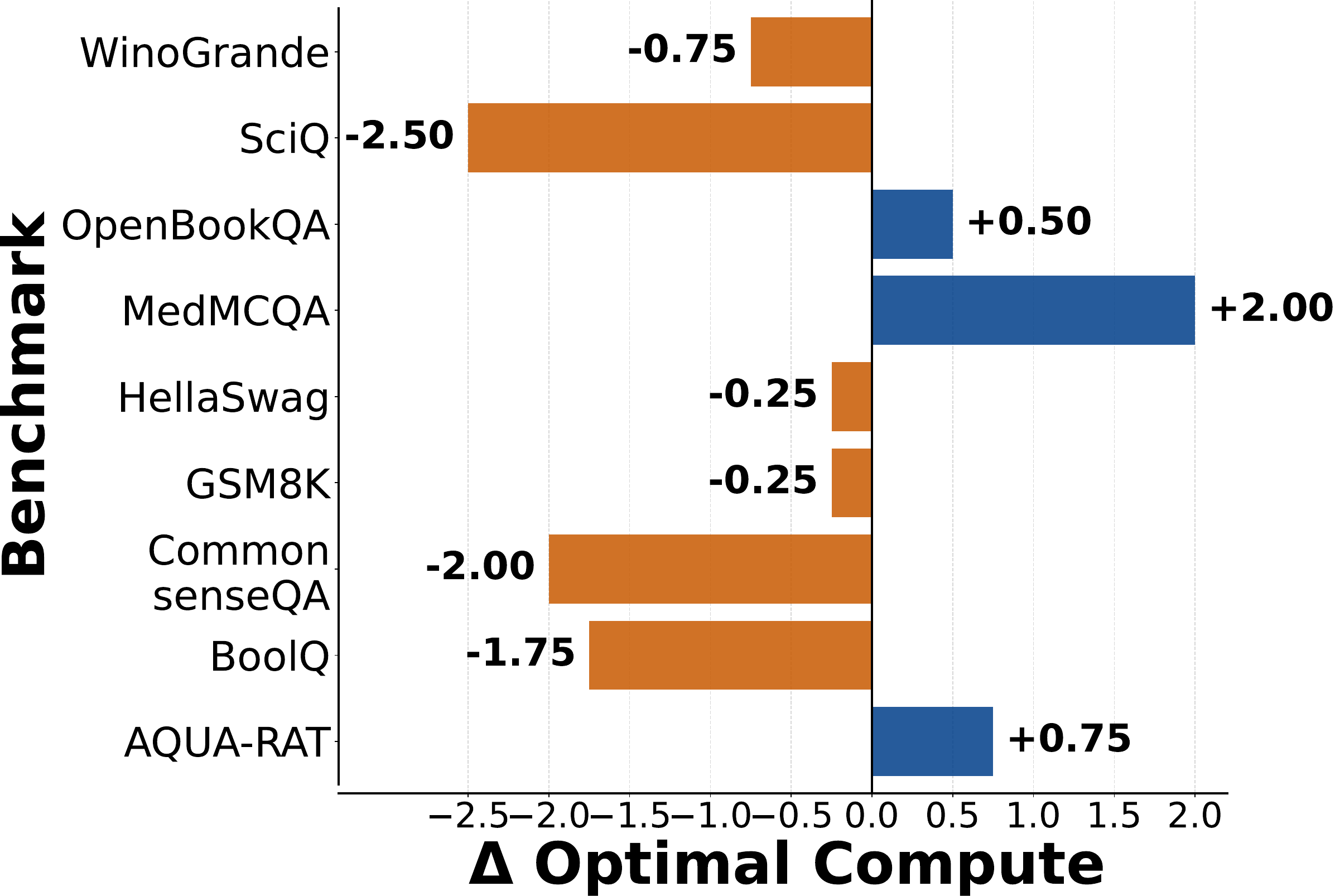}
        \caption{$\Delta c^*_j$ ($\Delta$ Optimal Compute) for individual benchmarks on \ttt{Qwen3 8B}.}
        \label{fig:delta_benchmark}
    \end{subfigure}
    \hfill
    \begin{subfigure}[b]{0.495\textwidth}
        \centering
        \includegraphics[width=\linewidth]{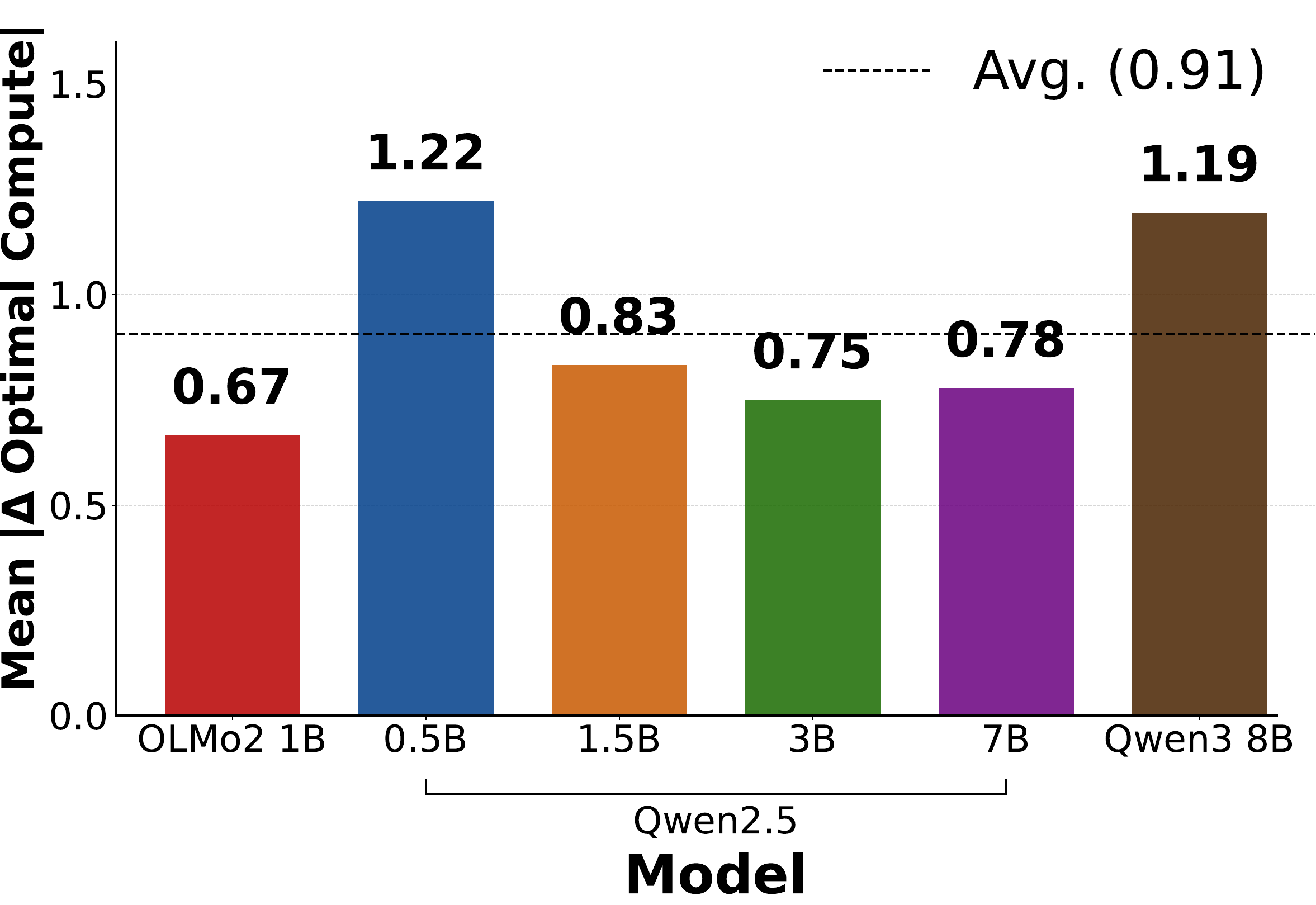}
        \caption{Mean absolute shift in optimal compute across various model architectures and scales.}
        \label{fig:delta_mean}
    \end{subfigure}
    \caption{\textbf{Divergence of optimal compute upon dataset exclusion.} Excluding a small fraction of the training mixture alters the optimization trajectory, shifting optimal stopping points for remaining tasks. (a) $\Delta$ optimal compute varies across individual sub-tasks. (b) This divergence is consistent across model families and scales, averaging an absolute shift of 0.91 epochs. Detailed decomposition across other models available in Appendix \ref{app:results_delta}}
    \label{fig:diverge}
\end{figure*}

\paragraph{Empirical Analysis.} 
We empirically validate whether the parameter divergence $\theta'_t \neq \theta_t$ (Eq. \ref{eq:theta}, \ref{eq:theta_prime}) translates into shifted optimal compute. We construct an equal-weighted mixture of $N=10$ sub-datasets, each containing $\vert \mc{D}_i \vert = 1800$ samples. We train a model on the full mixture $\mc{D}$ until the first sub-dataset, which we denote as $\mc{D}_k$, overfits. At this exact checkpoint, we bifurcate the training process into two branches: one continues training on the full mixture $\mc{D}$, while the other continues on the reduced mixture $\mc{D} \setminus \{\mc{D}_k\}$. For each of the 9 remaining tasks ($j \neq k$), we compare the optimal compute achieved on the full mixture ($c^*_j$) against the optimal compute on the reduced mixture ($c'^*_j$). We report the shift, defined as $\Delta c^*_j := c'^*_j - c^*_j$, in Fig. \ref{fig:diverge}. The results clearly demonstrate that excluding even a small fraction of the training data (1/10) significantly alters the optimal stopping points for the remaining tasks, confirming our hypothesis that $c'^*_j \neq c^*_j$.

\subsection{Iterative Overfitting-Aware Search}\label{sec:ours}

In response to this limitation, we propose \tsc{mSFT}, a training algorithm that ensures that the search and train phase is aligned. \tsc{mSFT} follows an iterative \textcolor{mygreen}{\textbf{roll-out}} and \textcolor{myred}{\textbf{roll-back}} search algorithm described below and conceptualized in Alg. \ref{alg:msft}.

\paragraph{Initialization.} First, the algorithm initializes the exclusion set $\mc{E}$ that keeps track of the excluded sub-datasets, and the parameter $\hat{\theta}$ is set to the base model $\theta_0$ (line 1). The algorithm loops as long as there is at least one active sub-dataset (line 2).

\paragraph{\textcolor{mygreen}{Roll-out}.} For every active sub-dataset $\mc{D} \setminus \mc{E}$ the model $\hat{\theta}$ is trained by a pre-determined compute budget $C$ hyperparameter (line 3). $C$ is analogous to epochs in the literature, however, we call it compute budget (e.g., 1/4 of an epoch) as we aim to record more granular levels of compute as we observe granular overfitting behavior in our preliminary analysis in Fig. \ref{fig:motivation} and Appendix \ref{app:motivation_app}. For each active sub-dataset, the optimal compute is recorded (line 4). The sub-dataset that over-fitted earliest is expected to be excluded $\mc{D}_\text{exclude}$ (line 5). In the rare case that no sub-dataset $\mc{D}_i$ over-fitted within the compute budget $C$, the algorithm continues without rolling back.   
\paragraph{\textcolor{myred}{Roll-back}.} The earliest over-fitted dataset $\mc{D}_\text{exclude}$ will no longer be included in the active set (line 9), and the model is reverted to the point at which it overfit (line 10).

\begin{algorithm2e}[tbh]
    \caption{\tsc{mSFT}}
    \label{alg:msft}
    \SetAlgoLined
    \LinesNumbered
    \Input{Dataset mixture $\mc{D}$, base model $\theta_0$, compute budget $C$}

    $\mc{E} \leftarrow \emptyset ;$  $\hat{\theta} \leftarrow \theta_0$ \tcp*{Initialization}

    \While{$\mc{D} \setminus \mc{E} \neq \emptyset$}{
        \tcc{\textcolor{mygreen}{\textbf{Roll-out}}: Search for per-sub-dataset peaks}
        $\theta, \;\{\text{acc}(\mc{D}_i, c)\}_{i,c} \leftarrow \tsc{SFT-Roll-out}\!\left(\hat{\theta},\; \mc{D} \setminus \mc{E},\; C\right)$ \;
        
        $c_i^* \leftarrow \arg\max_{c}\; \text{acc}(\mc{D}_i, c) \quad \forall \mc{D}_i \notin \mc{E}$ \tcp*{Optimal compute per sub-dataset}

        $c_{\min}, \mc{D}_\text{exclude} \leftarrow \arg\min_{\mc{D}_i \notin \mc{E}}\; c_i^*$ \;

        \eIf{$c_{\min} = C$}{
            \tcc{No overfitting: update model and continue}
            $\hat{\theta} \leftarrow \theta(C)$ \;
        }{
            \tcc{\textcolor{myred}{\textbf{Roll-back}}: Revert to the checkpoint where the sub-dataset overfit}
            $\mc{E} \leftarrow \mc{E} \cup \{\mc{D}_{\text{exclude}}\}$ \;
            $\hat{\theta} \leftarrow \theta(c_{\min})$ \tcp*{Revert to checkpoint at $c_{\min}$}
        }
    }
\end{algorithm2e}

\section{Empirical Study} \label{sec:exp}

\subsection{Experiment Set-up}

\paragraph{Base Models.} For a broad range of model sizes and families, we employ \ttt{OLMo 2 1B} \citep{walsh2025}, \ttt{Qwen2.5 0.5, 1.5, 3, 7B} \citep{qwen2025qwen25technicalreport}, and \ttt{Qwen3 8B} \citep{yang2025qwen3technicalreport}. 

\paragraph{Baselines.} We compare our approach with four baselines: \textbf{[1]} standard SFT \citep{rastogi2025magistral,groeneveld-etal-2024-olmo,walsh2025,olmo2025olmo,liu2024deepseek,guo2025deepseek,qwen2025qwen25technicalreport,yang2025qwen3technicalreport,adler2024nemotron}, the \textit{de facto} norm , \textbf{[2]} continual SFT \citep{scialom-etal-2022-fine} which trains each of the sub-datasets sequentially, allowing each of them to arrive at the optimal early-stopping point, \textbf{[3]} DynamixSFT \citep{shin2025dynamixsft} which optimizes dataset mixture ratios using multi-armed bandits with 1-step roll-out, and \textbf{[4]} Instance-dependant Early Stopping (IES; \cite{yuan2025instancedependent}) which computes second-order differences for each instance, and leverages a threshold hyperparameter for exclusion.

\paragraph{Training and Evaluation Setting.} For fair comparison, all overlapping training configurations are equalized across methods. Overlapping hyperparameters were optimized for standard SFT. We use $N=10$ sub-datasets: CommonsenseQA \citep{talmor-etal-2019-commonsenseqa}, OpenBookQA \citep{mihaylov-etal-2018-suit}, AQUA-RAT \citep{ling-etal-2017-program}, GSM8K \citep{cobbe2021trainingverifierssolvemath}, SciQ \citep{welbl-etal-2017-crowdsourcing}, ARC-Easy \citep{clark2018thinksolvedquestionanswering}, HellaSwag \citep{zellers-etal-2019-hellaswag}, Winogrande \citep{sakaguchi2020winogrande}, BoolQ \citep{clark-etal-2019-boolq}, and MedMCQA \citep{pmlr-v174-pal22a}. All methods are greedy decoding evaluated 5-shot \citep{brown2020language} on the validation set in intervals of 1/4 epochs, with the best performing checkpoint being reported.  Further training details can be found in Appendix \ref{app:further_exp}.

\begin{table*}[tbp]
\centering
\resizebox{\textwidth}{!}{
\begin{tabular}{l *{14}{r}}
\toprule
\textbf{Model:} & \multicolumn{2}{c}{\textbf{\ttt{OLMo 2}}} & \multicolumn{8}{c}{\textbf{\ttt{Qwen2.5}}} & \multicolumn{2}{c}{\textbf{\ttt{Qwen3}}} & \multicolumn{2}{c}{} \\
\cmidrule(lr){2-3} \cmidrule(lr){4-11} \cmidrule(lr){12-13}
\textbf{Size:} & \multicolumn{2}{c}{\textbf{\ttt{1B}}} & \multicolumn{2}{c}{\textbf{\ttt{0.5B}}} & \multicolumn{2}{c}{\textbf{\ttt{1.5B}}} & \multicolumn{2}{c}{\textbf{\ttt{3B}}} & \multicolumn{2}{c}{\textbf{\ttt{7B}}} & \multicolumn{2}{c}{\textbf{\ttt{8B}}} & \multicolumn{2}{c}{\cellcolor{gray!15}\textbf{Average}} \\
\cmidrule(lr){2-3} \cmidrule(lr){4-5} \cmidrule(lr){6-7} \cmidrule(lr){8-9} \cmidrule(lr){10-11} \cmidrule(lr){12-13} \cmidrule(lr){14-15}
& \textbf{Acc.} & \textbf{Ep.} & \textbf{Acc.} & \textbf{Ep.} & \textbf{Acc.} & \textbf{Ep.} & \textbf{Acc.} & \textbf{Ep.} & \textbf{Acc.} & \textbf{Ep.} & \textbf{Acc.} & \textbf{Ep.} & \cellcolor{gray!15}\textbf{Acc.} & \cellcolor{gray!15}\textbf{Ep.} \\
\midrule
\multicolumn{15}{c}{\cellcolor{myblue!15}\textbf{Science and Knowledge}} \\
\midrule
Base & 32.4 & --- & 26.1 & --- & 54.6 & --- & 12.1 & --- & 4.0 & --- & 24.6 & --- & \cellcolor{gray!15}25.6 & \cellcolor{gray!15}--- \\
SFT & 47.9 & 9.75 & 37.5 & 0.50 & 65.8 & 3.00 & 71.8 & 5.00 & 74.5 & 2.00 & 77.9 & 3.00 & \cellcolor{gray!15}62.5 & \cellcolor{gray!15}3.88 \\
Continual SFT & 48.5 & 1.90 & 24.6 & 1.95 & 66.6 & 2.08 & 71.4 & 1.80 & 72.9 & 1.40 & 77.5 & 1.15 & \cellcolor{gray!15}60.2$_{\color{myred}-2.3}$ & \cellcolor{gray!15}1.71 \\
DynamixSFT & 47.9 & 5.75 & 39.5 & 0.50 & 65.6 & 2.75 & 71.5 & 3.00 & 74.5 & 7.25 & 75.2 & 5.00 & \cellcolor{gray!15}62.4$_{\color{myred}-0.1}$ & \cellcolor{gray!15}4.04 \\
IES & 47.6 & 10.00 & 39.5 & 0.50 & 65.4 & 4.00 & 71.9 & 3.50 & 74.4 & 3.00 & 78.1 & 2.25 & \cellcolor{gray!15}\underline{62.8}$_{\color{mygreen}+0.3}$ & \cellcolor{gray!15}3.88 \\
\tsc{mSFT} (\textbf{ours}) & 50.4 & 9.75 & 39.2 & 0.25 & 65.4 & 4.75 & 72.9 & 5.50 & 73.6 & 1.50 & 78.0 & 3.00 & \cellcolor{gray!15}\textbf{63.2}$_{\color{mygreen}+0.7}$ & \cellcolor{gray!15}4.12 \\
\midrule
\multicolumn{15}{c}{\cellcolor{myblue!15}\textbf{Commonsense and Language}} \\
\midrule
Base & 9.9 & --- & 22.2 & --- & 42.5 & --- & 8.1 & --- & 8.4 & --- & 19.0 & --- & \cellcolor{gray!15}18.4 & \cellcolor{gray!15}--- \\
SFT & 50.9 & 9.75 & 32.9 & 0.50 & 73.0 & 3.00 & 81.6 & 5.00 & 84.2 & 2.00 & 86.9 & 3.00 & \cellcolor{gray!15}68.2 & \cellcolor{gray!15}3.88 \\
Continual SFT & 48.6 & 1.90 & 19.0 & 1.95 & 71.1 & 2.08 & 80.2 & 1.80 & 86.1 & 1.40 & 86.0 & 1.15 & \cellcolor{gray!15}65.2$_{\color{myred}-3.0}$ & \cellcolor{gray!15}1.71 \\
DynamixSFT & 49.0 & 5.75 & 39.9 & 0.50 & 72.6 & 2.75 & 83.0 & 3.00 & 84.6 & 7.25 & 84.9 & 5.00 & \cellcolor{gray!15}69.0$_{\color{mygreen}+0.8}$ & \cellcolor{gray!15}4.04 \\
IES & 51.0 & 10.00 & 38.8 & 0.50 & 72.6 & 4.00 & 82.4 & 3.50 & 85.5 & 3.00 & 86.1 & 2.25 & \cellcolor{gray!15}\underline{69.4}$_{\color{mygreen}+1.2}$ & \cellcolor{gray!15}3.88 \\
\tsc{mSFT} (\textbf{ours}) & 53.8 & 9.75 & 42.5 & 0.25 & 72.8 & 4.75 & 80.6 & 5.50 & 86.5 & 1.50 & 87.6 & 3.00 & \cellcolor{gray!15}\textbf{70.6}$_{\color{mygreen}+2.4}$ & \cellcolor{gray!15}4.12 \\
\midrule
\multicolumn{15}{c}{\cellcolor{myblue!15}\textbf{Mathematic and Quantitative}} \\
\midrule
Base & 19.5 & --- & 26.2 & --- & 42.8 & --- & 58.0 & --- & 68.0 & --- & 71.0 & --- & \cellcolor{gray!15}47.6 & \cellcolor{gray!15}--- \\
SFT & 20.2 & 9.75 & 24.2 & 0.50 & 43.0 & 3.00 & 59.5 & 5.00 & 66.5 & 2.00 & 74.5 & 3.00 & \cellcolor{gray!15}\underline{48.0} & \cellcolor{gray!15}3.88 \\
Continual SFT & 18.5 & 1.90 & 23.8 & 1.95 & 45.0 & 2.08 & 60.0 & 1.80 & 67.0 & 1.40 & 72.5 & 1.15 & \cellcolor{gray!15}47.8$_{\color{myred}-0.2}$ & \cellcolor{gray!15}1.71 \\
DynamixSFT & 20.8 & 5.75 & 25.0 & 0.50 & 43.2 & 2.75 & 58.2 & 3.00 & 65.8 & 7.25 & 74.2 & 5.00 & \cellcolor{gray!15}47.9$_{\color{myred}-0.1}$ & \cellcolor{gray!15}4.04 \\
IES & 21.5 & 10.00 & 25.5 & 0.50 & 43.0 & 4.00 & 60.2 & 3.50 & 65.2 & 3.00 & 72.5 & 2.25 & \cellcolor{gray!15}\underline{48.0}$_{\color{myred}-0.0}$ & \cellcolor{gray!15}3.88 \\
\tsc{mSFT} (\textbf{ours}) & 23.2 & 9.75 & 23.5 & 0.25 & 48.8 & 4.75 & 64.2 & 5.50 & 70.0 & 1.50 & 76.0 & 3.00 & \cellcolor{gray!15}\textbf{51.0}$_{\color{mygreen}+3.0}$ & \cellcolor{gray!15}4.12 \\
\midrule
\multicolumn{15}{c}{\cellcolor{myorange!15}\textbf{Average Accuracy Across 10 Benchmarks}} \\
\midrule
Base & 20.8 & --- & 24.6 & --- & 47.4 & --- & 19.7 & --- & 18.6 & --- & 31.6 & --- & \cellcolor{gray!15}27.1 & \cellcolor{gray!15}--- \\
SFT & 43.6 & 9.75 & 33.0 & 0.50 & \underline{64.1} & 3.00 & 73.2 & 5.00 & 76.8 & 2.00 & \underline{80.8} & 3.00 & \cellcolor{gray!15}61.9 & \cellcolor{gray!15}3.88 \\
Continual SFT & 42.6 & 1.90 & 22.2 & 1.95 & \underline{64.1} & 2.08 & 72.6 & 1.80 & \underline{77.0} & 1.40 & 79.9 & 1.15 & \cellcolor{gray!15}59.7$_{\color{myred}-2.2}$ & \cellcolor{gray!15}1.71 \\
DynamixSFT & 42.9 & 5.75 & \underline{36.8} & 0.50 & 64.0 & 2.75 & 73.4 & 3.00 & 76.8 & 7.25 & 78.9 & 5.00 & \cellcolor{gray!15}62.1$_{\color{mygreen}+0.2}$ & \cellcolor{gray!15}4.04 \\
IES & \underline{43.8} & 10.00 & 36.4 & 0.50 & 63.8 & 4.00 & \underline{73.8} & 3.50 & 77.0 & 3.00 & 80.2 & 2.25 & \cellcolor{gray!15}\underline{62.5}$_{\color{mygreen}+0.6}$ & \cellcolor{gray!15}3.88 \\
\tsc{mSFT} (\textbf{ours}) & \textbf{46.3} & 9.75 & \textbf{37.4} & 0.25 & \textbf{65.0} & 4.75 & \textbf{74.2} & 5.50 & \textbf{78.0} & 1.50 & \textbf{81.4} & 3.00 & \cellcolor{gray!15}\textbf{63.7}$_{\color{mygreen}+1.8}$ & \cellcolor{gray!15}4.12 \\
\bottomrule
\end{tabular}
}
\caption{\textbf{Main results.} Comparison of six methodologies across six underlying models (\ttt{OLMo 2}, \ttt{Qwen2.5}, and \ttt{Qwen3}), evaluating performance across three major task categories. We report both accuracy (Acc.) and the epoch (Ep.) at which the best accuracy was achieved. Continual SFT's Ep. is the average across benchmarks making values not in intervals of 1/4 epochs like others. The best scores are \textbf{bolded}, and second best \underline{underlined}.}
\label{tab:main}
\end{table*}

\subsection{Main Results}\label{sec:exp_main_result}

\paragraph{Overall Performance and Robustness.} As detailed in Tab. \ref{tab:main}, \tsc{mSFT} consistently outperforms all baseline methodologies across the six evaluated models (\ttt{OLMo 2}, \ttt{Qwen2.5}, \ttt{Qwen3}), achieving the highest average accuracy. While advanced baselines like DynamixSFT and IES yield marginal gains, and Continual SFT suffers from catastrophic forgetting (-2.2\%), \tsc{mSFT} remains uniquely robust. It is the only approach to exhibit consistent improvements across all three major domains: Science \& Knowledge (+0.7\%), Commonsense \& Language (+2.4\%), and Mathematical \& Quantitative reasoning (+3.0\%).

\begin{figure*}[tbp]
    \centering
    \begin{minipage}{0.495\linewidth}
        \centering
         \label{fig:std}
         \includegraphics[width=\linewidth]{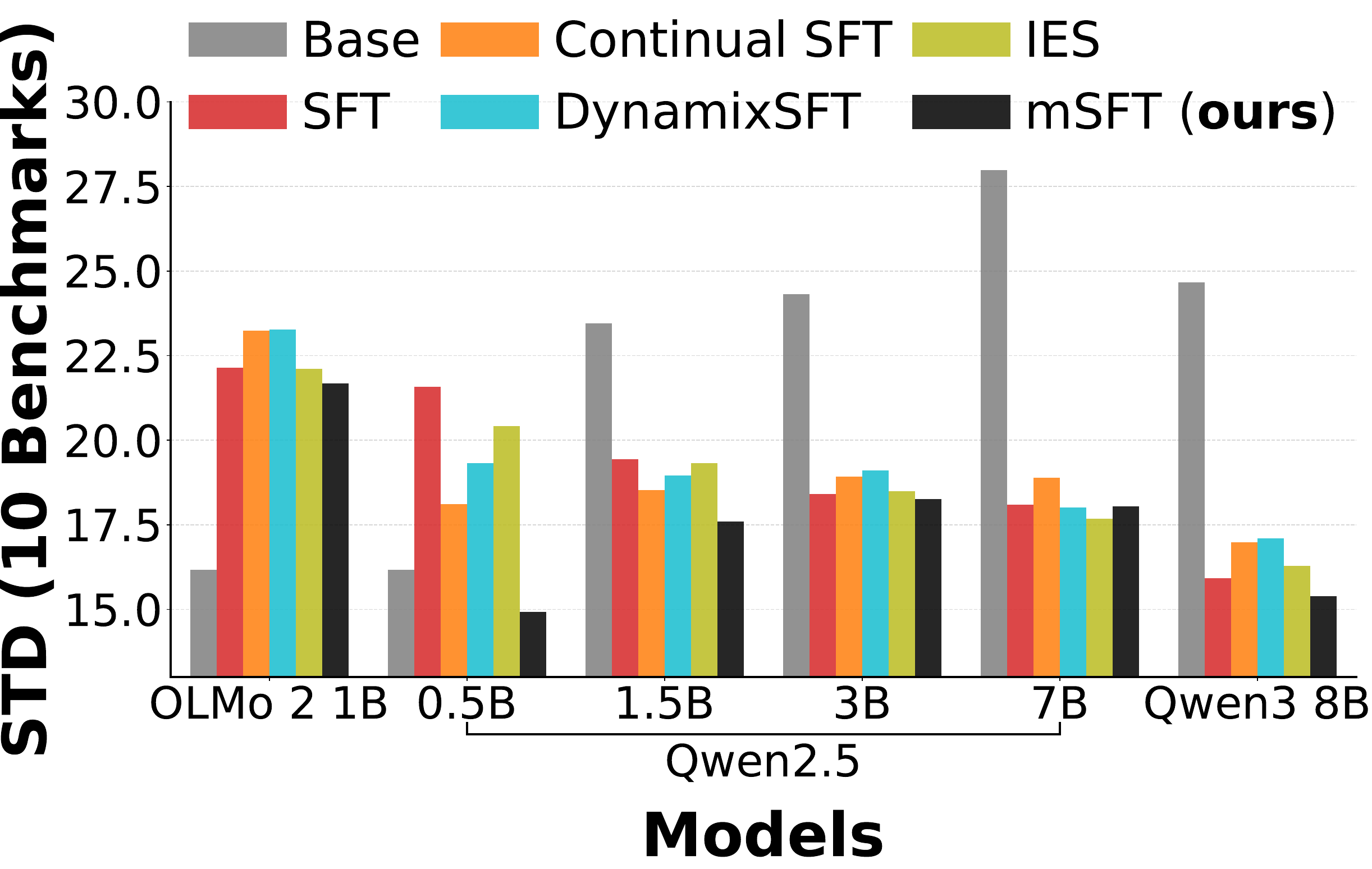}
    \end{minipage}\hfill
    \begin{minipage}{0.495\linewidth}
        \centering
        \includegraphics[width=\linewidth]{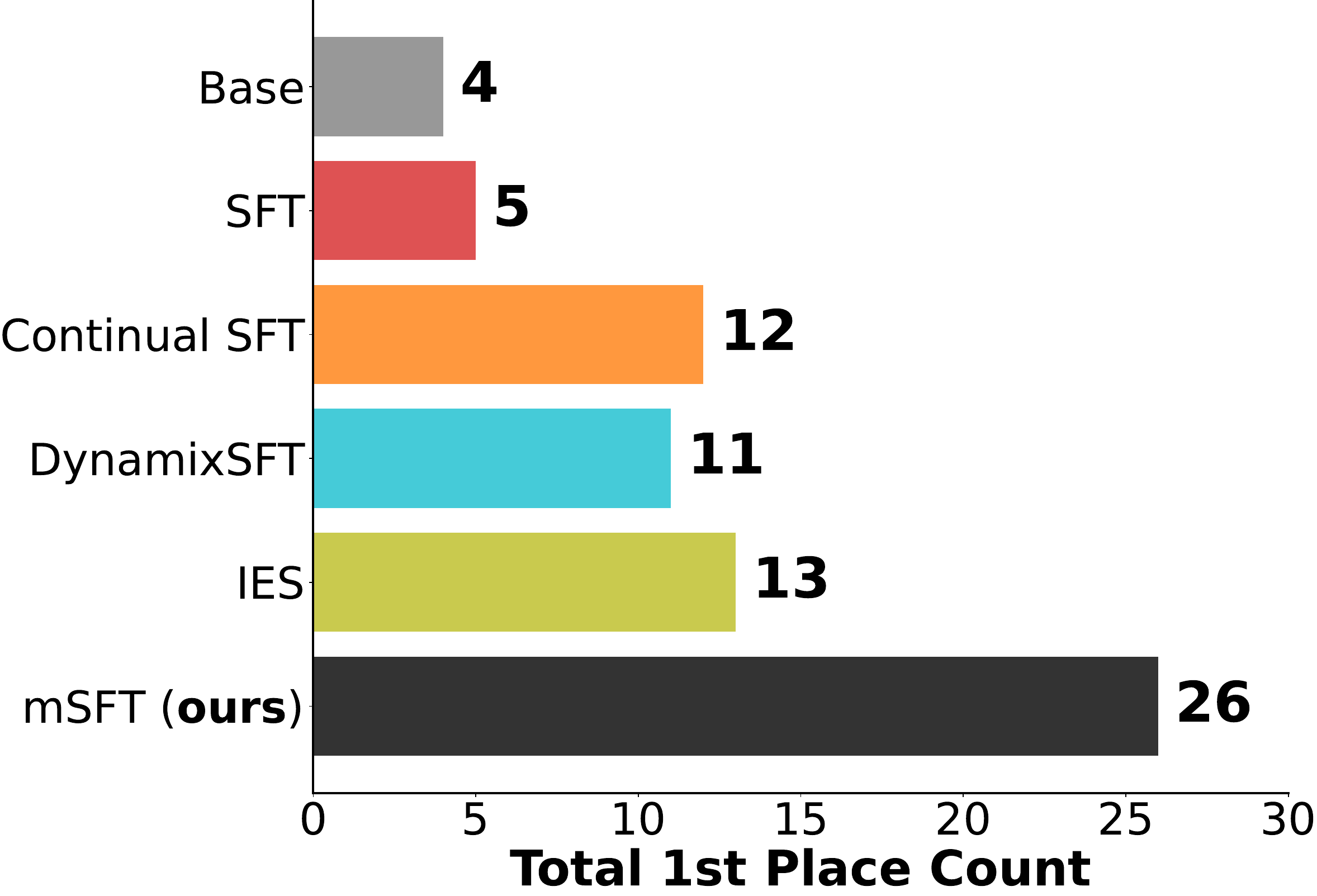}
        \label{fig:right_placeholder}
    \end{minipage}
    
\caption{\textbf{Further details of main results.} \textbf{[left]} \tsc{mSFT} achieves the lowest levels of standard deviation across benchmarks (STD), indicating performance gains are not due to large outliers. \textbf{[right]} Across models, \tsc{mSFT} achieves 1st place the most. The 1st place count does not add up to 60 = 6 $\cdot$ 10 (models $\cdot$ benchmarks) as there are cases where 1st place is tied.}
    \label{fig:further}
\end{figure*}

\paragraph{Consistency and Outlier Analysis.} Beyond aggregate accuracy, \tsc{mSFT} demonstrates superior systematic stability. As illustrated in Fig. \ref{fig:further} [left], it generally maintains the lowest standard deviation across benchmarks, confirming that the average improvements stem from uniformly distributed gains rather than skewed outlier performances. Furthermore, Fig. \ref{fig:further} [right] shows that \tsc{mSFT} achieves 1st place on individual benchmarks 26 times across all model configurations, doubling the frequency of the next best baseline (IES, 13 times). This affirms that \tsc{mSFT} reliably elevates both the performance floor and ceiling across a diverse suite of tasks.

\begin{figure*}[tbp]
    \centering
    \begin{minipage}{0.495\linewidth}
        \centering
        \includegraphics[width=\linewidth]{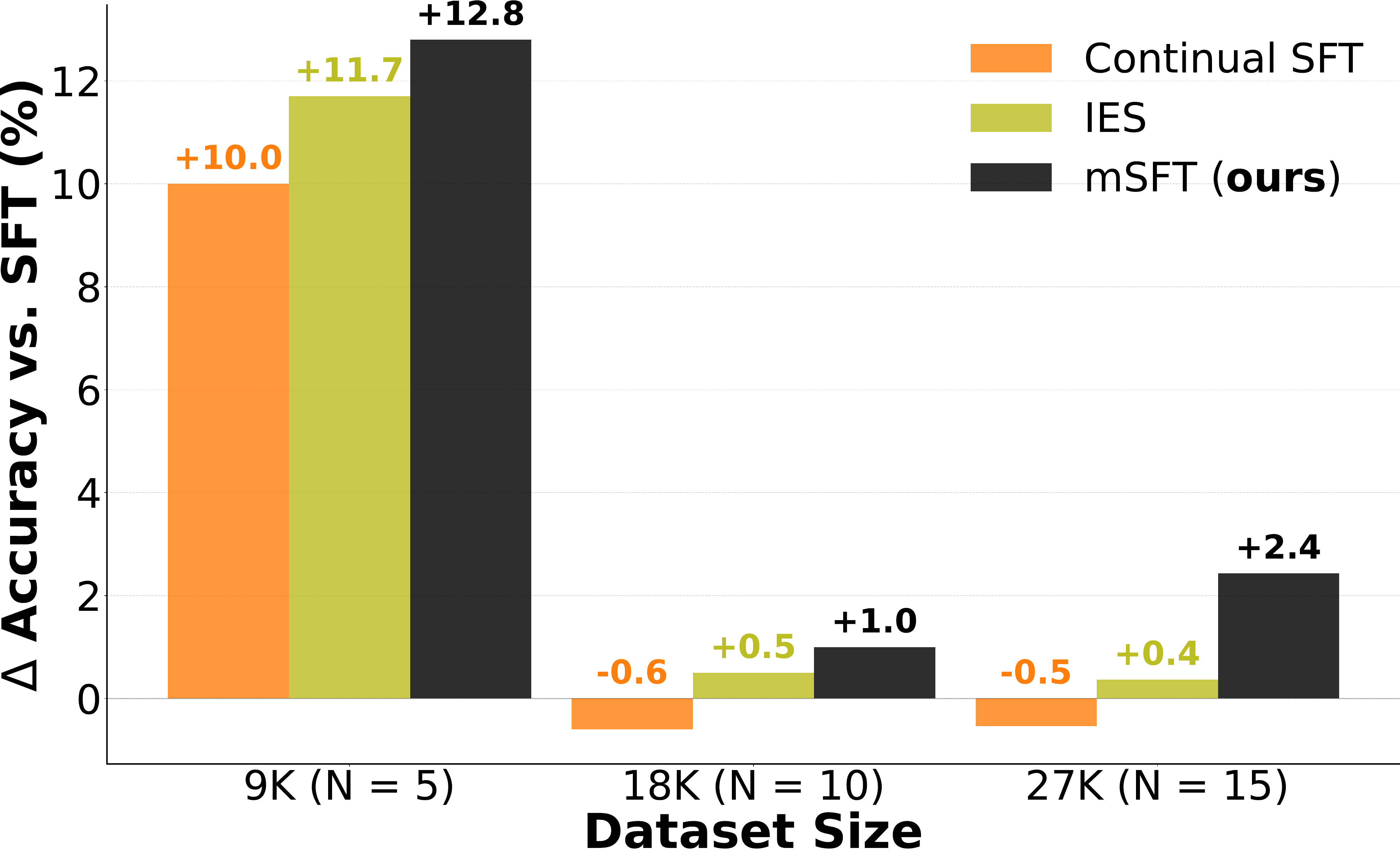}
        \caption{\textbf{Robustness across varying dataset sizes.} $\Delta$ Accuracy of Continual SFT, IES, and \tsc{mSFT} relative to SFT. \tsc{mSFT} consistently achieves the highest performance gains across different total dataset sizes and tasks ($N$), avoiding the degradation seen in Continual SFT at larger scales.}
        \label{fig:diverse_N}
    \end{minipage}\hfill
    \begin{minipage}{0.495\linewidth}
        \centering
        \includegraphics[width=\linewidth]{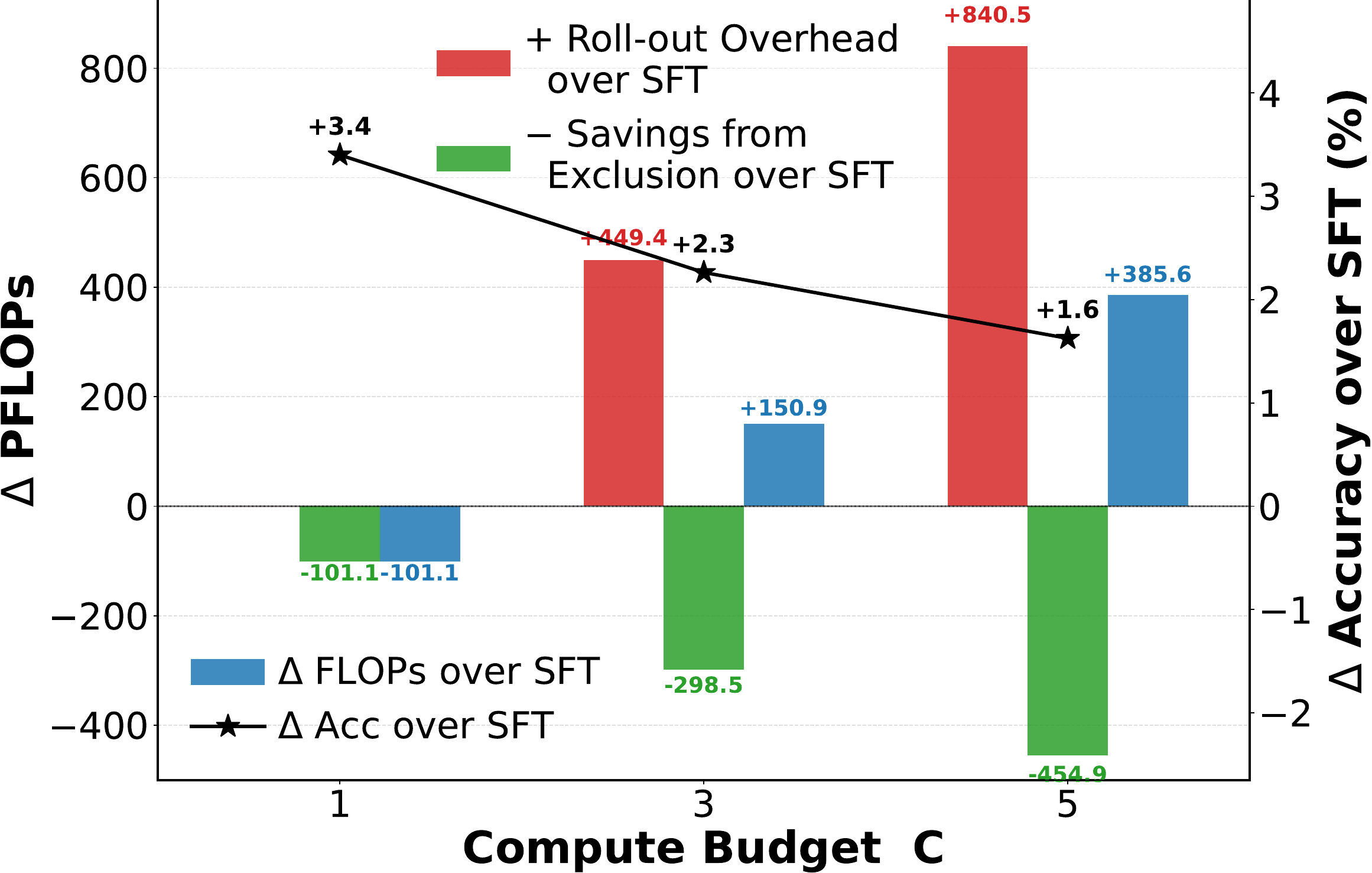}
        \caption{\textbf{Accuracy and FLOPs across compute budget.} FLOPs decomposition (bars) and $\Delta$ Accuracy (line) of \tsc{mSFT} across compute budgets ($C$). At $C=1$, zero roll-out overhead lets \tsc{mSFT} gain accuracy while strictly reducing net compute; for $C>1$, roll-out overhead outweighs the savings from halting completed tasks.}
        \label{fig:granular_flops}
    \end{minipage}
\end{figure*}

\subsection{Ablation Study}\label{sec:abl}

\paragraph{Set-up.} We examine two naïve alternative heterogeneous early-stopping algorithms, which serve as ablation studies: \textbf{[4]} Single roll-out searched SFT (SRO SFT), and \textbf{[5]} Soft SRO SFT. SRO SFT is introduced in \S\ \ref{sec:naive}, and Soft SRO SFT is the soft version, which aims to replicate SRO SFT via mixture ratios rather than hard exclusions, reducing catastrophic forgetting. SRO SFT and Soft SRO SFT are introduced with pseudo-codes in Appendix \ref{app:baselines}.

\paragraph{Result.} As observed in Tab. \ref{tab:ablation}, \tsc{mSFT}'s average performance is superior to both SRO SFT and Soft SRO SFT. This verifies that the naïve approach of using approximate optimal compute $c_i^*$ through single roll-out search introduced in \S\ \ref{sec:naive} is sub-optimal.

\begin{table}[bp]
\centering
\small
\begin{tabular}{l rr}
\toprule
& \multicolumn{2}{c}{\textbf{Average}} \\
\cmidrule(lr){2-3} 
& \textbf{Acc.} & \textbf{Ep.} \\
\midrule
SFT & 61.9 & 3.88 \\
SRO SFT & \underline{63.4} & 3.75 \\
Soft SRO SFT & 62.1 & 3.79 \\
\tsc{mSFT} (\textbf{ours}) & \textbf{63.7} & 4.12 \\
\bottomrule
\end{tabular}
\caption{\textbf{Ablation study results.} Comparison of our proposed method (\tsc{mSFT}) against two naïve alternative heterogeneous early-stopping algorithms averaged across six underlying models.}
\label{tab:ablation}
\end{table}

\subsection{Further Analysis}\label{sec:further}

To rigorously evaluate the practical utility of \tsc{mSFT}, we conduct additional analyses. Unless otherwise stated, we use \ttt{Qwen2.5 3B} and primarily benchmark against standard SFT, the most widely adopted paradigm, and IES, which emerged as the strongest baseline in \S\ \ref{sec:exp_main_result}.

\paragraph{(I) \tsc{mSFT} Gains are Robust Across Dataset Scales.}
We find that the performance gains of \tsc{mSFT} remain robust across varying dataset sizes and task counts ($N \in \{5, 10, 15\}$) indicating that \tsc{mSFT} is valuable across a wide range of real-world scenarios. Across all three configurations, \tsc{mSFT} consistently outperforms SFT, yielding an average improvement of +5.4\% (see Fig. \ref{fig:diverse_N}).

\paragraph{(II) \tsc{mSFT} is Insensitive to Compute Budget $C$, with Simultaneous FLOPs Savings and Performance Gains.}
We demonstrate that under restricted compute budget, \tsc{mSFT} improves downstream performance while simultaneously reducing FLOPs. When $C = 1$, we observe a +3.4\% performance gain alongside an average compute reduction of 120.3 PFLOPs (see Fig. \ref{fig:granular_flops}). This efficiency is achieved because \tsc{mSFT} introduces no additional roll-out overhead compared to SFT, while dynamically excluding sub-datasets during training to save compute. Notably, these performance gains do not degrade as the budget $C$ decreases. Refer to Appendix \ref{app:FLOPS} for details on how FLOPs are measured across all methods.

\begin{figure*}[tbp]
    \centering
    \includegraphics[width=1\linewidth]{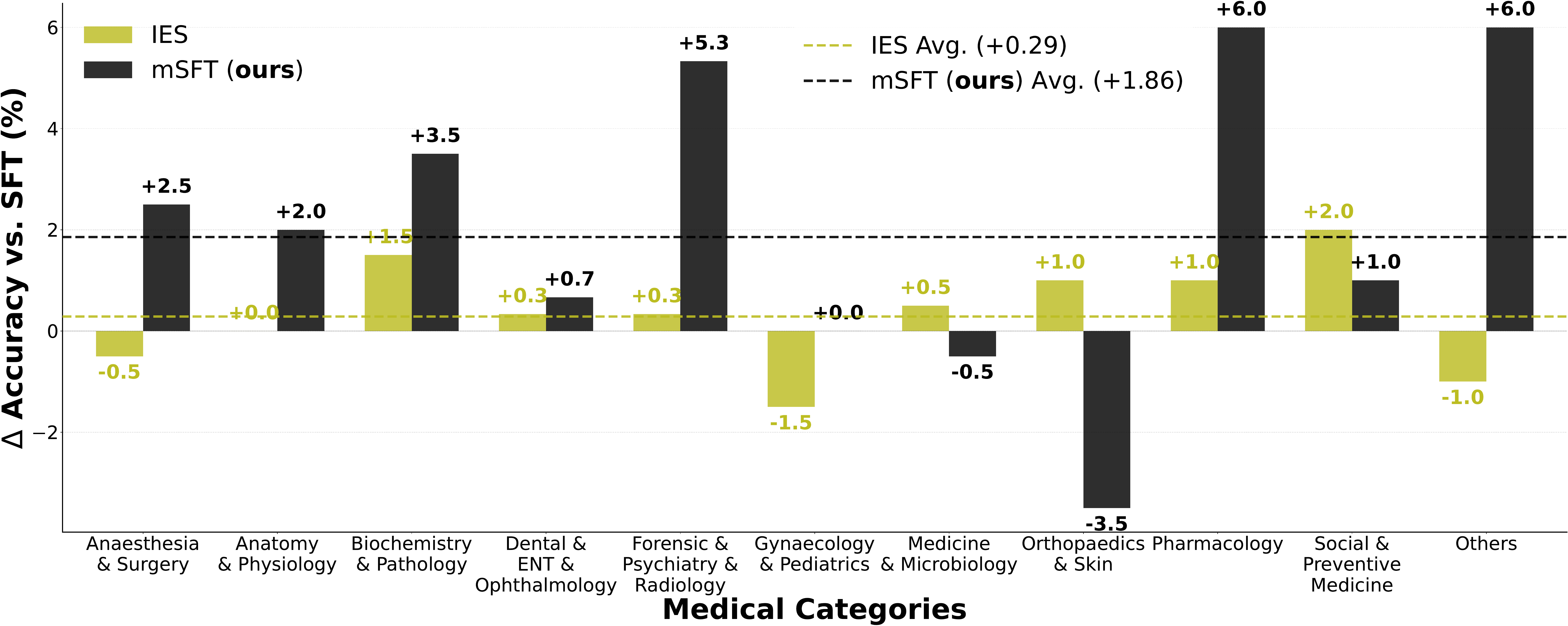}
    \caption{\textbf{Performance on further granular decompositions.} Evaluating \tsc{mSFT} across MedMCQA sub-categories using \ttt{Qwen2.5 3B} demonstrates an average accuracy improvement of +1.86\% over the SFT baseline, outperforming IES (+0.29\%).}
    \label{fig:med}
\end{figure*}

\paragraph{(III) \tsc{mSFT} Remains Effective on Granular Decompositions.}
We further investigate whether \tsc{mSFT} remains effective at a highly granular level by applying it to the 21 pre-defined sub-categories of the MedMCQA dataset \citep{pmlr-v174-pal22a}. As shown in Fig. \ref{fig:med} (grouped into 11 broad categories for legibility), \tsc{mSFT} yields an average accuracy improvement of +1.86\% over SFT, outperforming IES (+0.29\%). We observe particularly pronounced gains in specialized domains such as Pharmacology (+6.0\%) and Forensic, Psychiatry \& Radiology (+5.3\%). Despite topic-specific variance, \tsc{mSFT} consistently improves performance across most sub-categories, validating its efficacy on fine-grained task distributions.

\begin{figure}[tbp]
    \centering
    \includegraphics[width=\linewidth]{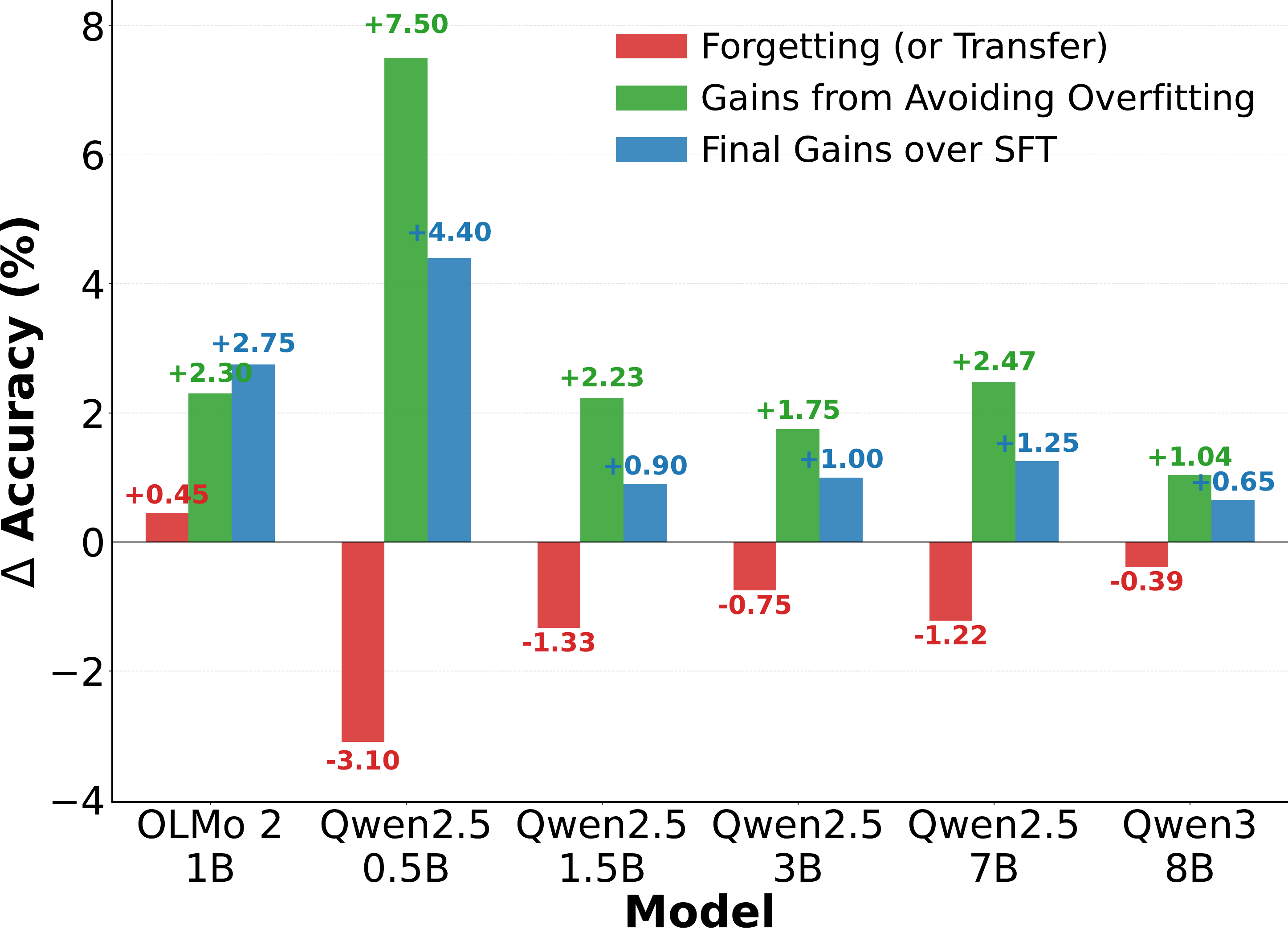}
    \caption{\textbf{Decomposition of performance gains.} \tsc{mSFT}'s accuracy improvement over SFT is decomposed into overfitting prevention benefits and dataset exclusion effects. Minor catastrophic forgetting from hard exclusion is outweighed by gains from mitigating heterogeneous overfitting.}
    \label{fig:decomp}
\end{figure}

\paragraph{(IV) Decomposing Overfitting Prevention and Catastrophic Forgetting.} To better understand the trade-off between preventing overfitting and the risk of catastrophic forgetting, we decompose \tsc{mSFT}'s performance gains relative to SFT (Fig. \ref{fig:decomp}). Specifically, we quantify the effect of dataset exclusion as:
\begin{align}
    &\text{Forgetting (or Transfer)} \notag \\ 
    &\quad:= \text{Metric}(c_{\text{final}}) - \text{Metric}(c_{\text{min}}),
    \label{eq:forget}
\end{align}
where $c_{\text{final}}$ denotes the globally optimal checkpoint and $c_{\text{min}}$ represents the peak performance checkpoint identified during the roll-out search (Alg. \ref{alg:msft}, line 5). 

A negative Eq. \ref{eq:forget} indicates forgetting from hard exclusions, which is the most common empirical outcome. Conversely, a positive value, as occasionally observed, suggests that continued training on the remaining mixture induces positive transfer. By subtracting Eq. \ref{eq:forget} from the overall performance gain over standard SFT, we isolate the benefit of overfitting prevention. Ultimately, our analysis reveals that while hard exclusion incurs minor forgetting penalties on average, the performance gains achieved by mitigating heterogeneous overfitting outweigh these losses, driving the overall superiority of \tsc{mSFT}.

\begin{figure}[tb]
    \centering
    \includegraphics[width=\linewidth]{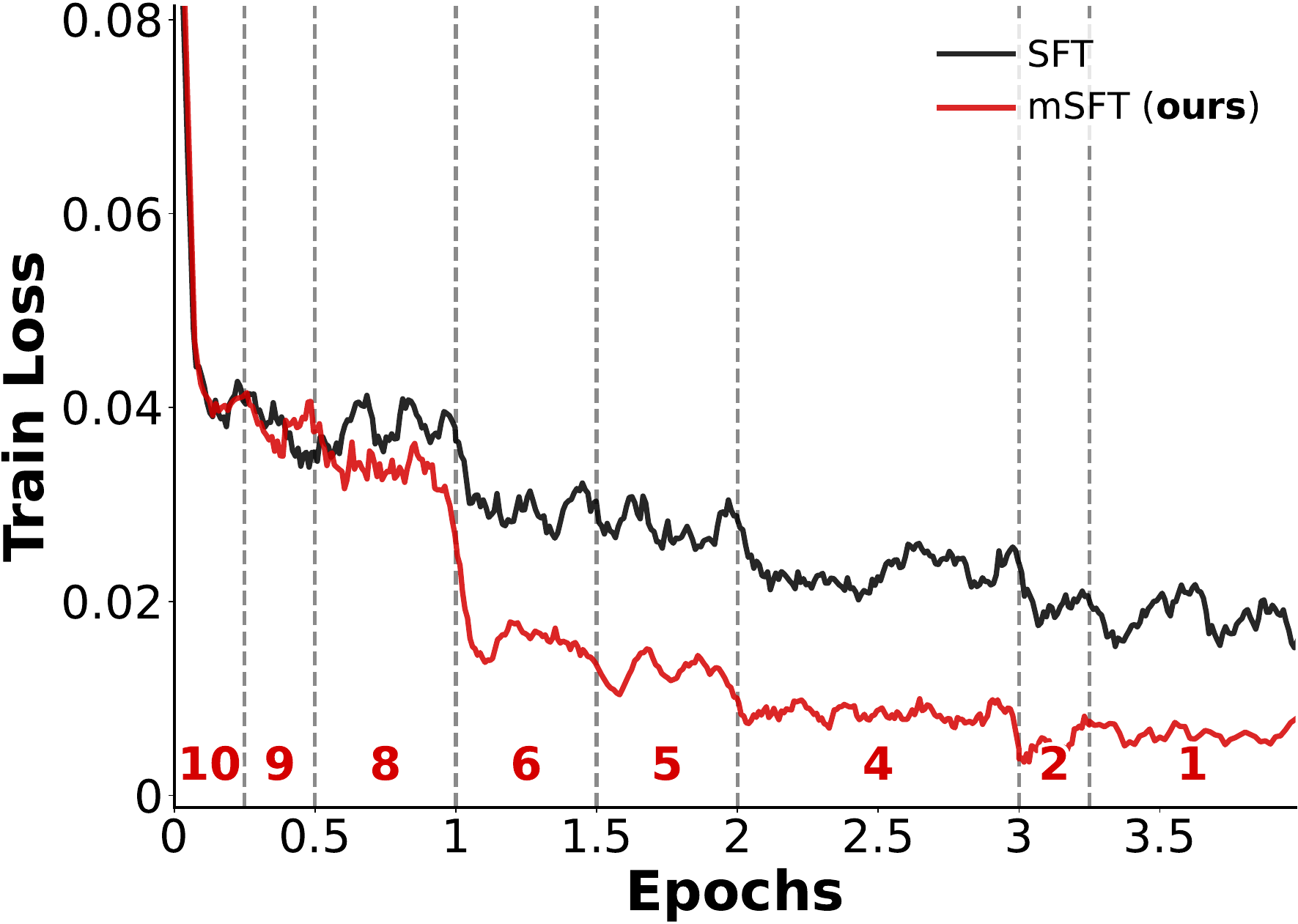}
    \caption{\textbf{Training loss curve comparison at \ttt{8B}.} Smoothed with moving average with sliding window 10. Dashed vertical lines denote roll-back where a sub-dataset is excluded. Numerical annotation at the bottom indicate the number of remaining sub-datasets at each interval.}
    \label{fig:loss-qwen3-8b}
\end{figure}

\paragraph{(V) \tsc{mSFT} Commonly Embodies Lower Training Loss.}
As seen in Fig. \ref{fig:loss-qwen3-8b} (and Appendix \ref{app:loss}), \tsc{mSFT} commonly achieves a consistently lower training loss than standard SFT. With base model \ttt{Qwen3 8B}, the curve occasionally exhibits sharp, step-wise loss descents immediately after overfitted sub-datasets are excluded. We hypothesize this reflects a relief from gradient conflict. In SFT, simultaneous updates can cause progress on some tasks to actively disrupt others. Furthermore, once a fast-learning dataset passes its optimal compute point, it likely introduces noisy, over-specialized gradients. By dynamically filtering out these post-peak datasets, \tsc{mSFT} unburdens the optimizer, enabling the model to reallocate its capacity and more efficiently minimize the loss of the remaining, slower-learning tasks.

\paragraph{(VI) \tsc{mSFT} Generalizes to Other Model Families.}
To verify that \tsc{mSFT} is not specific to the \ttt{OLMo} and \ttt{Qwen} families, we additionally evaluate \ttt{Gemma 3 4B} \citep{gemmateam2025gemma3technicalreport}. As shown in Tab. \ref{tab:gemma}, \tsc{mSFT} improves over SFT by +3.7\% on average, with gains in all three task categories.

\paragraph{(VII) \tsc{mSFT} Outperforms Loss-Reweighting.}
We further compare against CoBa \citep{gong2024coba}, which dynamically reweights task losses to balance convergence rates (see \S\ \ref{sec:discussion}). As shown in Tab. \ref{tab:coba}, while CoBa improves over SFT (+1.2\%), \tsc{mSFT} outperforms CoBa on all six models (+1.8\%), without per-step gradient-level interventions or additional sensitive hyperparameters.

\begin{table*}[tbp]
\centering
\small
\begin{tabularx}{\textwidth}{l RRR R}
\toprule
\textbf{Model:} & \multicolumn{4}{c}{\textbf{\ttt{Gemma 3 4B}}} \\
\cmidrule(lr){2-5}
& \textbf{Science and Knowledge} & \textbf{Commonsense and Language} & \textbf{Mathematic and Quantitative} &
\cellcolor{gray!15}\textbf{Average} \\
\midrule
SFT & 59.2 & 69.0 & 38.5 & \cellcolor{gray!15}59.0 \\
\tsc{mSFT} (\textbf{ours}) & \textbf{63.0}$_{\color{mygreen}+3.8}$ & \textbf{74.1}$_{\color{mygreen}+5.1}$ &
\textbf{39.5}$_{\color{mygreen}+1.0}$ & \cellcolor{gray!15}\textbf{62.7}$_{\color{mygreen}+3.7}$ \\
\bottomrule
\end{tabularx}
\caption{\textbf{Results on a different model family.} \tsc{mSFT} ($C=3$) on \ttt{Gemma 3 4B}. Average denotes the average
accuracy across 10 benchmarks. Subscripts denote $\Delta$ relative to SFT.}
\label{tab:gemma}
\end{table*}

\begin{table*}[tbp]
\centering
\small
\begin{tabularx}{\textwidth}{l RRRRRR R}
\toprule
\textbf{Model:} & \textbf{\ttt{OLMo 2}} & \multicolumn{4}{c}{\textbf{\ttt{Qwen2.5}}} & \textbf{\ttt{Qwen3}} & \\
\cmidrule(lr){2-2} \cmidrule(lr){3-6} \cmidrule(lr){7-7}
\textbf{Size:} & \textbf{\ttt{1B}} & \textbf{\ttt{0.5B}} & \textbf{\ttt{1.5B}} & \textbf{\ttt{3B}} & \textbf{\ttt{7B}} &
\textbf{\ttt{8B}} & \cellcolor{gray!15}\textbf{Average} \\
\midrule
\multicolumn{8}{c}{\cellcolor{myorange!15}\textbf{Average Accuracy Across 10 Benchmarks}} \\
\midrule
SFT & 43.6 & 33.0 & 64.1 & 73.2 & 76.8 & 80.8 & \cellcolor{gray!15}61.9 \\
CoBa & 46.1 & 36.8 & 64.6 & 72.5 & 77.4 & 80.9 &
\cellcolor{gray!15}63.1$_{\color{mygreen}+1.2}$ \\
\tsc{mSFT} (\textbf{ours}) & \textbf{46.3} & \textbf{37.4} & \textbf{65.0} & \textbf{74.2} & \textbf{78.0} & \textbf{81.4} &
\cellcolor{gray!15}\textbf{63.7}$_{\color{mygreen}+1.8}$ \\
\bottomrule
\end{tabularx}
\caption{\textbf{Comparison with loss-reweighting.} \tsc{mSFT} against CoBa across six underlying
models. Subscripts denote $\Delta$ relative to SFT. The best scores are \textbf{bolded}.}
\label{tab:coba}
\end{table*}

\section{Discussion}\label{sec:discussion}

\paragraph{Additional Related Work.} Numerous works explore which datasets to include in the SFT stage \citep{dong-etal-2024-abilities, li-etal-2024-quantity}, and the optimal mixture ratios \citep{xiao2024sftmix,zhu-etal-2025-dynamic,shi2025damodatamixingoptimizer,wang2026hbo, li2025data}. Another line of research addresses task imbalance through continuous loss-reweighting or gradient manipulation, primarily studied in computer vision, reinforcement learning, and early LM multi-tasking \citep{chen2018gradnorm, yu2020gradient, liu2021conflict, liu2023famo, gong2024coba}. While  \cite{gong2024coba} dynamically adjust task weights to balance convergence rates, they require continuous gradient-level interventions during the forward-backward pass and introduce multiple sensitive hyperparameters (e.g., history windows, warm-up steps, temperature). In contrast, \tsc{mSFT} operates strictly at the data-scheduling level and hard exclusions, entirely avoiding this per-step computational overhead. Empirically, \tsc{mSFT} also outperforms CoBa across all six models (\S\ \ref{sec:further}, Tab. \ref{tab:coba}).

\newpage
\section{Limitation}
\paragraph{Efficient Disk Management.} An operational limitation of \tsc{mSFT} is the additional storage overhead incurred by saving intermediate checkpoints during the roll-out phase. To mitigate this, we introduce a dynamic checkpoint pruning algorithm in Appendix \ref{app:efficient_disk} that actively discards redundant model states. Empirically, this strategy results in average storage footprint by approximately 4.44$\times$ SFT (see Appendix \ref{app:storage}). Because disk space is rarely the primary bottleneck in large-scale LM training, especially given the negligible cost of storage relative to compute, we consider this an acceptable trade-off. Nevertheless, future work could further optimize this process to reduce disk overhead entirely.

\paragraph{Experiments on Limited Compute.} 
Our experiments are conducted under academic compute constraints, and as such, our dataset sizes (9K--27K samples) and model scales (0.5B--8B parameters) do not reach the frontier-lab scale. Nevertheless, our empirical study provides substantial evidence for the practical utility of \tsc{mSFT}. We evaluate across \textbf{6} models spanning three families (\ttt{OLMo 2}, \ttt{Qwen2.5}, \ttt{Qwen3}) and a wide range of scales (0.5B, 1B, 1.5B, 3B, 7B, 8B), and further validate robustness across heterogeneous dataset sizes ($N \in \{5, 10, 15\}$ tasks with 9K--27K samples; \S\ \ref{sec:further}, Fig. \ref{fig:diverse_N}) and granular task decompositions (\S\ \ref{sec:further}, Fig. \ref{fig:med}). We further confirm that the gains extend beyond these three families with \ttt{Gemma 3 4B} (\S\ \ref{sec:further}, Tab. \ref{tab:gemma}). The robustness of \tsc{mSFT} across these varied conditions suggests the underlying phenomenon generalizes beyond our experimental scale. We leave frontier-scale validation to future work.

\section*{Acknowledgments}
This work was supported by the Institute of Information \& communications Technology Planning \& Evaluation (IITP) grant funded by the Korea government (MSIT) (No. 2022-0-00871, Development of AI Autonomy and Knowledge Enhancement for AI Agent Collaboration), No. RS-2024-00457882, AI Research Hub Project, and No. RS-2019-II190075, Artificial Intelligence Graduate School Program (KAIST)).

\clearpage

\bibliography{custom}

\newpage
\appendix

\begin{table*}[b]
\centering
\setlength{\tabcolsep}{5pt}
\begin{tabular}{lcccccc}
\toprule
 & \multicolumn{3}{c}{Source} & \multicolumn{3}{c}{FLOPs} \\
\cmidrule(lr){2-4}\cmidrule(lr){5-7}
Stage & 7B & 13B & 32B & 7B & 13B & 32B \\
\midrule
Pre-training  & \multicolumn{3}{c}{paper §2.3}
  & $1.64{\times}10^{23}$
  & $3.90{\times}10^{23}$
  & $1.16{\times}10^{24}$ \\
Mid-training  & \multicolumn{3}{c}{paper §4.5}
  & $6.30{\times}10^{21}$
  & $4.68{\times}10^{22}$
  & $1.04{\times}10^{23}$ \\
SFT           & \multicolumn{3}{c}{HF dataset}
  & $2.97{\times}10^{19}$& $5.52{\times}10^{19}$& $1.24{\times}10^{20}$\\
DPO           & \multicolumn{3}{c}{HF dataset}
  & $1.89{\times}10^{19}$& $3.61{\times}10^{19}$& $1.28{\times}10^{20}$\\
RLVR-grad     & \multicolumn{3}{c}{\texttt{tulu3.md} script}
  & $7.15{\times}10^{19}$& $1.33{\times}10^{20}$& $3.27{\times}10^{20}$\\
RLVR-roll     & \multicolumn{3}{c}{\texttt{tulu3.md} script}
  & $7.63{\times}10^{20}$& $1.42{\times}10^{21}$& $3.49{\times}10^{21}$\\
\midrule
Post total    & \multicolumn{3}{c}{---}
  & $8.83{\times}10^{20}$& $1.64{\times}10^{21}$& $4.07{\times}10^{21}$\\
Post / Total  & \multicolumn{3}{c}{---}
  & $0.517\%$
  & $0.374\%$
  & $0.320\%$\\
SFT / Post    & \multicolumn{3}{c}{---}
  & $3.36\%$& $3.36\%$& $3.07\%$\\
\bottomrule
\end{tabular}
\caption{\textbf{\ttt{OLMo~2} training FLOPs by stage.} ``Post'' denotes the
sum of SFT, DPO, RLVR-grad, and RLVR-roll. Post/Total is the ratio of total
post-training FLOPs to total training FLOPs. SFT/Post is the fraction of
post-training compute spent on SFT.}
\label{tab:olmo2_flops}
\end{table*}

\section{Computation of FLOPs Proportion} \label{app:train_proportion}

The \ttt{OLMo~2} technical paper \citep{walsh2025} reports total FLOPs computed via the standard formula from \citet{kaplan2020scalinglawsneurallanguage}. We adopt the same formula and extend it to each training stage to compute proportional contributions. We use the reported parameter size ($|\theta| \in \{7\text{B}, 13\text{B}, 32\text{B}\}$).

\paragraph{Pre-training and mid-training.} Pre-training token counts are taken from \citet{walsh2025}~§2.3. Mid-training tokens follow from the model souping procedure (§4.5): \ttt{7B} performs three annealing runs of 50B tokens each (150B total); \ttt{13B} performs three 100B runs plus one 300B run (600B total); \ttt{32B} is derived by subtracting pre-training from the overall base (pre- + mid-training) total ($6.60\text{T} - 6.06\text{T} = 0.54\text{T}$).

\paragraph{SFT.}
Data is from \path{allenai/tulu-3-sft-olmo-2-mixture} (7B, 13B; $n_{\text{sft}} = 939{,}334$) and \path{allenai/tulu-3-sft-olmo-2-mixture-0225} (32B; $n_{\text{sft}} = 866{,}138$). Per \path{docs/tulu3.md}, maximum sequence length is 4,096 tokens and training runs for 2 epochs:
\[
  \mathrm{FLOPs}_{\text{SFT}}
    = 6\,|\theta|\times n_{\text{sft}}\times \bar{l}_{\text{SFT}}\times 2,
\]
where $n_{\text{sft}}$ is the number of samples, and  $\bar{l}_{\text{SFT}}$ is the average token length per sample, capped at 4,096 and computed by streaming the full dataset with the \ttt{OLMo 2} tokenizer. 

\paragraph{DPO.}
Pair counts are from \path{allenai/olmo-2-1124\--7b-preference-mix} (366,700 pairs, 7B), \path{allenai/olmo-2-1124-13b-preference-mix} (377,700 pairs, 13B), and \path{allenai/olmo-2-0325-32b-preference-mix} (377,900 pairs, 32B). Per \path{docs/tulu3.md}, training uses 1 epoch and maximum sequence length is 2,048 tokens. Each pair is processed as two separate forward--backward passes:
\[
  \mathrm{FLOPs}_{\text{DPO}}
    = 6\,|\theta|\times n_{\text{pairs}}\times 2\bar{l}_{\text{DPO}},
\]
where $\bar{l}_{\text{DPO}}$ is the average token length across all chosen and rejected sequences pooled together, capped at 2,048.

\paragraph{RLVR.}

The 7B and 13B models use PPO; the 32B model uses GRPO. All sizes use 10M total episodes. For PPO (7B, 13B), rollouts are collected in batches of 32, giving $n_{\text{grad}} = 10\text{M}/32 = 312{,}500$ gradient update steps. For GRPO (32B), 16 completions are sampled per prompt, giving
$n_{\text{grad}} = 10\text{M}/16 = 625{,}000$ gradient update steps. Prompt and response are each capped at 2,048 tokens. 
FLOPs split into forward-only (RLVR-roll) and forward--backward (RLVR-grad) passes:
\begin{align*}
  \mathrm{FLOPs}_{\text{RLVR-roll}}
    &= 2\,|\Theta|\times 10\text{M}\times 4096\times 2, \\[4pt]
  \mathrm{FLOPs}_{\text{RLVR-grad}}
    &= 6\,|\Theta|\times n_{\text{grad}}\times 4096 \times
       {\begin{cases}
         2 ,\\
         1 ,
       \end{cases}}
\end{align*}
where the factor of 2 in $\mathrm{FLOPs}_{\text{RLVR-roll}}$ covers policy
rollout and the frozen reference model (one forward pass each per episode), and the factor of 2 in the PPO $\mathrm{FLOPs}_{\text{RLVR-grad}}$ term covers
the policy and value model gradients.
$\mathrm{FLOPs}_{\text{RLVR}} = \mathrm{FLOPs}_{\text{RLVR-roll}}
+ \mathrm{FLOPs}_{\text{RLVR-grad}}$.

\paragraph{Results.}
Tab.~\ref{tab:olmo2_flops} reports the resulting FLOPs per stage.

\section{Additional Figures for Heterogeneous Overfitting}\label{app:motivation_app}
\begin{figure*}[t]
    \centering
    \begin{subfigure}[b]{0.49\textwidth}
        \centering
        \includegraphics[width=\linewidth]{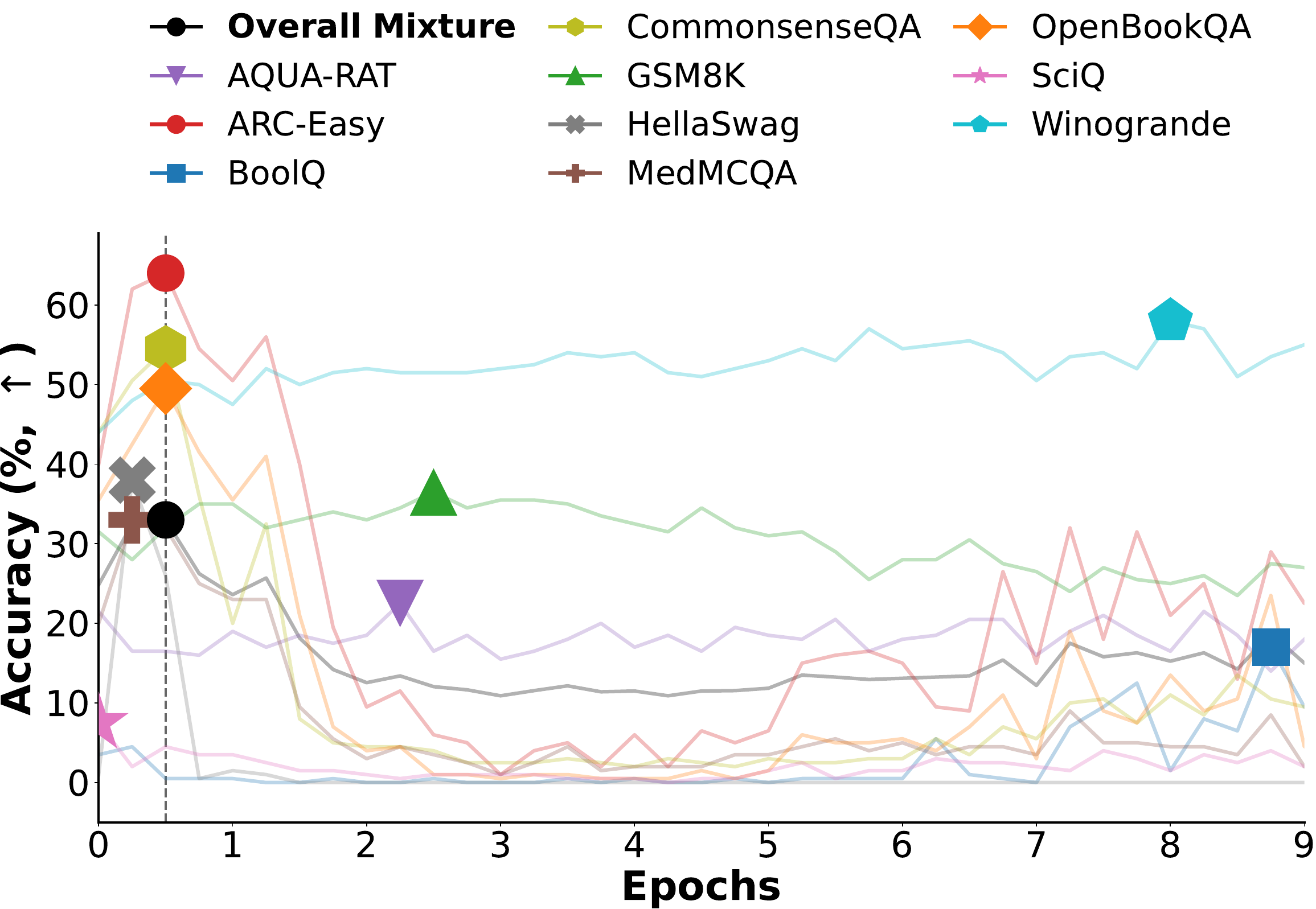}
        \caption{\ttt{Qwen2.5 0.5B}}
        \label{fig:app_qwen0_5b_a}
    \end{subfigure}
    \hfill
    \begin{subfigure}[b]{0.49\textwidth}
        \centering
        \includegraphics[width=\linewidth]{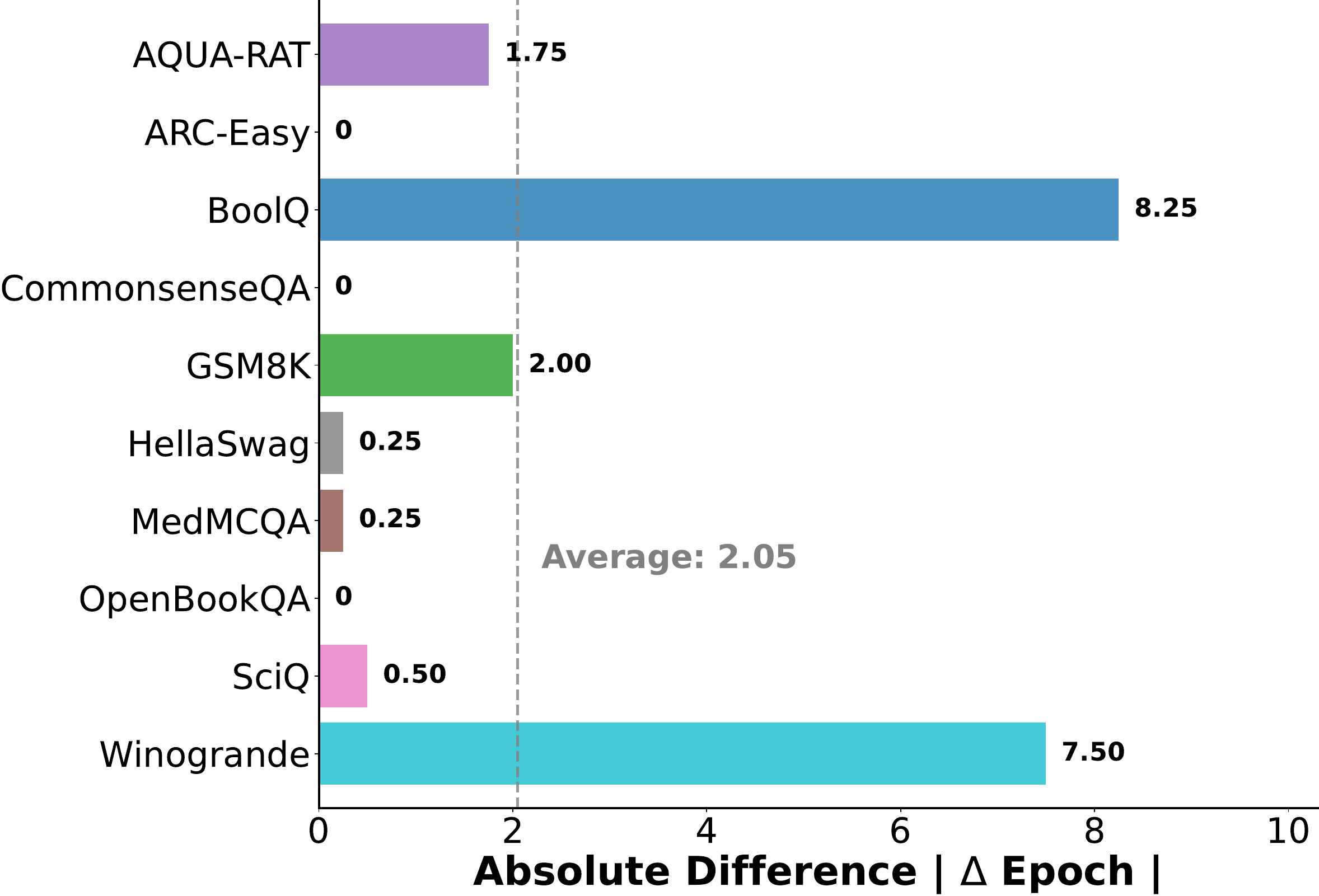}
        \caption{\ttt{Qwen2.5 0.5B}}
        \label{fig:app_qwen0_5b_b}
    \end{subfigure}

    \vspace{0.5em}

    \begin{subfigure}[b]{0.49\textwidth}
        \centering
        \includegraphics[width=\linewidth]{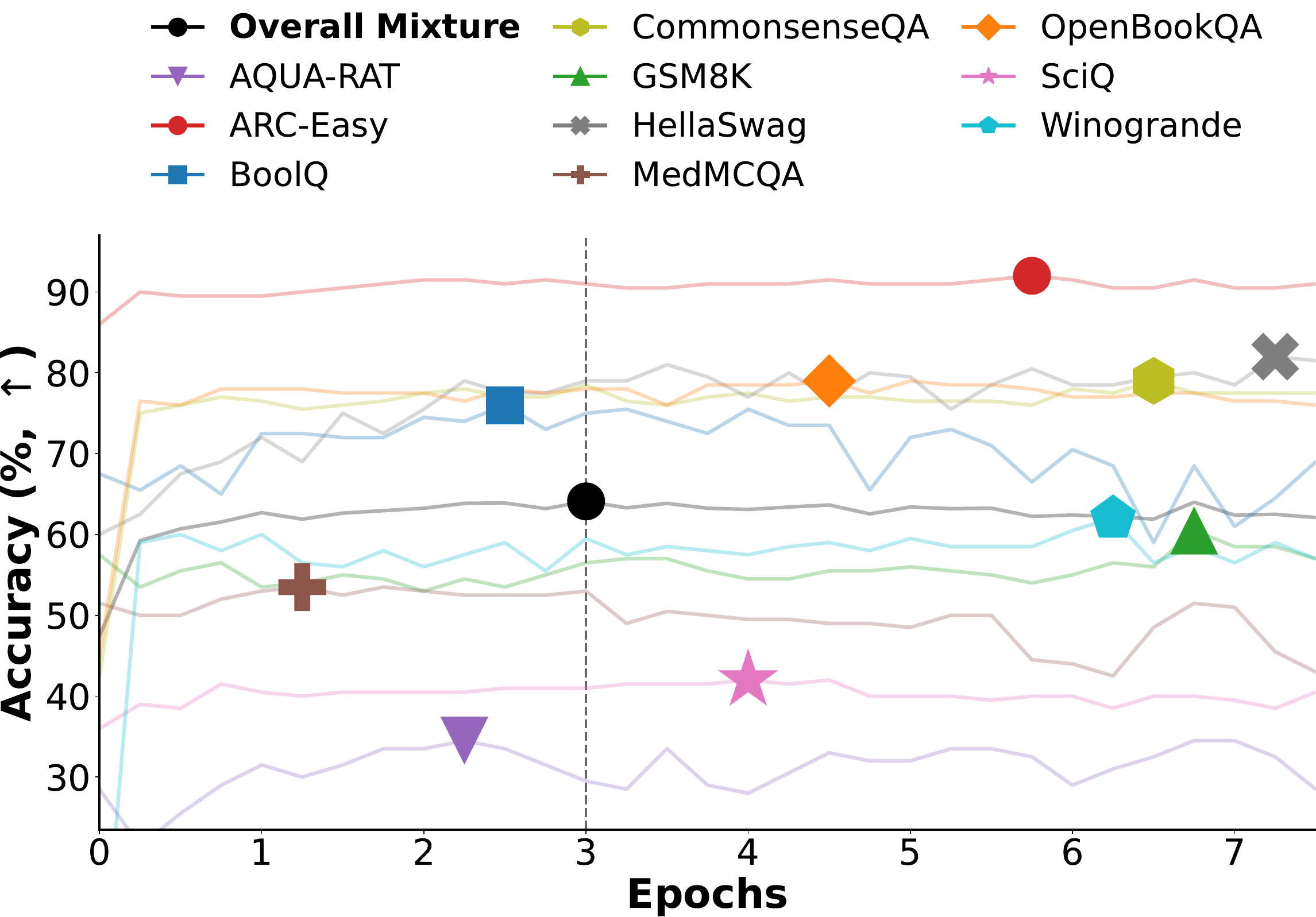}
        \caption{\ttt{Qwen2.5 1.5B}}
        \label{fig:app_qwen1_5b_a}
    \end{subfigure}
    \hfill
    \begin{subfigure}[b]{0.49\textwidth}
        \centering
        \includegraphics[width=\linewidth]{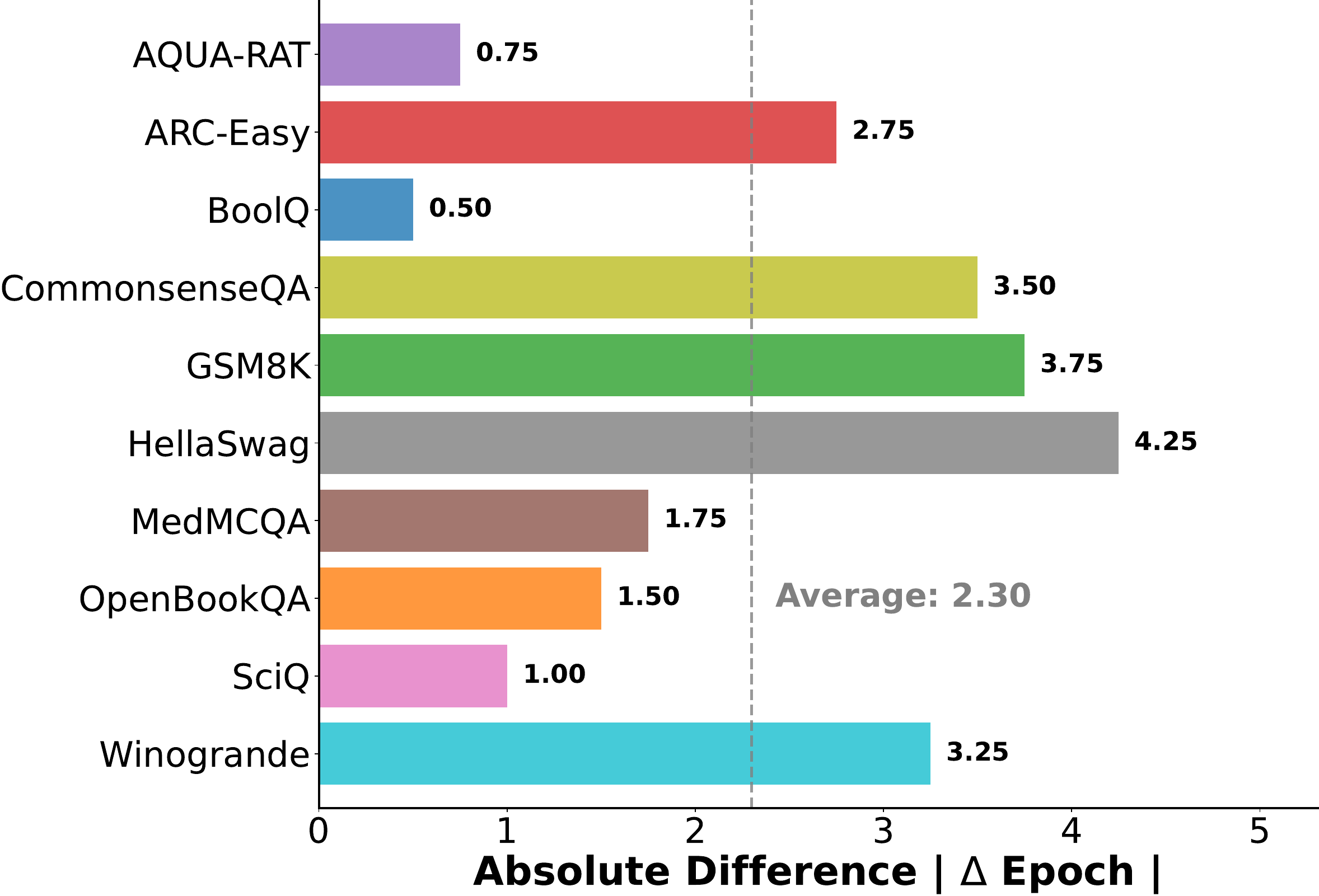}
        \caption{\ttt{Qwen2.5 1.5B}}
        \label{fig:app_qwen1_5b_b}
    \end{subfigure}

    \caption{\textbf{Heterogeneous learning dynamics.} Multi-task SFT demonstrates underlying sub-datasets overfitting dynamics vary greatly.}
    \label{fig:app_all_models_1}
\end{figure*}

\begin{figure*}[p]
    \centering
    \begin{subfigure}[b]{0.495\textwidth}
        \centering
        \includegraphics[width=\textwidth]{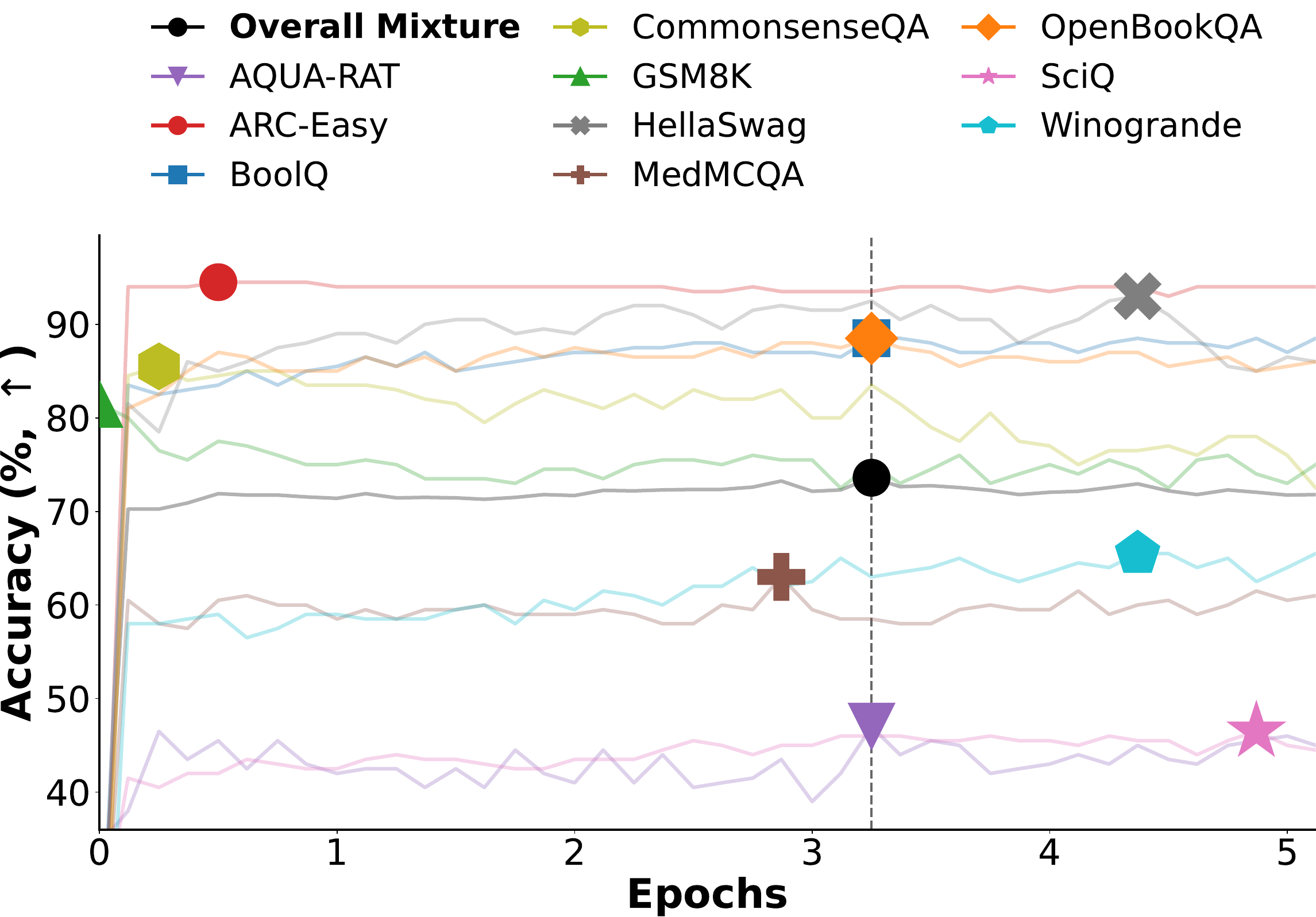}
        \caption{\ttt{Qwen2.5 3B}}
        \label{fig:app_qwen3b_a}
    \end{subfigure}
    \hfill
    \begin{subfigure}[b]{0.495\textwidth}
        \centering
        \includegraphics[width=\textwidth]{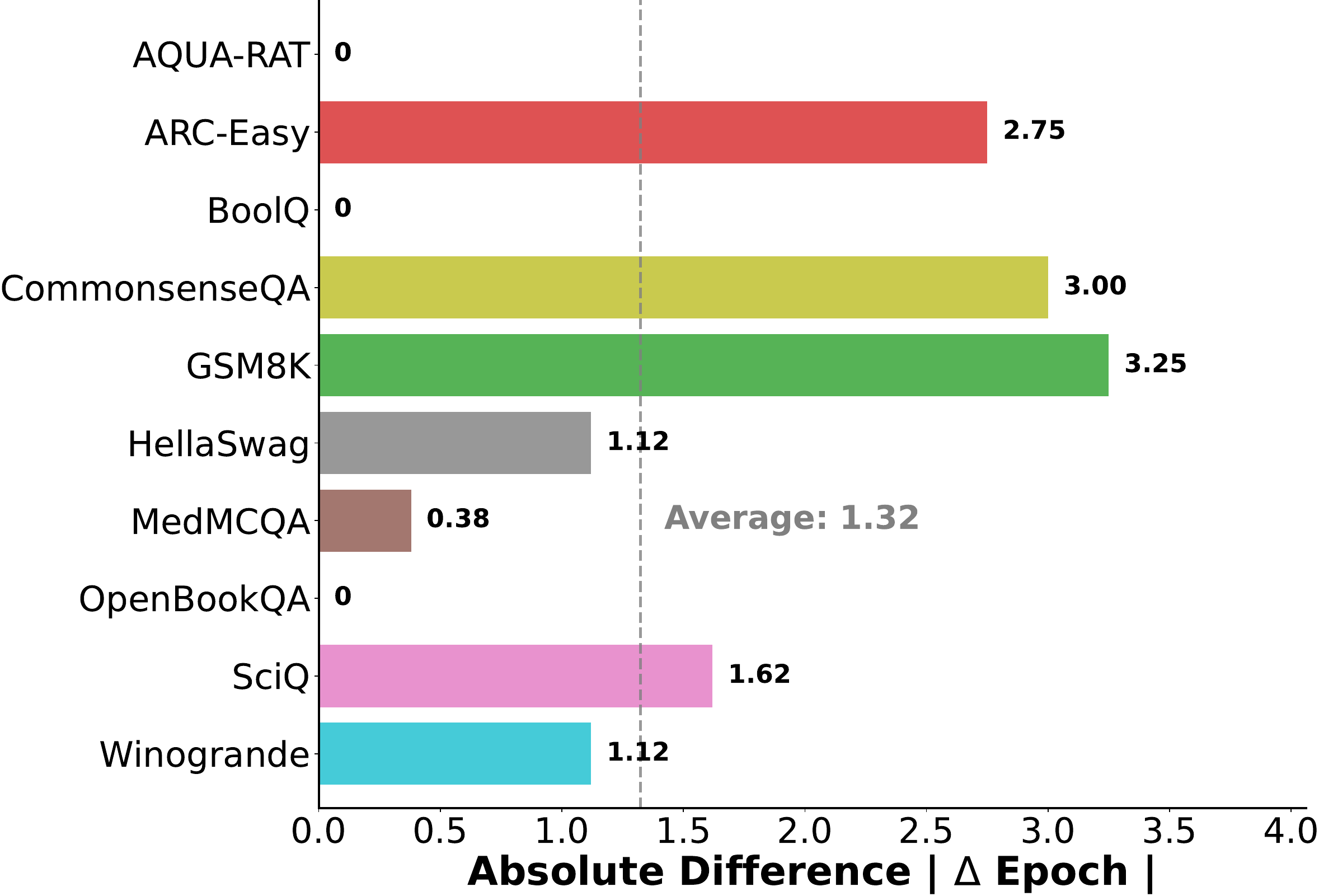}
        \caption{\ttt{Qwen2.5 3B}}
        \label{fig:app_qwen3b_b}
    \end{subfigure}
    
        \begin{subfigure}[b]{0.495\textwidth}
        \centering
        \includegraphics[width=\textwidth]{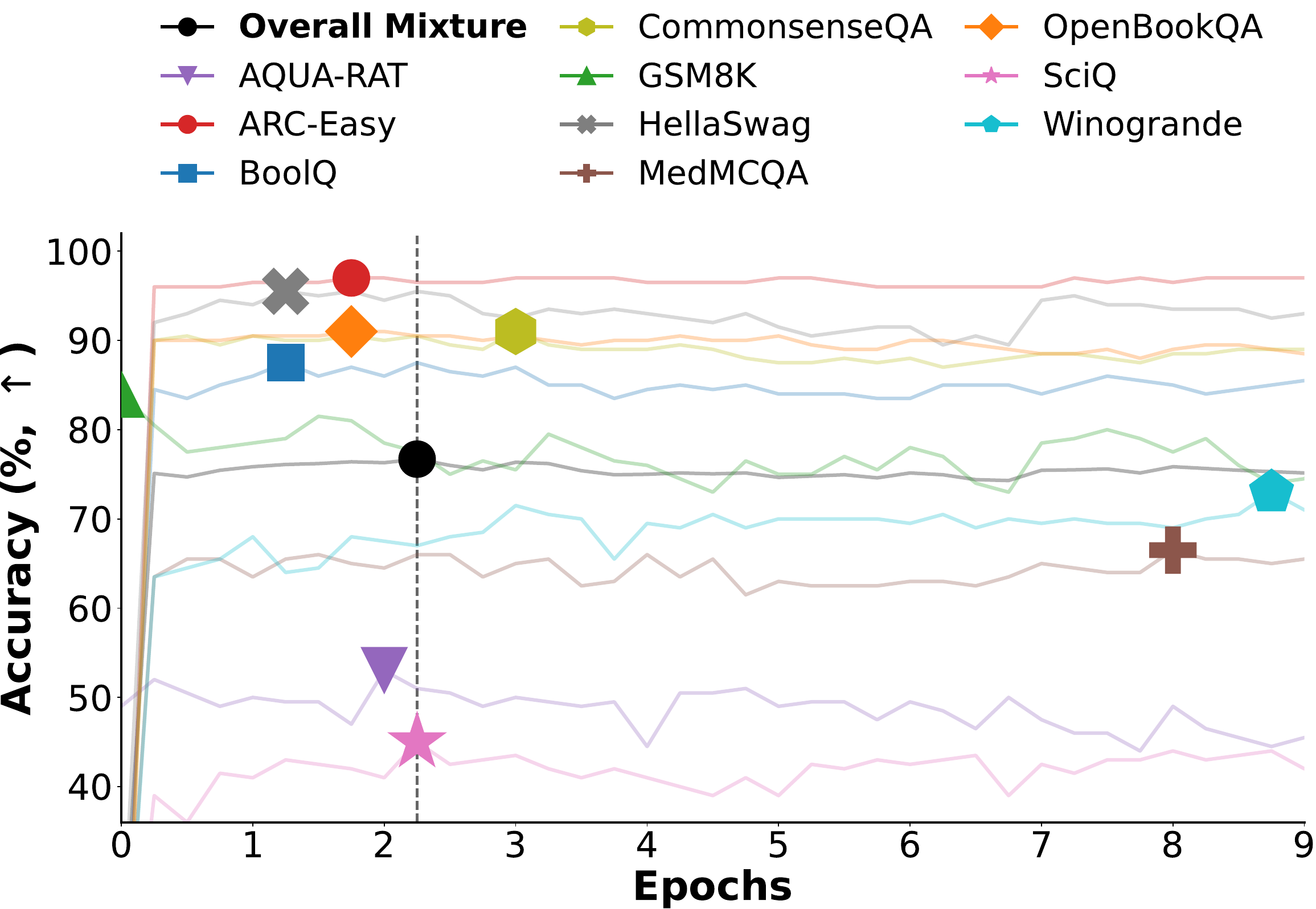}
        \caption{\ttt{Qwen2.5 7B}}
        \label{fig:app_qwen7b_a}
    \end{subfigure}
    \hfill
    \begin{subfigure}[b]{0.495\textwidth}
        \centering
        \includegraphics[width=\textwidth]{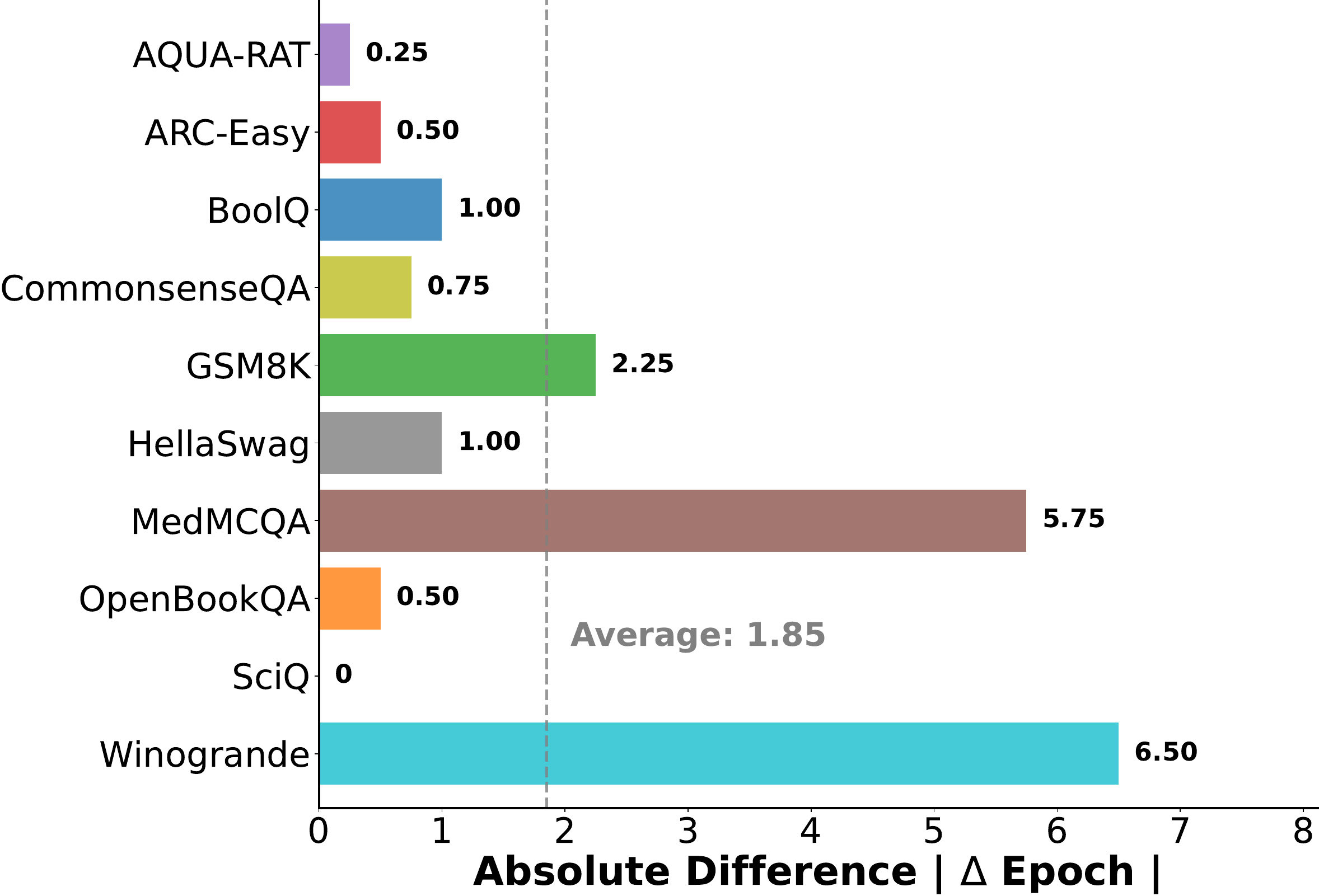}
        \caption{\ttt{Qwen2.5 7B}}
        \label{fig:app_qwen7b_b}
    \end{subfigure}
    
    \begin{subfigure}[b]{0.495\textwidth}
        \centering
        \includegraphics[width=\textwidth]{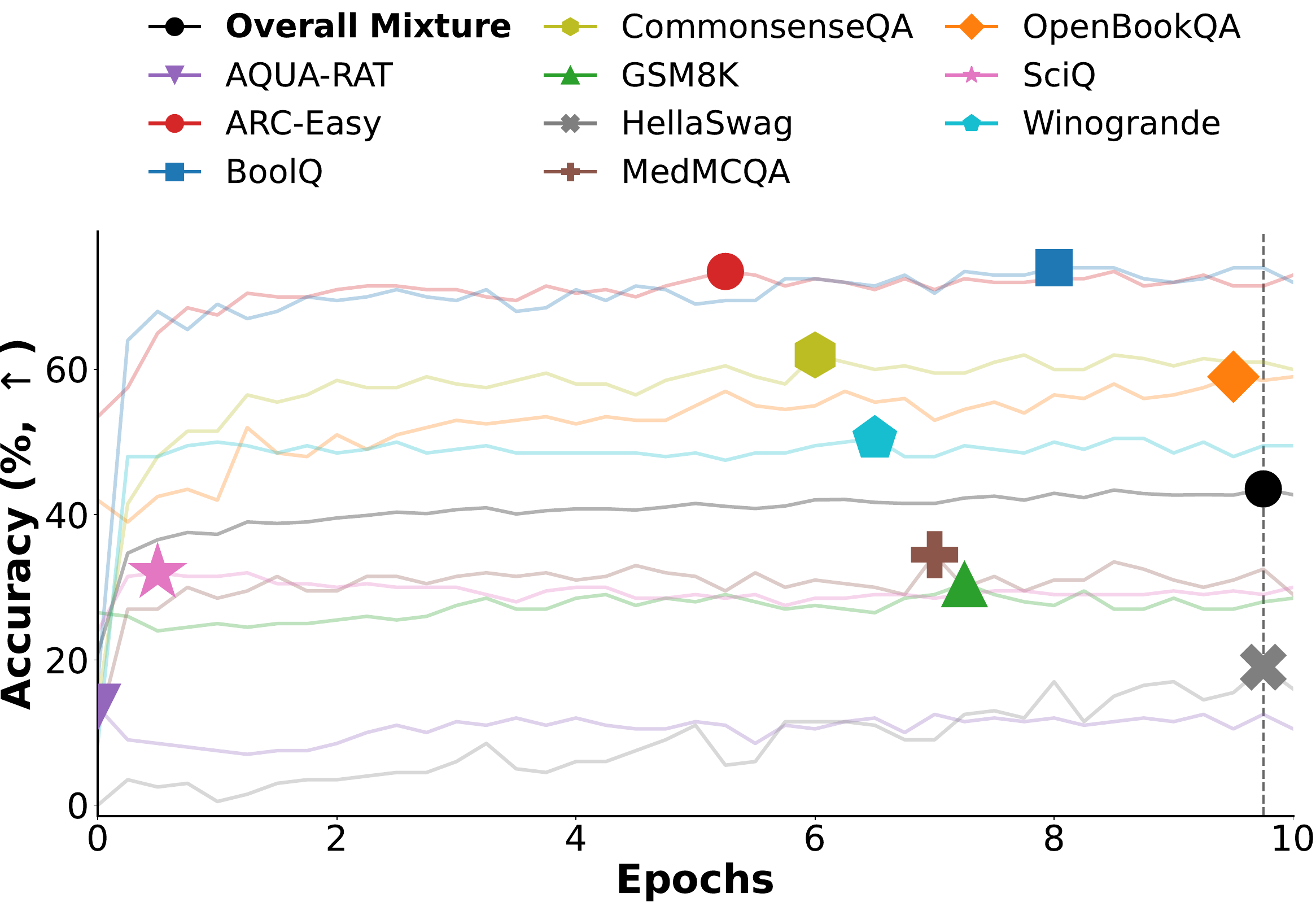}
        \caption{\ttt{OLMo2 1B}}
        \label{fig:app_olmo1b_a}
    \end{subfigure}
    \hfill
    \begin{subfigure}[b]{0.495\textwidth}
        \centering
        \includegraphics[width=\textwidth]{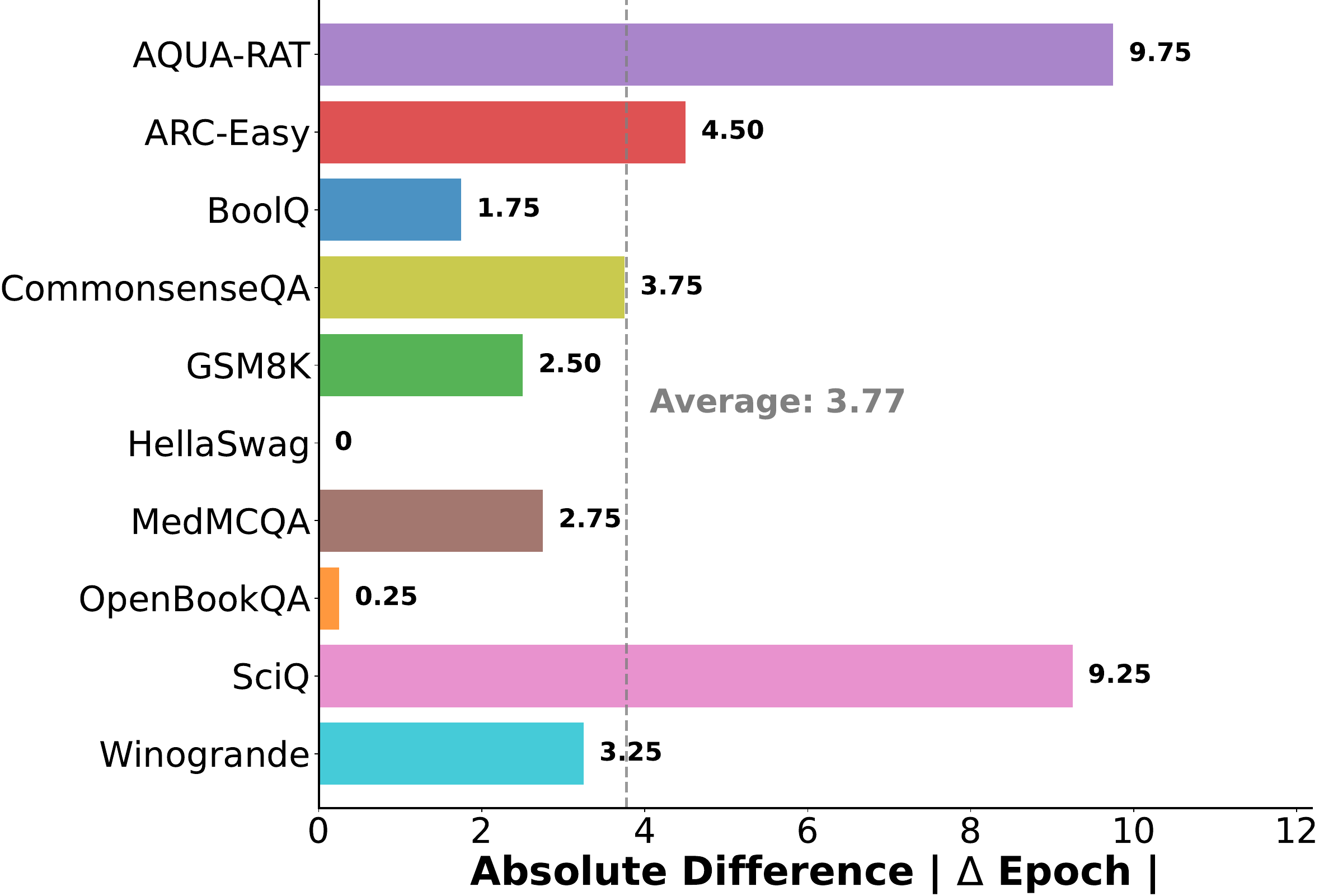}
        \caption{\ttt{OLMo2 1B}}
        \label{fig:app_olmo1b_b}
    \end{subfigure}
    \caption{\textbf{Heterogeneous learning dynamics.} Multi-task SFT demonstrates underlying sub-datasets overfitting dynamics vary greatly.}
    \label{fig:app_all_models_2}
\end{figure*}

Fig.~\ref{fig:app_all_models_1} and~\ref{fig:app_all_models_2} visualizes the per-sub-dataset validation accuracy for all remaining models. Across all models, each sub-dataset reaches its maximum accuracy at different training steps, confirming heterogeneous overfitting dynamics discussed in \S\ \ref{sec:motivation}.

\section{Further Details on SRO SFT and Soft SRO SFT}\label{app:baselines}

\paragraph{psuedocode for SRO and Soft SRO}
Alg. \ref{alg:sro} is the pseudocode for SRO, and Alg. \ref{alg:soft-sro} is the pseudocode for Soft SRO.

\begin{figure*}[t]
\centering
    \makeatletter
    \@twocolumnfalse
    \makeatother
    \begin{algorithm2e}[H]
    \caption{\tsc{SRO}}
    \label{alg:sro}
    \SetAlgoLined
    \LinesNumbered
    \Input{Dataset mixture $\mc{D}$, base model $\theta_0$, compute budget $C$}

    $\hat{\theta} \leftarrow \theta_0$ \tcp*{Initialization}

    \tcc{\textcolor{mygreen}{\textbf{Single roll-out search}}: Search for per-sub-dataset peaks}
    $\theta, \;\{\text{acc}(\mc{D}_i, c)\}_{i,c} \leftarrow \tsc{SFT-Roll-out}\!\left(\hat{\theta},\; \mc{D},\; C\right)$ \;
    $c_i^* \leftarrow \arg\max_{c}\; \text{acc}(\mc{D}_i, c)$ \tcp*{Optimal compute per sub-dataset}

    \tcc{\textcolor{myblue}{\textbf{Train from scratch}}: Start a new training run and exclude sub-datasets that have exhausted their budget}
    $\mc{E} \leftarrow \emptyset ;\; \hat{\theta} \leftarrow \theta_0 ;\; c_{\text{current}} \leftarrow 0$ \tcp*{Initialization}
    
    \While{$\mc{D} \setminus \mc{E} \neq \emptyset$}{
        \tcc{Find the next closest stopping point among active datasets}
        $c_{\text{next}} \leftarrow \min_{\mc{D}_i \in \mc{D} \setminus \mc{E}} c_i^*$ \;
        $\Delta c \leftarrow c_{\text{next}} - c_{\text{current}}$ \;
        
        \tcc{Roll-out active datasets for the delta compute and update model}
        $\hat{\theta}, \;\_ \leftarrow \tsc{SFT-Roll-out}\!\left(\hat{\theta},\; \mc{D} \setminus \mc{E},\; \Delta c\right)$ \;
        
        \tcc{Update current compute and exclude datasets that just peaked}
        $c_{\text{current}} \leftarrow c_{\text{next}}$ \;
        $\mc{E} \leftarrow \mc{E} \cup \{\mc{D}_i : c_i^* \le c_{\text{current}}\}$ \;
    }
\end{algorithm2e}

\begin{algorithm2e}[H]
    \caption{\tsc{Soft SRO}}
    \label{alg:soft-sro}
    \SetAlgoLined
    \LinesNumbered
    \Input{Dataset mixture $\mc{D}$, base model $\theta_0$, compute budget $C$}

    $\hat{\theta} \leftarrow \theta_0$ \tcp*{Initialization}

    \tcc{\textcolor{mygreen}{\textbf{Single roll-out search}}: Approximately search for per-sub-dataset peaks}
    $\theta, \;\{\text{acc}(\mc{D}_i, c)\}_{i,c} \leftarrow \tsc{SFT-Roll-out}\!\left(\hat{\theta},\; \mc{D},\; C\right)$ \;
    $c_i^* \leftarrow \arg\max_{c}\; \text{acc}(\mc{D}_i, c)$ \tcp*{Optimal compute per sub-dataset}

    \tcc{\textcolor{myblue}{\textbf{Train from scratch}}: Start a new training run with a new data mixture accounting for the optimal compute budgets}
    $\hat{\theta} \leftarrow \theta_0 ;\; \mc{D}' \leftarrow \emptyset ;\; Z \leftarrow \sum_{j} (c_j^* \cdot |\mc{D}_j|)$ \tcp*{Initialization and normalization factor}
    
    \For{$\mc{D}_i \in \mc{D}$}{
        $r \leftarrow (\sum_i|\mc{D}_i|) \cdot \frac{c_i^* \cdot |\mc{D}_i|}{Z}$ \tcp*{Target number of samples, preserving base proportions}
        $\mc{D}_i' \leftarrow \emptyset$ \;
        
        \While{$r \ge |\mc{D}_i|$}{
            \tcc{Add a full copy of $\mc{D}_i$ using multiset union}
            $\mc{D}_i' \leftarrow \mc{D}_i' \uplus \mc{D}_i$ \;
            $r \leftarrow r - |\mc{D}_i|$ \;
        }
        
        \If{$r > 0$}{
            $\tilde{\mc{D}}_{i} \leftarrow \text{Sample } \lfloor r \rfloor \text{ samples from } \mc{D}_i \text{ without replacement}$ \;
            $\mc{D}_i' \leftarrow \mc{D}_i' \uplus \tilde{\mc{D}}_{i}$ \;
        }
        $\mc{D}' \leftarrow \mc{D}' \uplus \mc{D}_i'$ \tcp*{Add the proportioned sub-dataset to the new mixture}
    }
    $\hat{\theta}, \;\_ \leftarrow \tsc{SFT-Roll-out}\!\left(\hat{\theta},\; \mc{D}',\; C\right)$ \;
\end{algorithm2e}

\end{figure*}

\clearpage
\section{Further Experimental Results on \texorpdfstring{$\Delta$}{Delta} Optimal Compute}\label{app:results_delta}

\begin{center}
\begin{tabular}{cc}
    \includegraphics[width=0.48\textwidth]{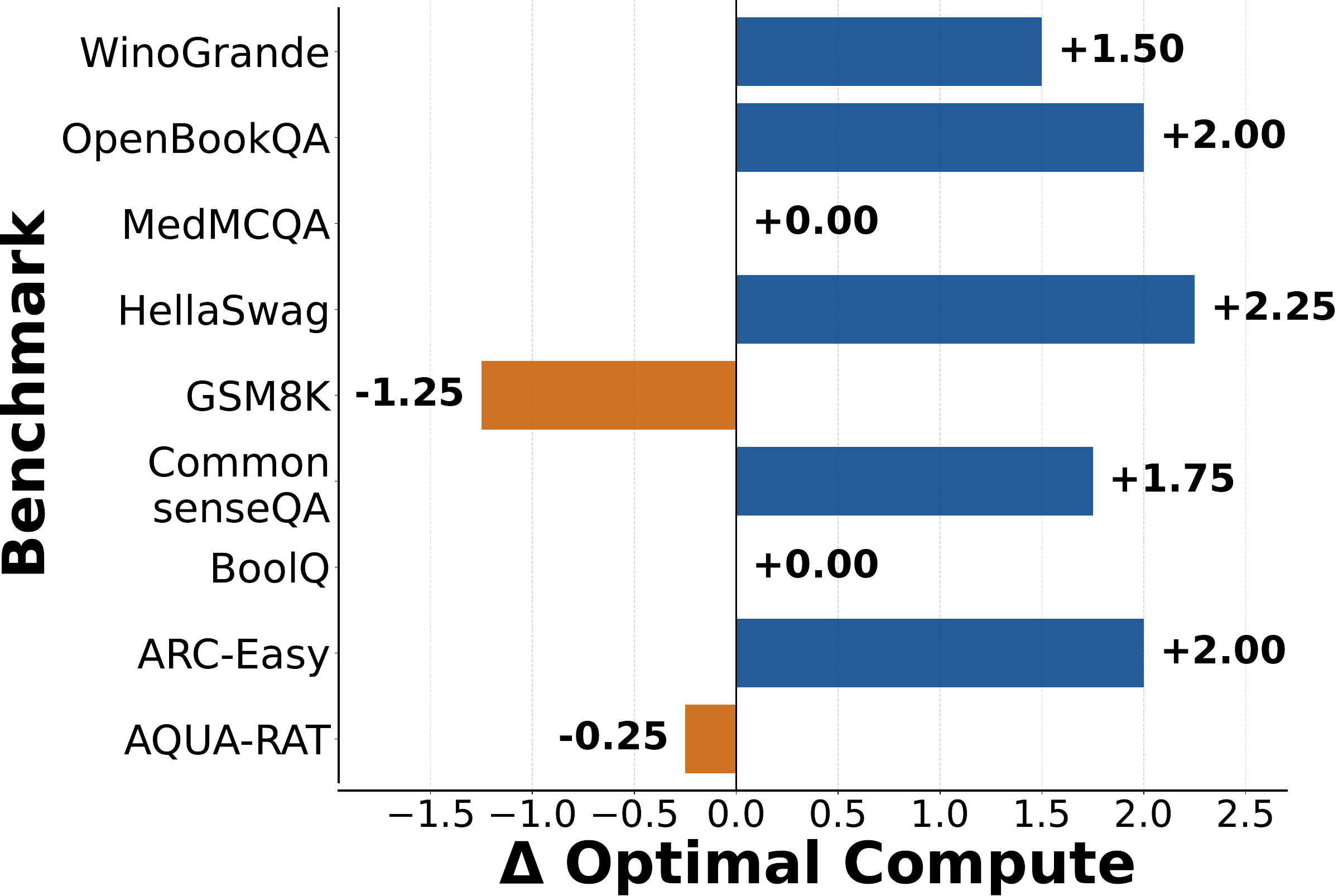} &
    \includegraphics[width=0.48\textwidth]{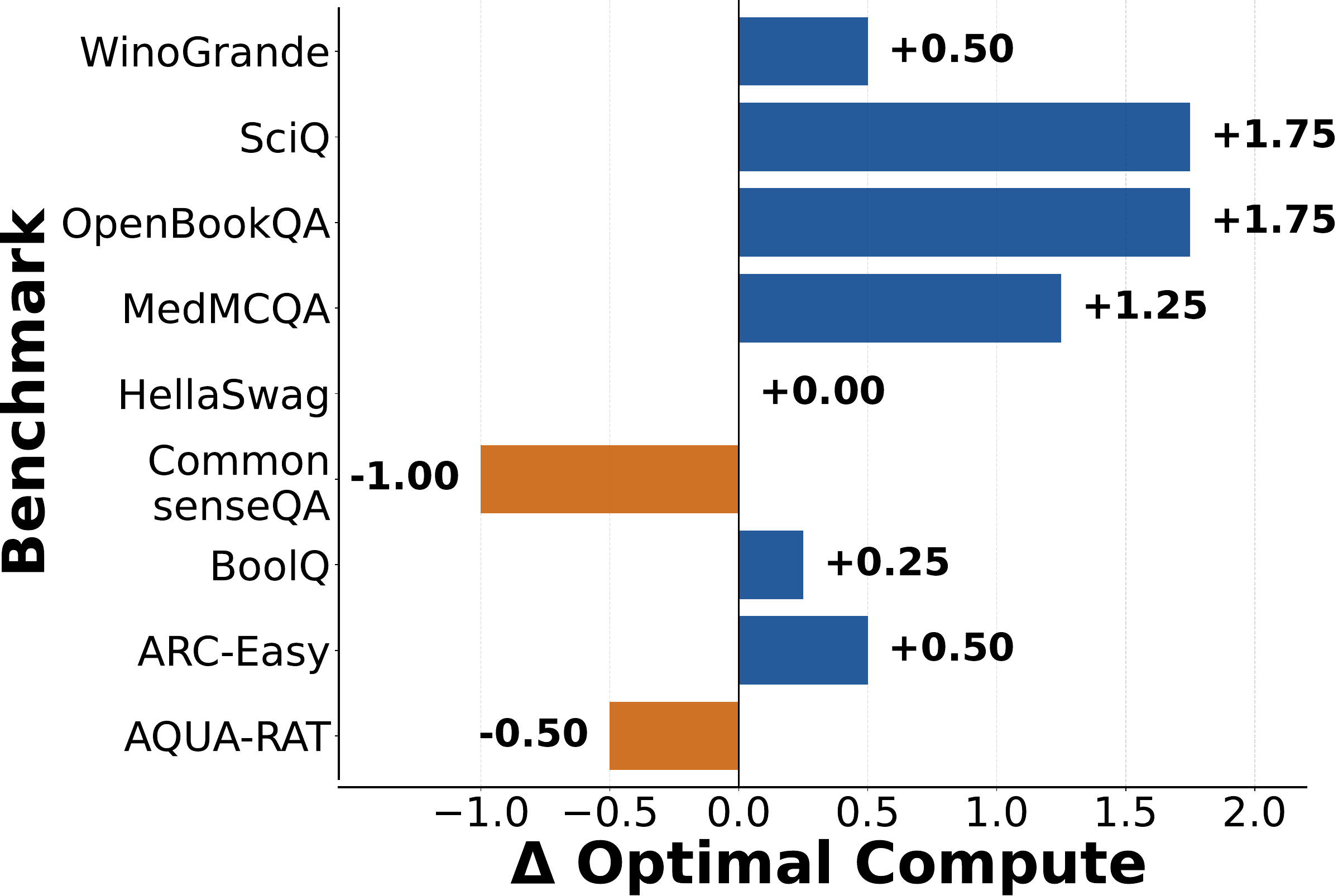} \\
    {\small (a) \ttt{Qwen2.5 0.5B}} &
    {\small (b) \ttt{Qwen2.5 1.5B}} \\[1em]
    \includegraphics[width=0.48\textwidth]{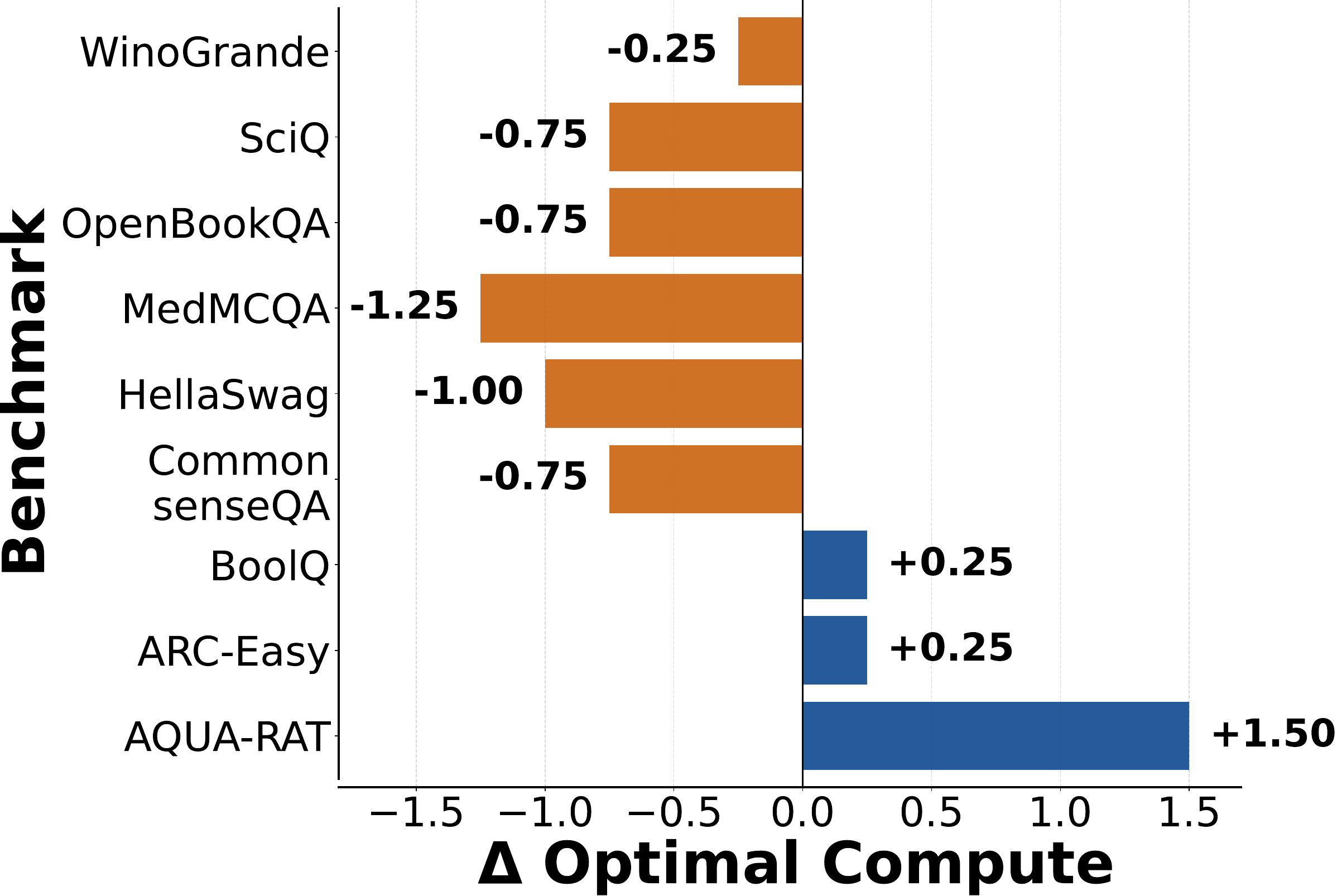} &
    \includegraphics[width=0.48\textwidth]{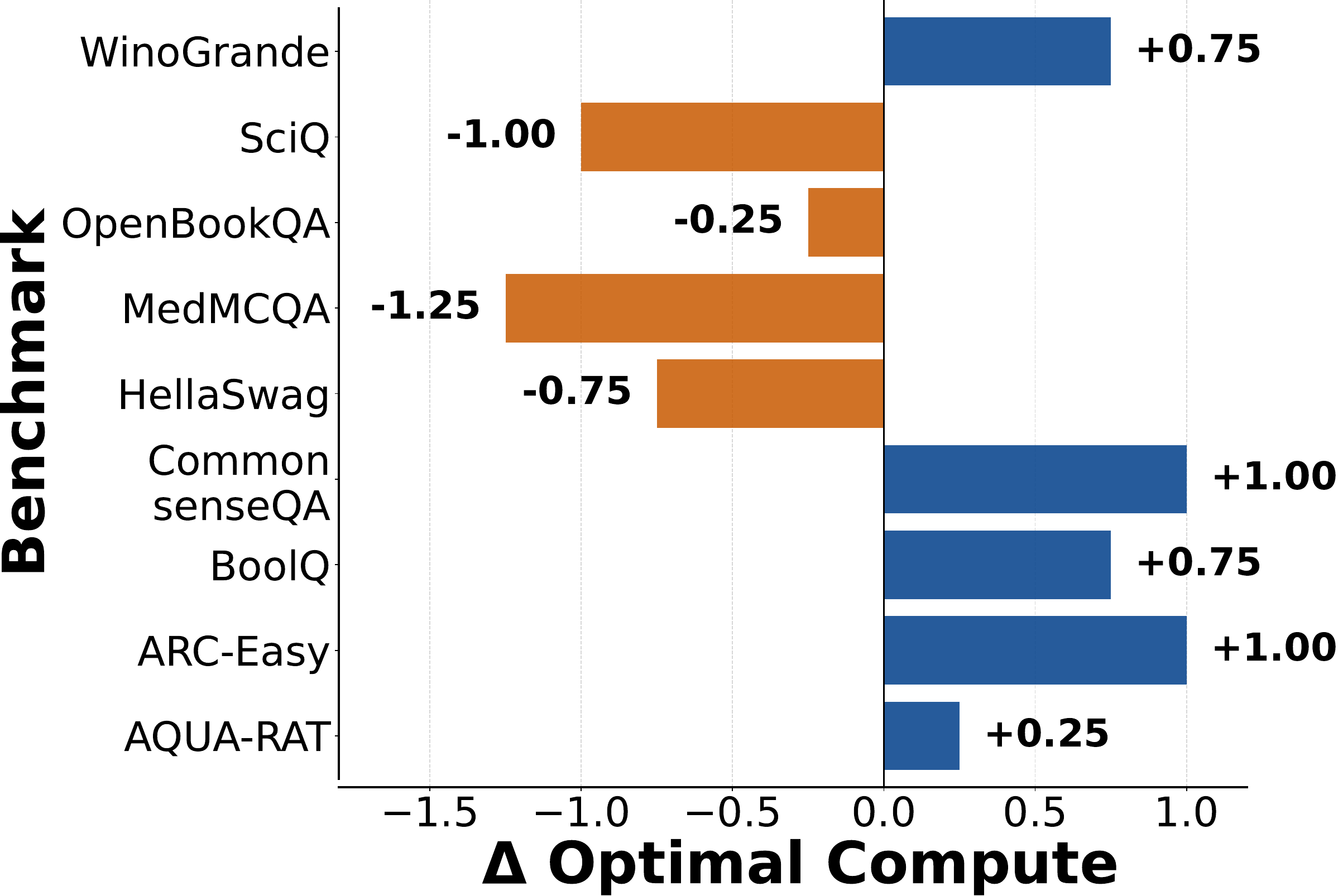} \\
    {\small (c) \ttt{Qwen2.5 3B}} &
    {\small (d) \ttt{Qwen2.5 7B}} \\[1em]
    \includegraphics[width=0.48\textwidth]{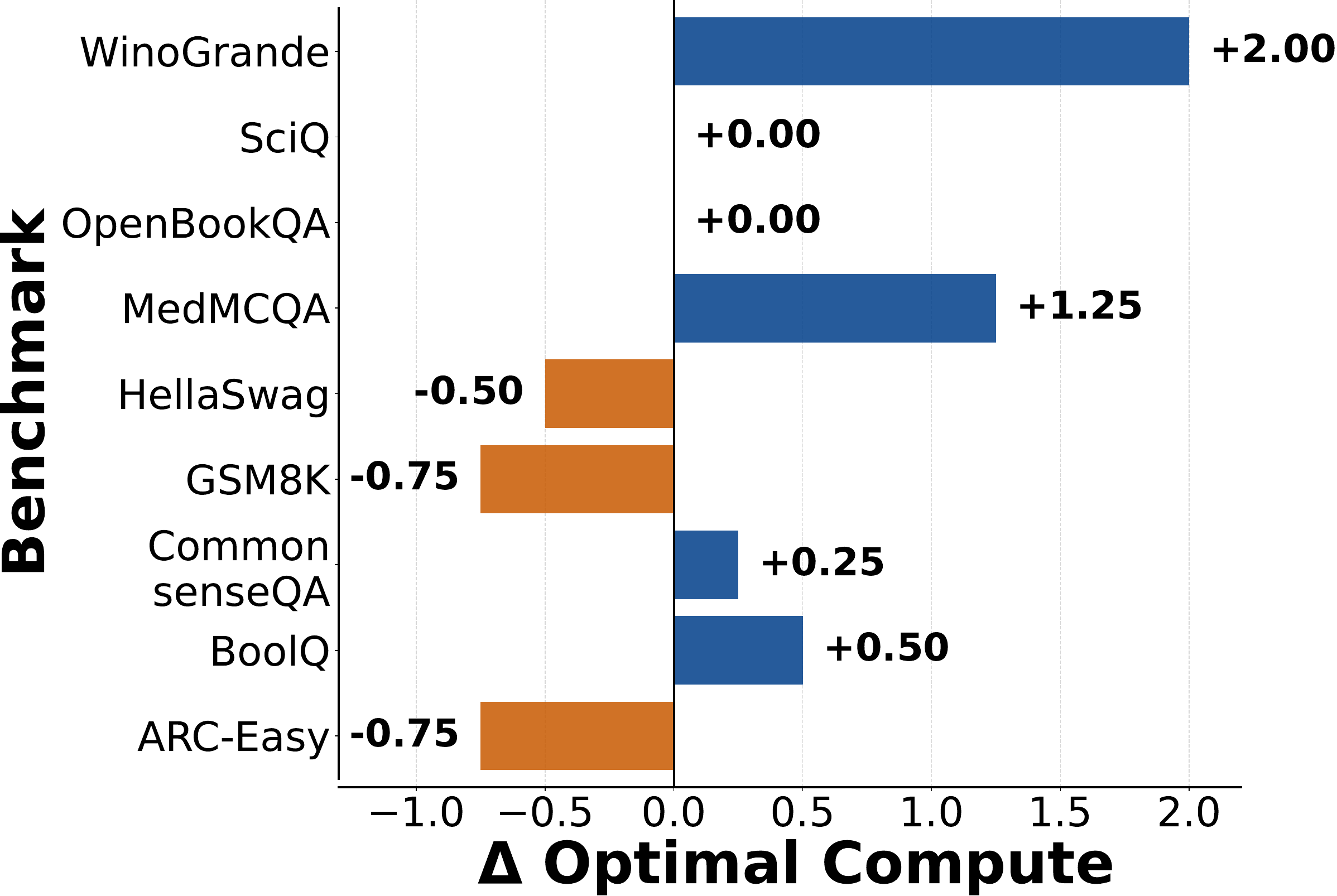} & \\
    {\small (e) \ttt{OLMo 2 1B}} & \\
\end{tabular}
\begin{minipage}{\textwidth}
\captionof{figure}{\textbf{Divergence of optimal compute upon dataset exclusion.} Excluding a small fraction of the training mixture alters the optimization trajectory, shifting optimal stopping points for remaining tasks. $\Delta$ optimal compute varies across individual sub-tasks.}
\label{fig:delta_all_models}
\end{minipage}
\end{center}

\clearpage
\section{Further Experimental Details} \label{app:further_exp}

\subsection{Datasets}\label{app:datasets}

\paragraph{Composition and Statistics.} We use $N=10$ publicly available English benchmarks, listed in Tab. \ref{tab:datasets} together with the task category we assign to each. Examples are drawn from the official splits by random sampling without replacement, at the per-sub-dataset sizes given in Tab. \ref{tab:sft_hyperparams}, yielding 18,000 training and 2,000 validation examples for $N=10$ (9,000 and 27,000 training examples for $N=5$ and $N=15$; \S\ \ref{sec:further}, Fig. \ref{fig:diverse_N}). No further preprocessing, filtering, or de-duplication is applied.

\paragraph{Documentation and Intended Use.} All benchmarks are English, publicly released for research, and are used here in a manner consistent with their intended use and licenses. Per-dataset domains, task-category assignments, and licenses are given in Tab. \ref{tab:datasets}. The base models are released under Apache 2.0, except \ttt{Qwen2.5 3B} (Qwen Research License) and \ttt{Gemma 3 4B} (Gemma Terms of Use), all permit our non-commercial research use. We did not collect, annotate, or release any new data. 

\begin{table*}[b]
\centering
\small
\begin{tabular}{l l l l}
\toprule
\textbf{Benchmark} & \textbf{Assigned Category} & \textbf{Domain} &
\textbf{License} \\
\midrule
SciQ \citep{welbl-etal-2017-crowdsourcing} & Science and Knowledge & Crowdsourced science exam & CC BY-NC 3.0 \\
ARC-Easy \citep{clark2018thinksolvedquestionanswering} & Science and Knowledge & Grade-school science exam & CC BY-SA 4.0 \\
OpenBookQA \citep{mihaylov-etal-2018-suit} & Science and Knowledge & Elementary science facts & Apache 2.0 \\
MedMCQA \citep{pmlr-v174-pal22a} & Science and Knowledge & Medical entrance exam & Apache 2.0 \\
\midrule
CommonsenseQA \citep{talmor-etal-2019-commonsenseqa} & Commonsense and Language &
General commonsense & MIT \\
HellaSwag \citep{zellers-etal-2019-hellaswag} & Commonsense and Language &
Grounded sentence completion & MIT \\
Winogrande \citep{sakaguchi2020winogrande} & Commonsense and Language & Pronoun
resolution & Apache 2.0 \\
BoolQ \citep{clark-etal-2019-boolq} & Commonsense and Language & Yes/no reading
comprehension & CC BY-SA 3.0 \\
\midrule
GSM8K \citep{cobbe2021trainingverifierssolvemath} & Mathematic and Quantitative &
Grade-school word problems & MIT \\
AQUA-RAT \citep{ling-etal-2017-program} & Mathematic and Quantitative & Algebraic
word problems & Apache 2.0 \\
\bottomrule
\end{tabular}
\caption{\textbf{Benchmarks used in our experiments.} All datasets are English.
Assigned Category denotes the grouping used to report results in Tab.
\ref{tab:main}. Each sub-dataset contributes 1,800 training and 200 validation
examples.}
\label{tab:datasets}
\end{table*}

\subsection{Hardware}
We use B200, H200, RTX A5000, and RTX 3090s for experiments. For other hardware like CPU and RAM we use commonly available ones, as these hardware did not induce any bottlenecks.

\subsection{Common Settings}
Default training settings universal across methods are available in Tab. \ref{tab:sft_hyperparams}. We use a single seed (20) as preliminary experiments with \ttt{Qwen2.5 3B} and \ttt{7B} on seeds 20, 30, 40 lead to virtually identical performance gains. Tab. \ref{tab:seed} shows that the gains of \tsc{mSFT} is stable (low standard deviation) and thus statistically significant. This likely due to our methods and experiments being non-stochastic in nature. 

\begin{table}[h]
\centering
\begin{tabular}{lcccc}
\toprule
\textbf{Hyperparameter} & \textbf{Value} \\
\midrule
Learning Rate & $1 \times 10^{-5}$ \\
Learning Rate Schedule & Constant \\
Batch Size & 64 \\
Seed & 20 \\
Sub-dataset Size (train) & 1800 \\
Sub-dataset Size (validation) & 200 \\
\bottomrule
\end{tabular}
\caption{Overlapping hyperparameters.}
\label{tab:sft_hyperparams}
\end{table} 

\begin{table}[h]
\centering
\small
\begin{tabular}{l rr rr}
\toprule
\textbf{Model:} & \multicolumn{4}{c}{\textbf{\ttt{Qwen2.5}}} \\
\cmidrule(lr){2-5}
\textbf{Size:} & \multicolumn{2}{c}{\textbf{\ttt{3B}}} & \multicolumn{2}{c}{\textbf{\ttt{7B}}} \\
\cmidrule(lr){2-3} \cmidrule(lr){4-5}
& \textbf{SFT} & \textbf{\tsc{mSFT}} & \textbf{SFT} & \textbf{\tsc{mSFT}} \\
\midrule
\multicolumn{5}{c}{\cellcolor{myorange!15}\textbf{Average Accuracy Across 10 Benchmarks}} \\
\midrule
Seed 20 & 73.25 & 74.25 & 76.80 & 78.05 \\
Seed 30 & 73.05 & 74.05 & 76.90 & 78.85 \\
Seed 40 & 72.65 & 73.25 & 76.80 & 78.05 \\
\midrule
\cellcolor{gray!15}Mean & \cellcolor{gray!15}72.98 & \cellcolor{gray!15}\textbf{73.85}$_{\color{mygreen}+0.87}$ &
\cellcolor{gray!15}76.83 & \cellcolor{gray!15}\textbf{78.32}$_{\color{mygreen}+1.48}$ \\
Std Dev & 0.31 & 0.53 & 0.06 & 0.46 \\
$p$-value & --- & 0.023$^{*}$ & --- & 0.024$^{*}$ \\
\bottomrule
\end{tabular}
\caption{\textbf{Seed stability on \ttt{Qwen2.5 3B} and \ttt{7B}.} Subscripts in the Mean row denote $\Delta$ relative to
SFT. $p$-values are from a two-sided paired $t$-test against SFT ($^{*}p < 0.05$).}
\label{tab:seed}
\end{table}

\begin{figure*}[tbp]
    \centering
    \includegraphics[width=\linewidth]{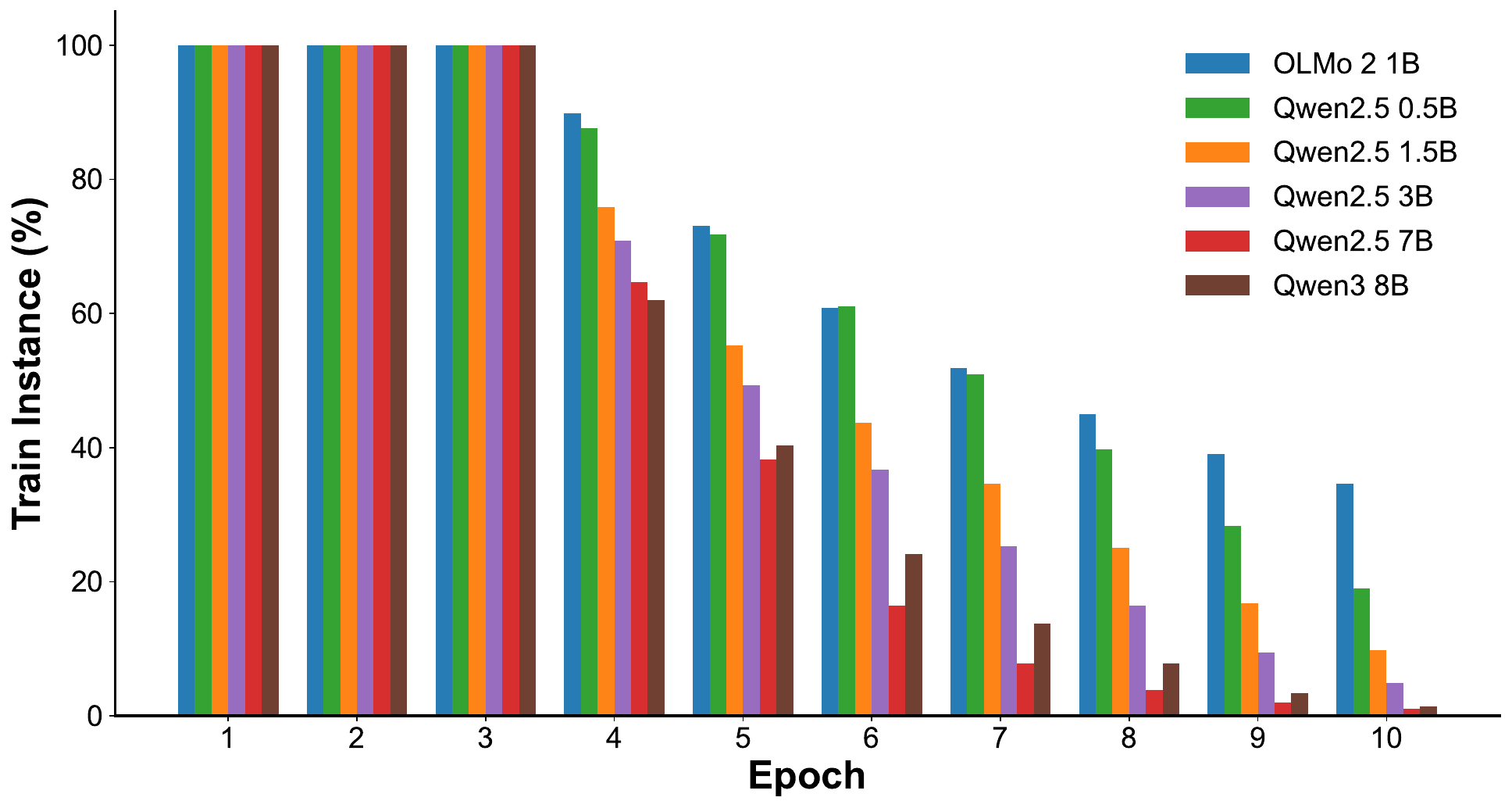} 
    \caption{\textbf{Training instances across epochs on IES.} Percentage of active training instances per epoch, relative to the initial dataset size at Epoch 1. All models process the complete dataset for the first three epochs, after which the proportion of active instances consistently decreases.}
    \label{fig:remaining_instances}
\end{figure*}

\subsection{Method-specific Settings}

SFT trains for 10 epochs as we observe that in some datasets do not overfit even up to 10 epochs (see Fig. \ref{fig:motivation} and Appendix \ref{app:motivation_app}). Continual SFT and \tsc{mSFT}'s compute budget is $C=3$ epochs. DynamixSFT was first run on the settings provided in the paper \citep{shin2025dynamixsft}, yet we found that further hyperparameter tuning, where sharpness factor $\beta = 5000$ improved performance in our environment so we used this for all reported experiment results. For IES, we adopt the threshold of $\delta = 0.01$ tuned for our experimental setting. \citep{yuan2025instancedependent}. The cumulative proportion of dropped instances over 10 epochs is visualized in Fig. \ref{fig:remaining_instances}. For SRO SFT and Soft SRO SFT the single search compute budget is set to $C=10$ as this is conceptually similar to the 10 epochs allocated in SFT.  

\section{Computation of Empirical FLOPS}
\label{app:FLOPS}

We calculate computation costs using the standard formula from \citet{kaplan2020scalinglawsneurallanguage}:
\begin{align*}
  \text{FLOPS}_{\text{train}} &= 6 \times \vert \theta \vert \times t, \\
  \text{FLOPS}_{\text{inference}} &= 2 \times \vert\theta\vert \times t.
\end{align*}
where $\vert\theta\vert$ is the number of model parameters and $t$ is the number of tokens.

\subsection{Method-specific FLOPs}
Let $t_\text{train}$ and $t_\text{validation}$ denote the total training and validation tokens per unit compute budget (1 epoch) over the full mixture $\mathcal{D}$.

\paragraph{[1] SFT.}
Standard supervised fine-tuning on all sub-datasets for $C$ units of compute budget.
\[
  \text{FLOPs}_{\text{SFT}}
  = \sum_{c=1}^{C} \bigl[6 \cdot \vert\theta\vert \cdot t_\text{train}
    + 2 \cdot \vert\theta\vert \cdot t_\text{validation}\bigr].
\]

\paragraph{[2] Continual SFT.}
Sequential training \citep{scialom-etal-2022-fine} configured with a 0\% rehearsal rate: each sub-dataset $\mathcal{D}_i$ is trained independently for $C$ units of compute budget before moving to the next.

\[
  \text{FLOPs}_{\text{Cont}}
  = \sum_{i}\bigl[
      6 \cdot \vert\theta\vert \cdot t_{tr,i}
    + 2 \cdot \vert\theta\vert \cdot t_\text{validation} 
  \bigr] \cdot C,
\]
summing over all $N$ sub-datasets trained sequentially.

\paragraph{[3] DynamixSFT.}
Dynamic mixture optimization \citep{shin2025dynamixsft} via multi-armed bandits with 1-step look-ahead. At each update step (1\% of total steps), the algorithm samples batches of size $B_{\text{look-ahead}}$ for all $N$ sub-datasets and performs forward-backward passes to estimate look-ahead rewards, incurring $8\vert\theta\vert$ FLOPs per token (2 forward pre-loss, 4 backward, 2 forward post-loss). Between updates, training proceeds with current mixture probabilities:
\begin{align*}
  \text{FLOPs}_{\text{Dynamix}}
  &= \underbrace{\sum_{c=1}^{C} 6 \cdot \vert\theta\vert \cdot t_\text{train}}_{\text{training}} \\
  &\quad + \underbrace{\sum_{t_u} N \cdot 8 \cdot \vert\theta\vert \cdot B_{\text{look-ahead}} \cdot t_\text{avg}}_{\text{look-ahead}} \\
  &\quad + \sum_{c=1}^{C} 2 \cdot \vert\theta\vert \cdot t_\text{validation},
\end{align*}

where $B_{\text{look-ahead}}$ is batch size for look-ahead,  $t_\text{avg}$ is the average tokens per sample and $t_u$ denotes update steps.

\begin{table*}[t]
\centering
\begin{tabular}{l|rrrrrrrr}
\toprule
 & & & & & & \multicolumn{1}{c}{\textbf{Soft}} & \multicolumn{2}{c}{\textbf{\tsc{mSFT}}} \\
\cmidrule(lr){8-9}
\textbf{Model} & \textbf{SFT} & \textbf{Cont.} & \textbf{Dynamix} & \textbf{IES} & \textbf{SRO} & \multicolumn{1}{c}{\textbf{SRO}} & (C=1) & (C=3) \\
\midrule
\ttt{OLMo 2 1B}    & 157.19& 124.52& 263.22& 112.05& 218.66& 273.35& 74.06& 210.11\\
\ttt{Qwen2.5 0.5B} & 58.10& 47.41& 103.54& 37.19& 80.35& 120.93& 31.24& 105.78\\
\ttt{Qwen2.5 1.5B} & 207.90& 308.98& 363.97& 135.80& 374.98& 443.58& 144.21& 416.26\\
\ttt{Qwen2.5 3B}   & 461.48& 727.13& 760.17& 313.30& 652.51& 879.80& 320.14& 770.21\\
\ttt{Qwen2.5 7B}   & 1089.42& 965.96& 1742.91& 663.91& 1717.30& 2057.52& --      & 1654.26\\
\ttt{Qwen3 8B}     & 942.34& 755.55& 1826.68& 418.69& 1409.65& 1986.82& --      & 1588.39\\
\midrule
\textbf{Average}   & 486.07& 488.26& 843.42& 280.16& 742.24& 960.33& --      & 790.83\\
\bottomrule
\end{tabular}
\caption{\textbf{Total PFLOPs for each method across model scales.}}
\label{tab:flops_comparison}
\end{table*}

\paragraph{[4] IES.}
Instance-dependent early stopping \citep{yuan2025instancedependent} computes second-order differences of per-sample loss trajectories to identify mastered instances. Samples satisfying the convergence criterion are excluded from gradient updates (typically from the 3rd unit onward). Training FLOPs decrease as more samples are excluded, while validation always covers the full dataset. 
\[
  \text{FLOPs}_{\text{IES}}
  = \sum_{c=1}^{C}\bigl[
      6 \cdot \vert\theta\vert \cdot t_\text{train}^{(c)}
    + 2 \cdot \vert\theta\vert \cdot t_\text{validation}
  \bigr],
\]
where $t_\text{train}^{(c)} \le t_\text{train}$ reflects the remaining active samples at $c$.

\paragraph{[5] SRO SFT.}
Single roll-out searched SFT: a two-step procedure.
\textbf{Step~1 (Search):} Standard SFT for $C$ units to determine per sub-dataset peak $c_i^*$, which is also their drop schedule.
\textbf{Step~2 (Train):} Training with sub-datasets exclusions applied at their respective peak checkpoints; dropped sub-datasets are removed from the active token count.
\begin{align*}
  \text{FLOPs}_{\text{SRO}}
  &= \underbrace{\text{FLOPs}_{\text{SFT}}}_{\text{step 1}} 
  + \sum_{c=1}^{C}\bigl[
      6 \cdot \vert\theta\vert \cdot t_\text{train}^{(c)} \\
     &\quad + 2 \cdot \vert\theta\vert \cdot t_\text{validation}
  \bigr]_{\text{step 2}},
\end{align*}

where $t_\text{train}^{(c)} \le t_\text{train}$ denotes training tokens over non-excluded sub-datasets at step $c$ in Step~2.

\paragraph{[6] Soft SRO SFT.}
\textbf{Step~1:} Identical to SRO SFT Step~1, recording per-sub-dataset peak $c_i^*$.
\textbf{Step~2:} Re-trains from scratch for $C$ units on a reweighted mixture, where the number of samples drawn from sub-dataset $\mathcal{D}_i$ is
\[
  n_i' = \left(\sum_j |\mathcal{D}_j|\right) \cdot \frac{c_i^* \cdot |\mathcal{D}_i|}{Z},
\]
where $Z = \sum_j c_j^* \cdot |\mathcal{D}_j|$,
so sub-datasets that peak later and are larger receive more exposure.
The total FLOPs are:
\begin{align*}
  \mathrm{FLOPs}_{\text{Soft}}
  &= \underbrace{\mathrm{FLOPs}_{\text{SFT}}}_{\text{step 1}} + \sum_{c=1}^{C}\Bigl[
      6\,\cdot |\theta| \cdot \sum_{i} n_i' \cdot \bar{l}_i \\
    &\quad+ 2\,\cdot |\theta| \cdot t_{\text{validation}}
  \Bigr]_{\text{step 2}},
\end{align*}
where $\bar{l}_i$ is the average token length of $\mathcal{D}_i$.

\paragraph{[7] \tsc{mSFT}.}
\tsc{mSFT} proceeds in $S$ stages indexed by $s = 1, \ldots, S$.
At each stage $s$, the model trains for $C$ units on active subsets $\mathcal{D} \setminus \mathcal{E}_s$, where $\mathcal{E}_s$ is the accumulated exclusion set at stage $s$.
Overfit sub-datasets are added to $\mathcal{E}_{s+1}$ and the model reverts to the earliest overfitting checkpoint (parameter rollback only; no additional FLOPs).
\begin{align*}
  \text{FLOPs}_{\text{stage}_s}
  &= \underbrace{6 \cdot \vert\theta\vert \cdot C \cdot t_\text{train}(\mathcal{D}{\setminus}\mathcal{E}_s)}_{\text{training on active sets}} \\
  &\quad+ \underbrace{2 \cdot \vert\theta\vert \cdot C \cdot t_\text{validation}(\mathcal{D}{\setminus}\mathcal{E}_s)}_{\text{validation on active sets}}\\  
  &\quad+ \underbrace{2 \cdot \vert\theta\vert \cdot t_\text{validation}(\mathcal{E}_s)}_{\text{validation on excluded sets}},
\end{align*}
where $t_\text{train}(\mathcal{D}{\setminus}\mathcal{E}_s)$ and $t_\text{validation}(\mathcal{D}{\setminus}\mathcal{E}_s)$ decreases as more sub-datasets are excluded. $t_\text{validation}(\mathcal{E}_s)$ denotes validation tokens of excluded sub-datasets. 
Note that the third term carries no compute budget $C$: excluded sets are validated only once at the rollback checkpoint to preserve the full validation trajectory, where as active sub-datasets are validated at every checkpoint throughout the stage. 
Total FLOPs: $\text{FLOPs}_{\tsc{mSFT}} = \sum_{s=1}^{S} \text{FLOPs}_{\text{stage}_s}$.

\paragraph{Empirical FLOPs comparison.}

Tab.~\ref{tab:flops_comparison} reports the total FLOPs for each method across six model scales. DynamixSFT incurs substantial look-ahead overhead (94.9\% of training FLOPs on average), while IES achieves costs smaller than SFT by dropping parts of samples from 3rd unit of compute budget onward. SRO SFT and Soft SRO SFT require an additional search phase (Step~1), resulting in higher total costs, though Soft SRO mitigates catastrophic forgetting via soft reweighting rather than hard exclusions.

\clearpage
\section{Further Loss Curves}\label{app:loss}

\begin{center}
\begin{tabular}{cc}
    \includegraphics[width=0.48\textwidth]{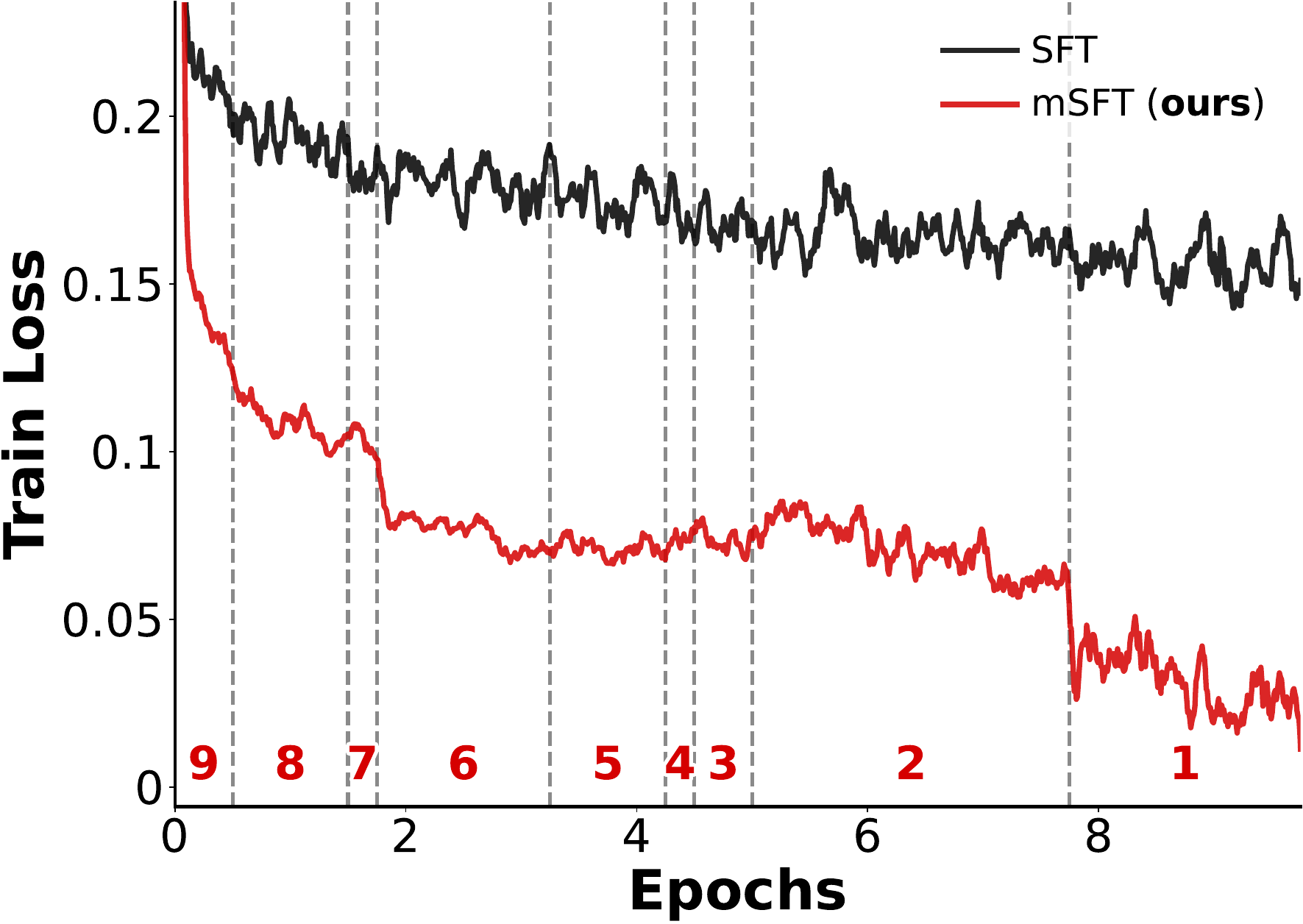} &
    \includegraphics[width=0.48\textwidth]{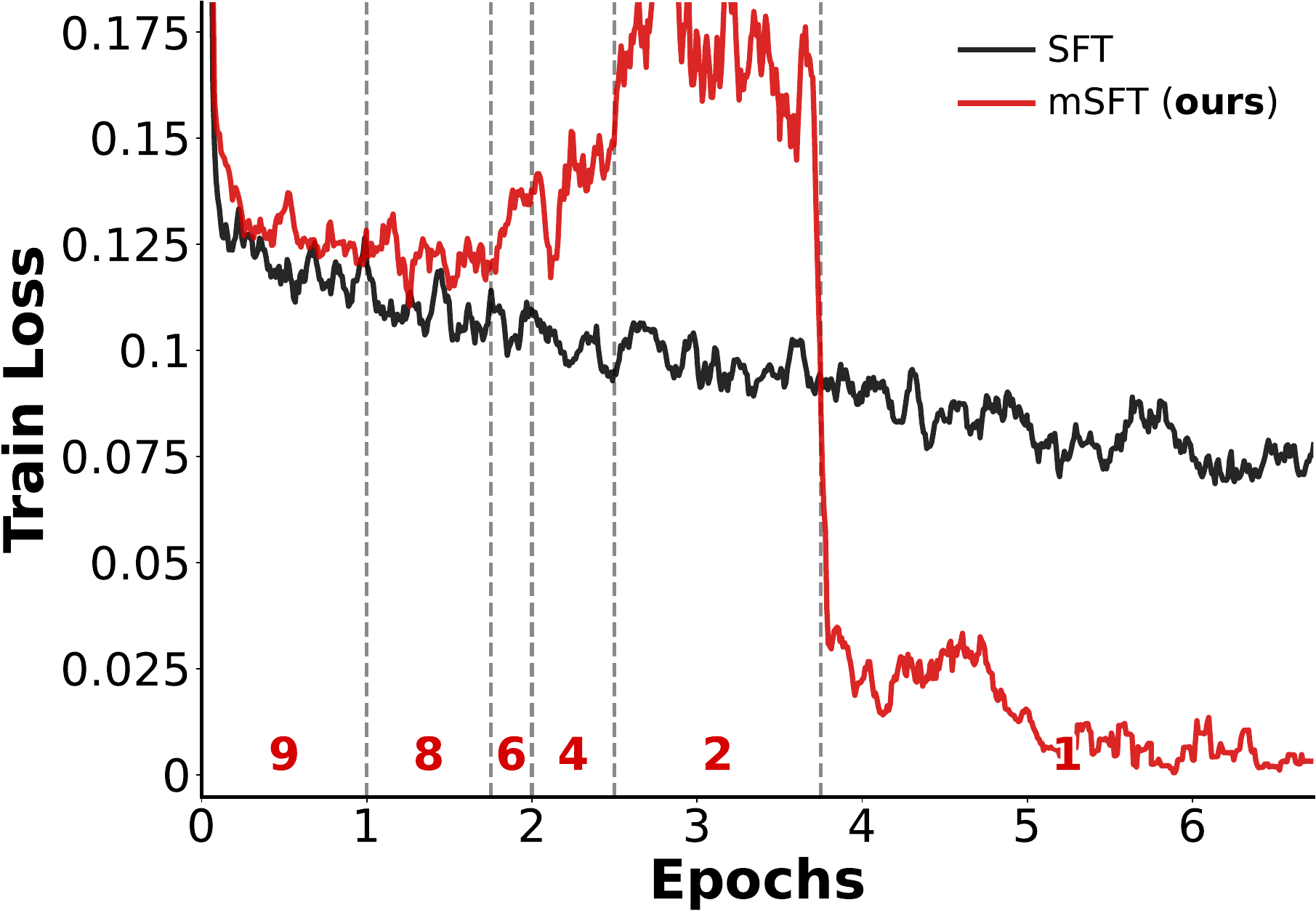} \\
    {\small (a) \ttt{Olmo 2 1B}, $C=3,\;N=10$} &
    {\small (b) \ttt{Qwen2.5 1.5B}, $C=3,\;N=10$} \\[1em]
    \includegraphics[width=0.48\textwidth]{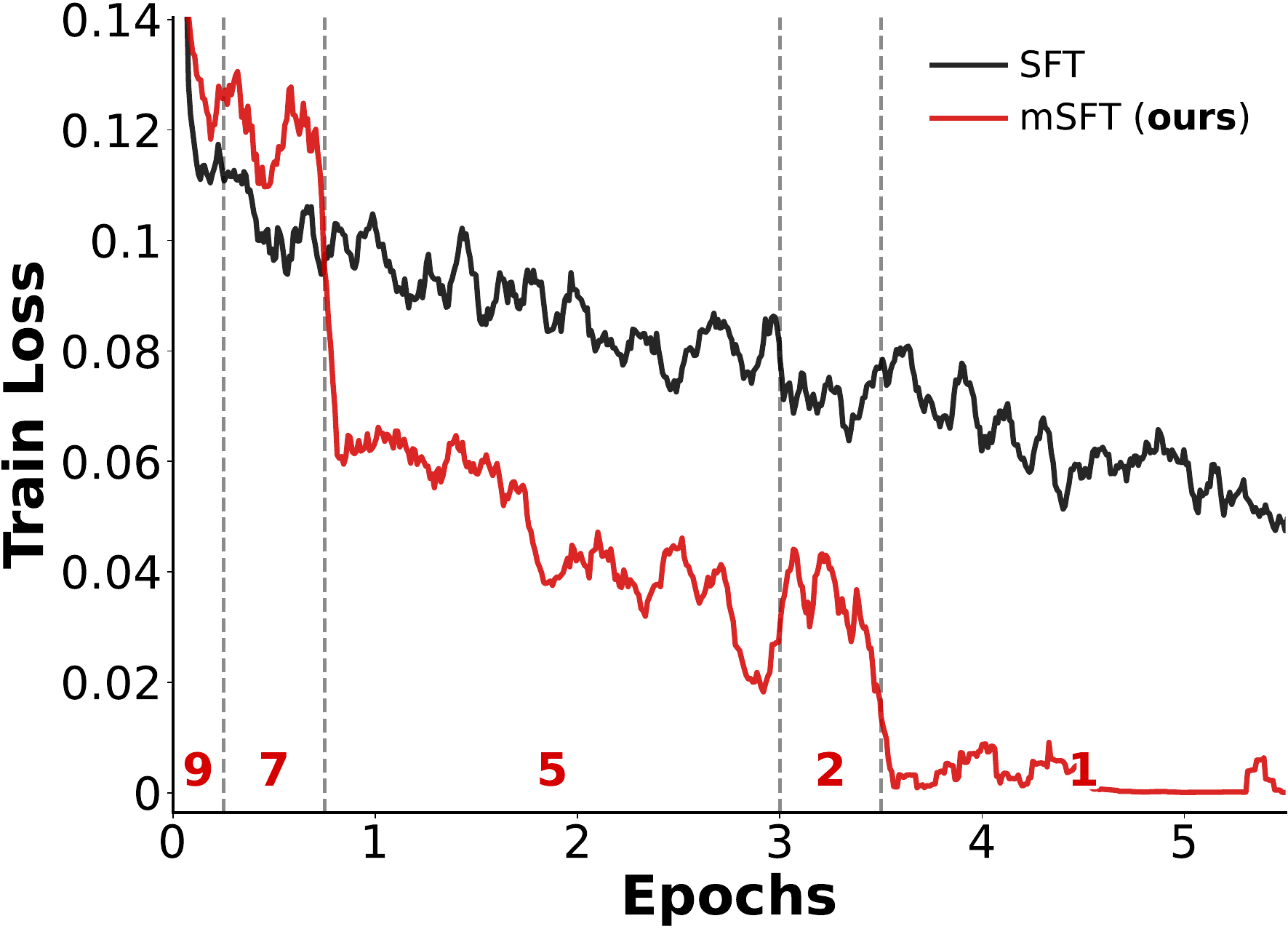} &
    \includegraphics[width=0.48\textwidth]{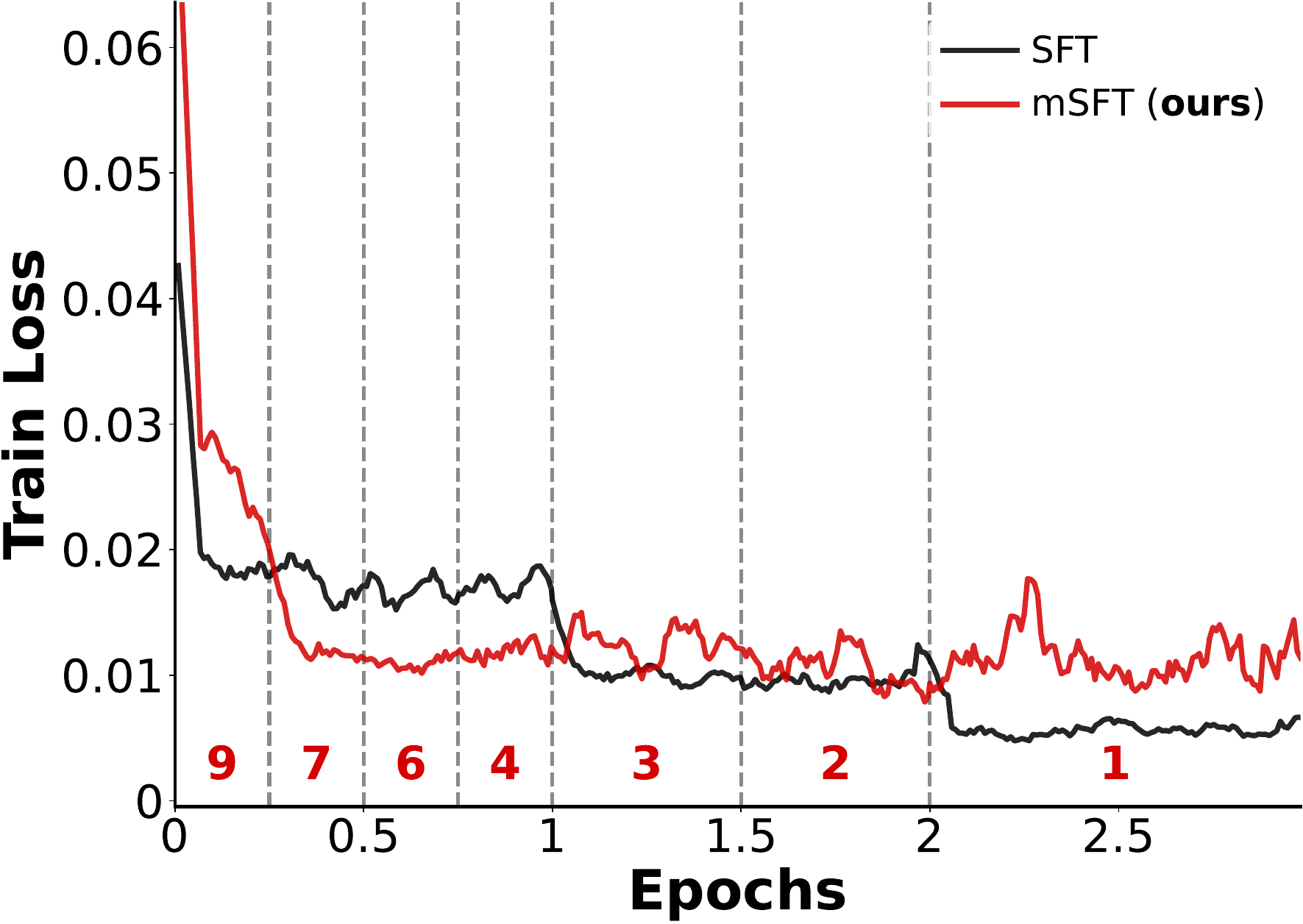} \\
    {\small (c) \ttt{Qwen2.5 3B}, $C=3,\;N=10$} &
    {\small (d) \ttt{Qwen2.5 7B}, $C=3,\;N=10$} \\[1em]
    \includegraphics[width=0.48\textwidth]{figs/loss_Qwen3_8B_c3.pdf} &
    \includegraphics[width=0.48\textwidth]{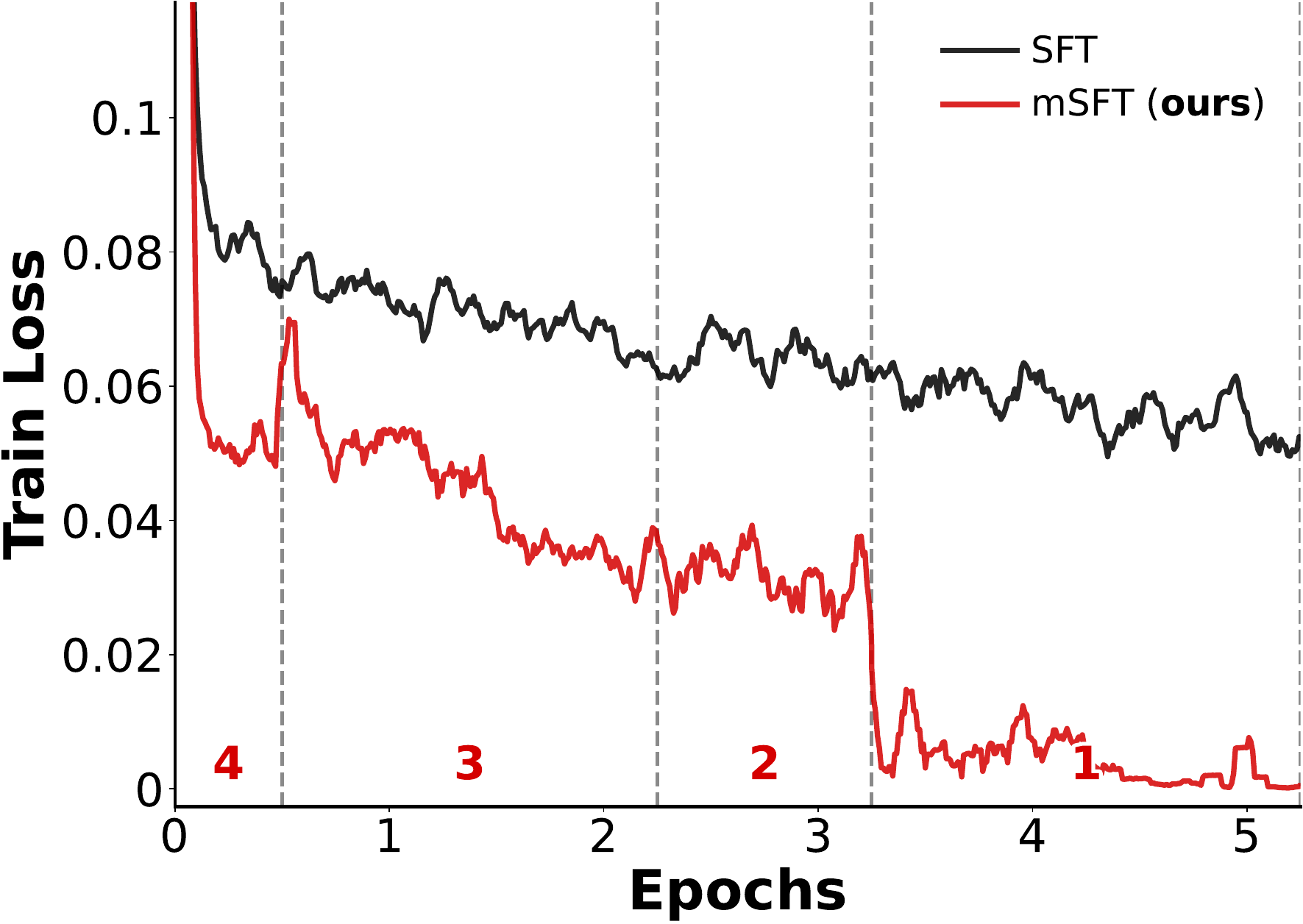} \\
    {\small (e) \ttt{Qwen3 8B}, $C=3,\;N=10$} &
    {\small (f) \ttt{Qwen2.5 3B}, $C=3,\;N=5$} \\
\end{tabular}
\begin{minipage}{\textwidth}
\captionof{figure}{\textbf{Training loss curve comparison.} Smoothed with moving average with sliding window 10. Dashed vertical lines denote roll-back where a sub-dataset is excluded. Numerical annotation at the bottom indicate the number of remaining sub-datasets at each interval.}
\label{fig:app_loss_1}
\end{minipage}
\end{center}

\clearpage

\begin{center}
\begin{tabular}{cc}
    \includegraphics[width=0.48\textwidth]{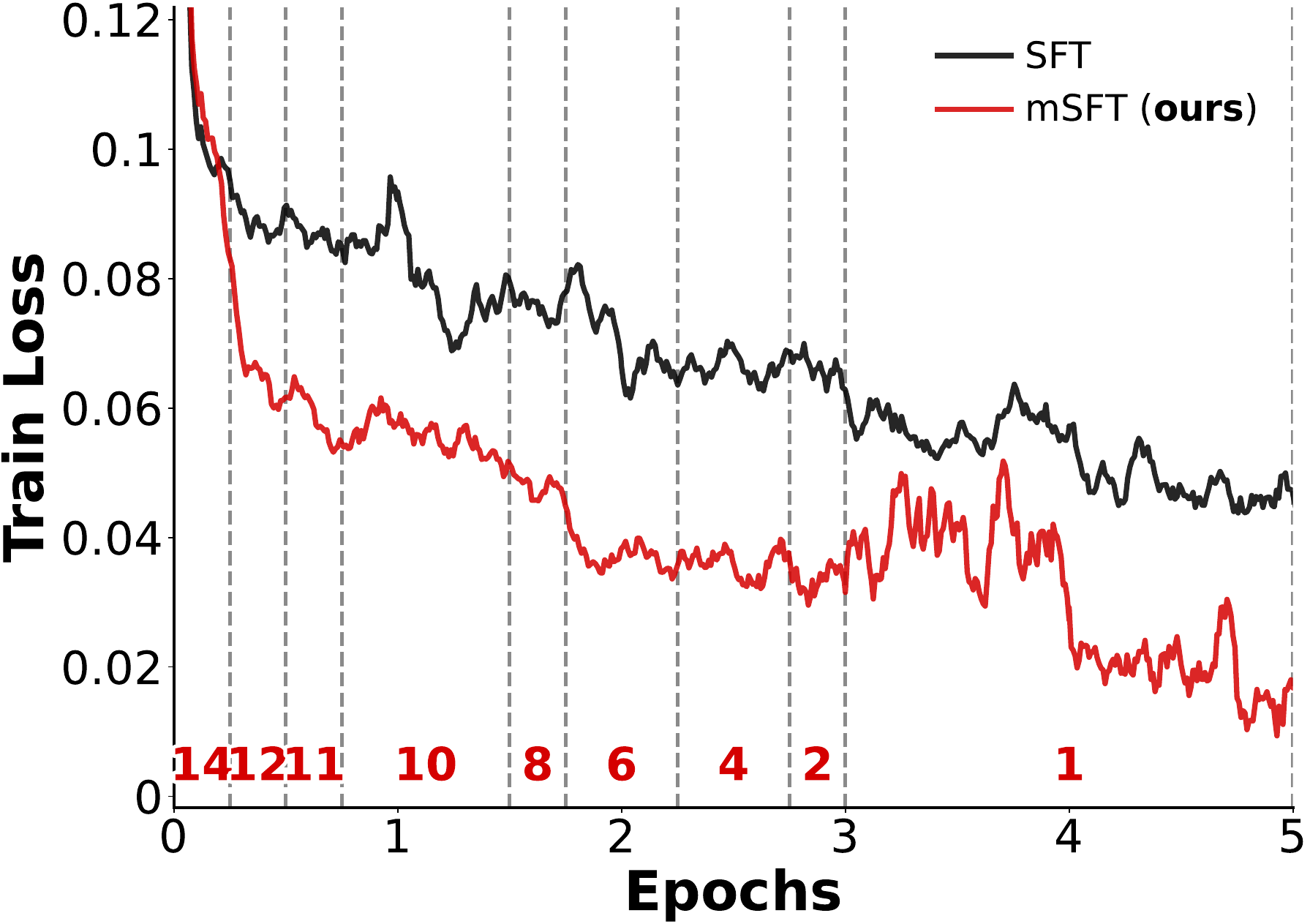} &
    \includegraphics[width=0.48\textwidth]{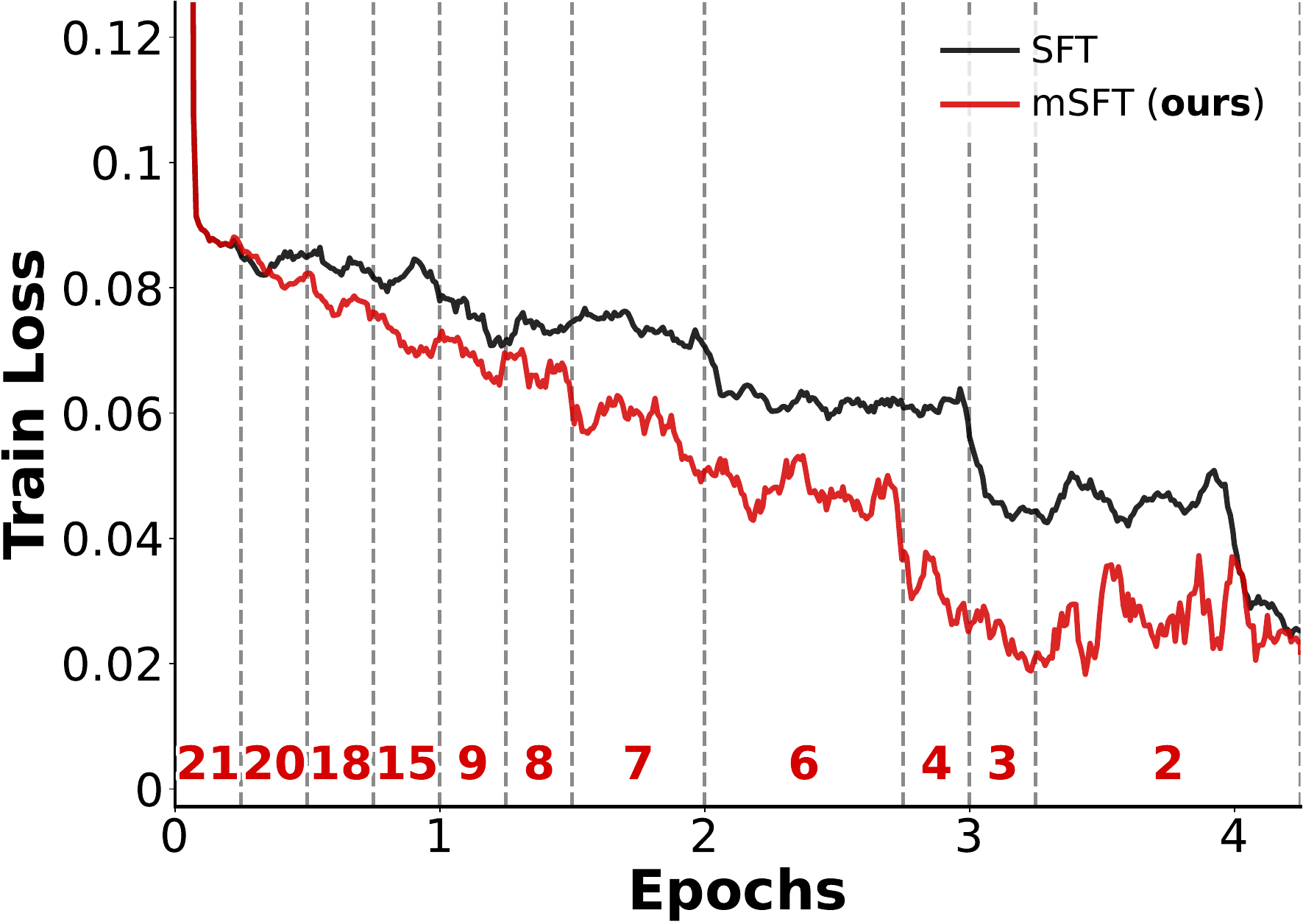} \\
    {\small (a) \ttt{Qwen2.5 3B}, $C=3,\;N=15$} &
    {\small (b) \ttt{Qwen2.5 3B} on MedMCQA, $C=3,\;N=21$} \\[1em]
    \includegraphics[width=0.48\textwidth]{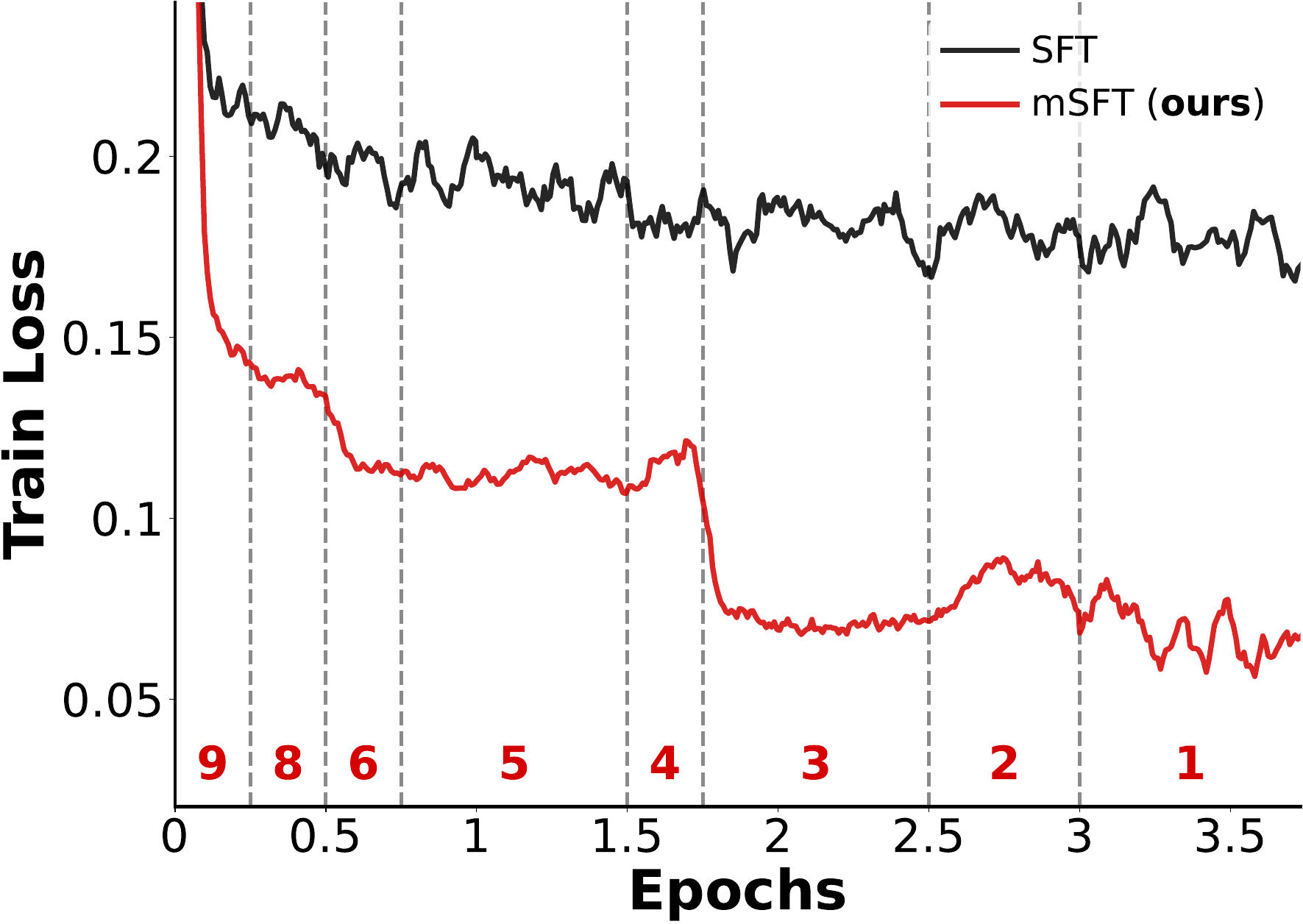} &
    \includegraphics[width=0.48\textwidth]{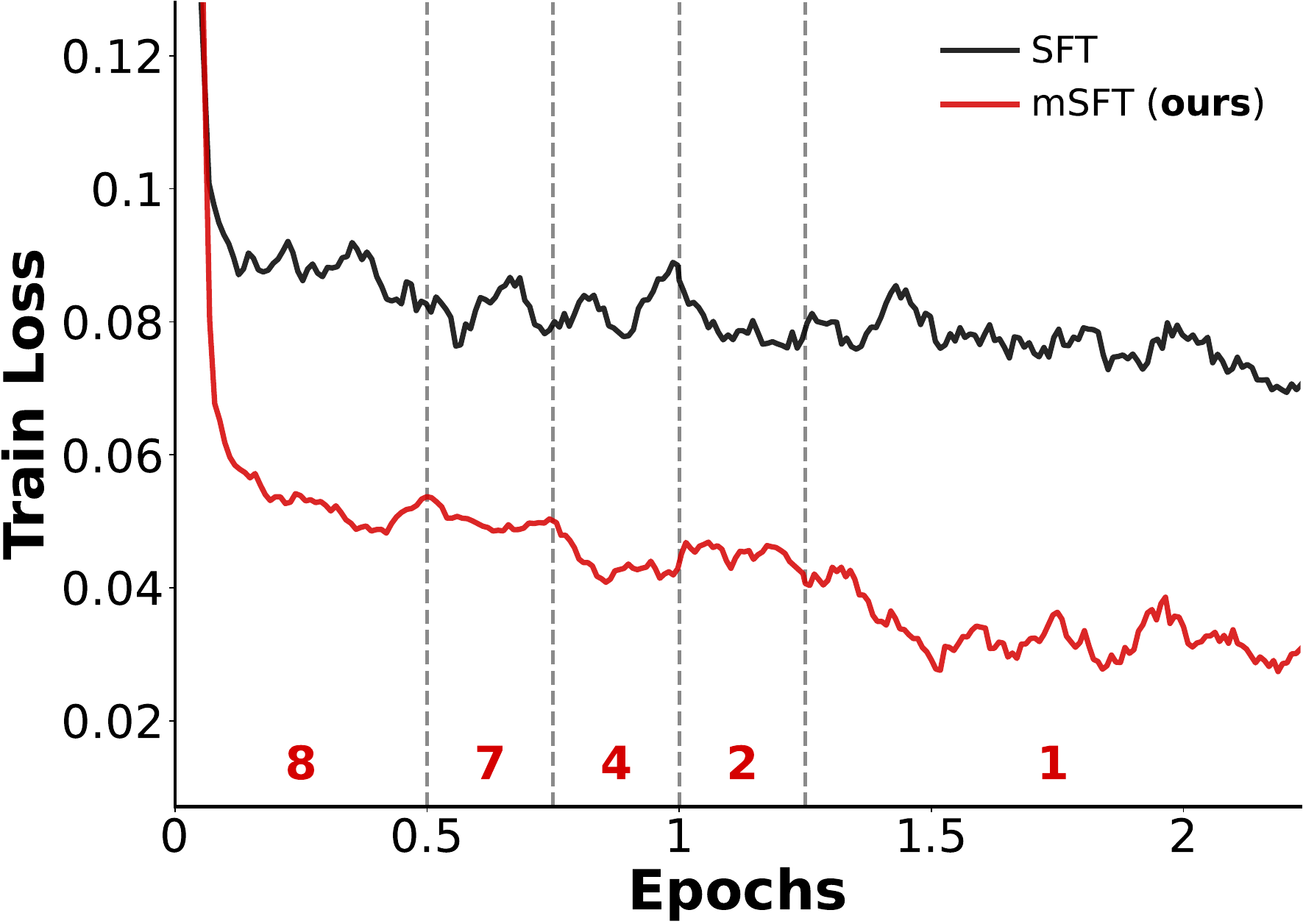} \\
    {\small (c) \ttt{Olmo 2 1B}, $C=1,\;N=10$} &
    {\small (d) \ttt{Qwen2.5 0.5B}, $C=1,\;N=10$} \\[1em]
    \includegraphics[width=0.48\textwidth]{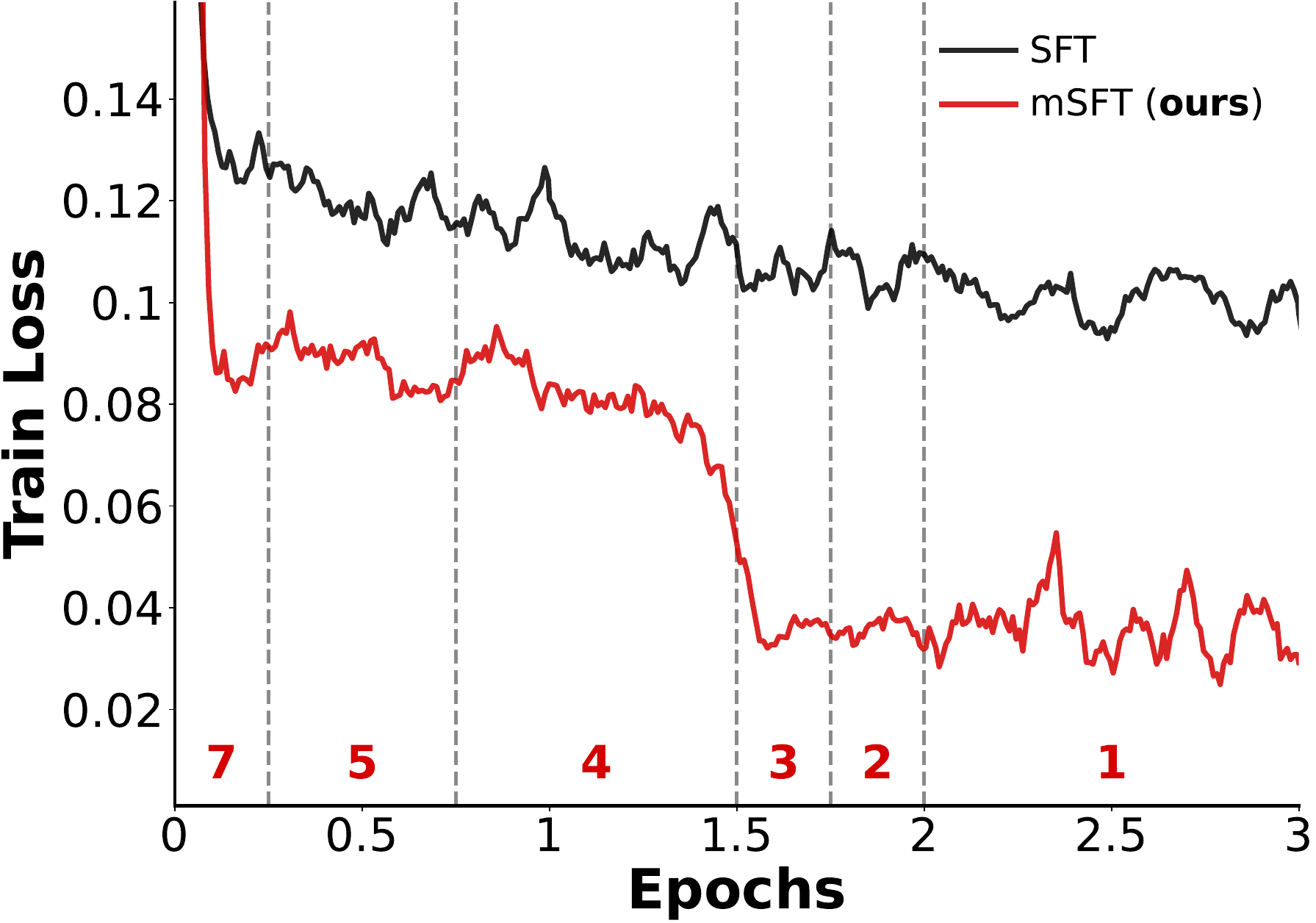} &
    \includegraphics[width=0.48\textwidth]{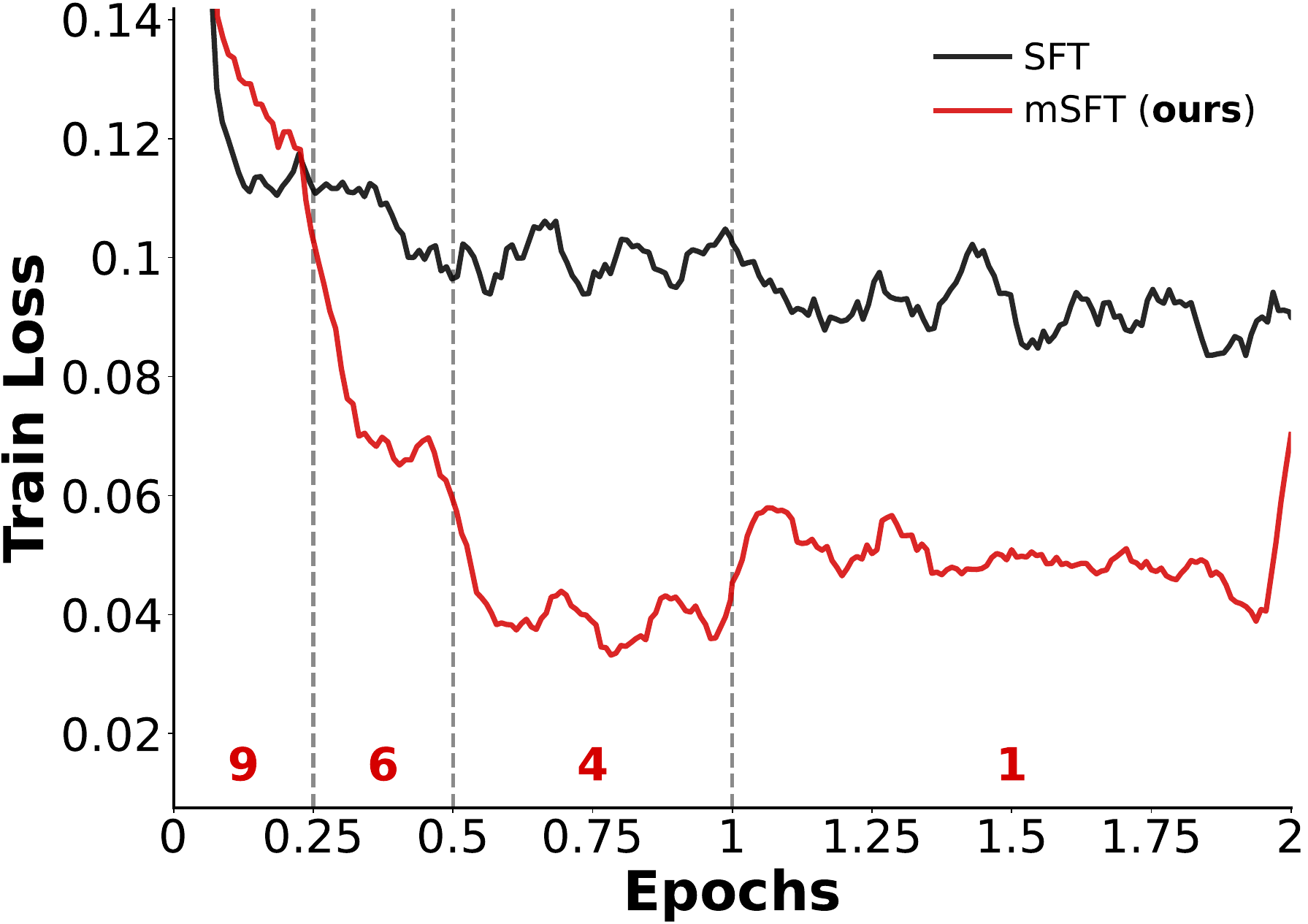} \\
    {\small (e) \ttt{Qwen2.5 1.5B}, $C=1,\;N=10$} &
    {\small (f) \ttt{Qwen2.5 3B}, $C=1,\;N=10$} \\
\end{tabular}
\begin{minipage}{\textwidth}
\captionof{figure}{\textbf{Training loss curve comparison.} Smoothed with moving average with sliding window 10. Dashed vertical lines denote roll-back where a sub-dataset is excluded. Numerical annotation at the bottom indicate the number of remaining sub-datasets at each interval.}
\label{fig:app_loss_2}
\end{minipage}
\end{center}

\clearpage
\section{\tsc{mSFT} with Efficient Disk Management}\label{app:efficient_disk}

\begin{algorithm2e}[t]
    \caption{\tsc{mSFT} with Checkpoint Management}
    \label{alg:msft-disk}
    \SetAlgoLined
    \LinesNumbered
    \Input{Dataset mixture $\mc{D}$, base model $\theta_0$, compute budget $C$}

    $\mc{E} \leftarrow \emptyset ;$  $\hat{\theta} \leftarrow \theta_0$
    \textcolor{darkblue}{$\theta^* \leftarrow \theta_0;\; a^* \leftarrow 0$} \tcp*{Initialization}

    \While{$\mc{D} \setminus \mc{E} \neq \emptyset$}{
        \tcc{\textcolor{mygreen}{\textbf{Roll-out}}: Search for per-sub-dataset peaks}
        $\theta, \;\{\text{acc}(\mc{D}_i, c)\}_{i,c} \leftarrow \tsc{SFT-Roll-out}\!\left(\hat{\theta},\; \mc{D} \setminus \mc{E},\; C\right)$ \;

        $c_i^* \leftarrow \arg\max_{c}\; \text{acc}(\mc{D}_i, c) \quad \forall \mc{D}_i \notin \mc{E}$ \tcp*{Optimal compute per sub-dataset}
        \textcolor{darkblue}{\tcc{During the roll-out, checkpoints $\theta(c_i^*)$ for remaining datasets $\forall \mc{D}_i \notin \mc{E}$ are written to Disk}}

        $c_{\min}, \mc{D}_\text{exclude} \leftarrow \arg\min_{\mc{D}_i \notin \mc{E}}\; c_i^*$ \;

        \eIf{$c_{\min} = C$}{
            \tcc{No overfitting: update model and continue}
            $\hat{\theta} \leftarrow \theta(C)$ \;
        }{
            \tcc{\textcolor{myred}{\textbf{Roll-back}}: Revert to the checkpoint where the sub-dataset overfit}
            $\mc{E} \leftarrow \mc{E} \cup \{\mc{D}_{\text{exclude}}\}$ \;
            $\hat{\theta} \leftarrow $ \textcolor{darkblue}{Load $\theta(c_\text{min})$ from Disk} \tcp*{Revert to checkpoint at $c_{\min}$}
        }
        \textcolor{darkblue}{\tcc{Update $\theta^*$ to be the model parameters of the highest accuracy}}

        \textcolor{darkblue}{$c_\text{best} \leftarrow \arg\max_c \;\text{acc}(\mc{D}, c);\quad a_\text{best} \leftarrow \text{acc}(\mc{D}, c_\text{best})$\;}
        \textcolor{darkblue}{\If{$a_{\text{best}} > a^*$}
        {$a^* \leftarrow a_{\text{best}};\;$ $\theta^* \leftarrow \theta(c_\text{best})$\;}}

        \textcolor{darkblue}{Discard all checkpoints from Disk except $\hat{\theta}$ and $\theta^*$\;}
    }
    \textcolor{darkblue}{\Return $\theta^*$}
\end{algorithm2e}

\paragraph{\textcolor{darkblue}{Checkpoint management.}} Algorithm~\ref{alg:msft-disk} details the checkpoint management strategy integrated into \tsc{mSFT}, where blue annotations denote disk-management operations added atop the base algorithm. While standard SFT retains only a single checkpoint on disk throughout training, \tsc{mSFT} requires additional storage during the \textcolor{mygreen}{\textbf{roll-out}} phase: per-dataset peak checkpoints ${\theta(c_i^*)}_{\mc{D}_i \notin \mc{E}}$ are persisted as they are identified (line~5), requiring up to $|\mc{D} \setminus \mc{E}_s|$ checkpoints at stage $s$.
Upon completing each iteration, the algorithm retains only the rollback checkpoint $\hat{\theta}$ and the global best checkpoint $\theta^*$ --- the model that achieved the highest overall accuracy across all stages --- and discards all remaining checkpoints (lines~13--18).
The theoretical peak occurs at the second stage, where $|\mc{D}| - 1$ live per-dataset peaks coexist with the two retained checkpoints ($\hat{\theta}$ and $\theta^*$), yielding a worst-case of $|\mc{D}| + 1$ model copies on disk.
Averaging the per-stage peaks across all $|\mc{D}|$ stages gives:
\[
  \frac{1}{|\mc{D}|}\sum_{s=1}^{|\mc{D}|}\bigl(\min(|\mc{D}| - s + 1,\; E) + 2\bigr),
\]
where $E$ is the number of evaluation steps per stage and $+2$ accounts for the retained $\hat{\theta}$ and $\theta^*$ (for $s \geq 2$; stage~1 retains none, but the over-count vanishes as $|\mc{D}|$ grows).
When $E \geq |\mc{D}|$, i.e.\ the evaluation grid is finer than the number of sub-datasets, the $\min$ reduces to $|\mc{D}| - s + 1$ and the average simplifies to $\frac{|\mc{D}| + 5}{2}$.
For our experiments with $|\mc{D}| = 10$ and $C = 3$ epochs evaluated every $0.25$ epochs ($E = 12 > |\mc{D}|$), this predicts a peak of $11$ and an average of $7.5$ model copies.
In practice, multiple categories often share the same peak epoch, so several per-dataset champions collapse onto a single checkpoint. Empirically, across \tsc{mSFT} runs with compute budgets $C \in \{1, 3\}$ on multiple dataset mixtures, we observe an average disk utilization of $4.44|\theta|$, well below the $|\mc{D}|+1$ theoretical bound. (see Appendix~\ref{app:storage} Figs.~\ref{fig:app_disk_1}, ~\ref{fig:app_disk_2}, ~\ref{fig:app_disk_3})

\clearpage
\section{Disk Storage Footprint}\label{app:storage}

\begin{center}
\begin{tabular}{cc}
    \includegraphics[width=0.48\textwidth]{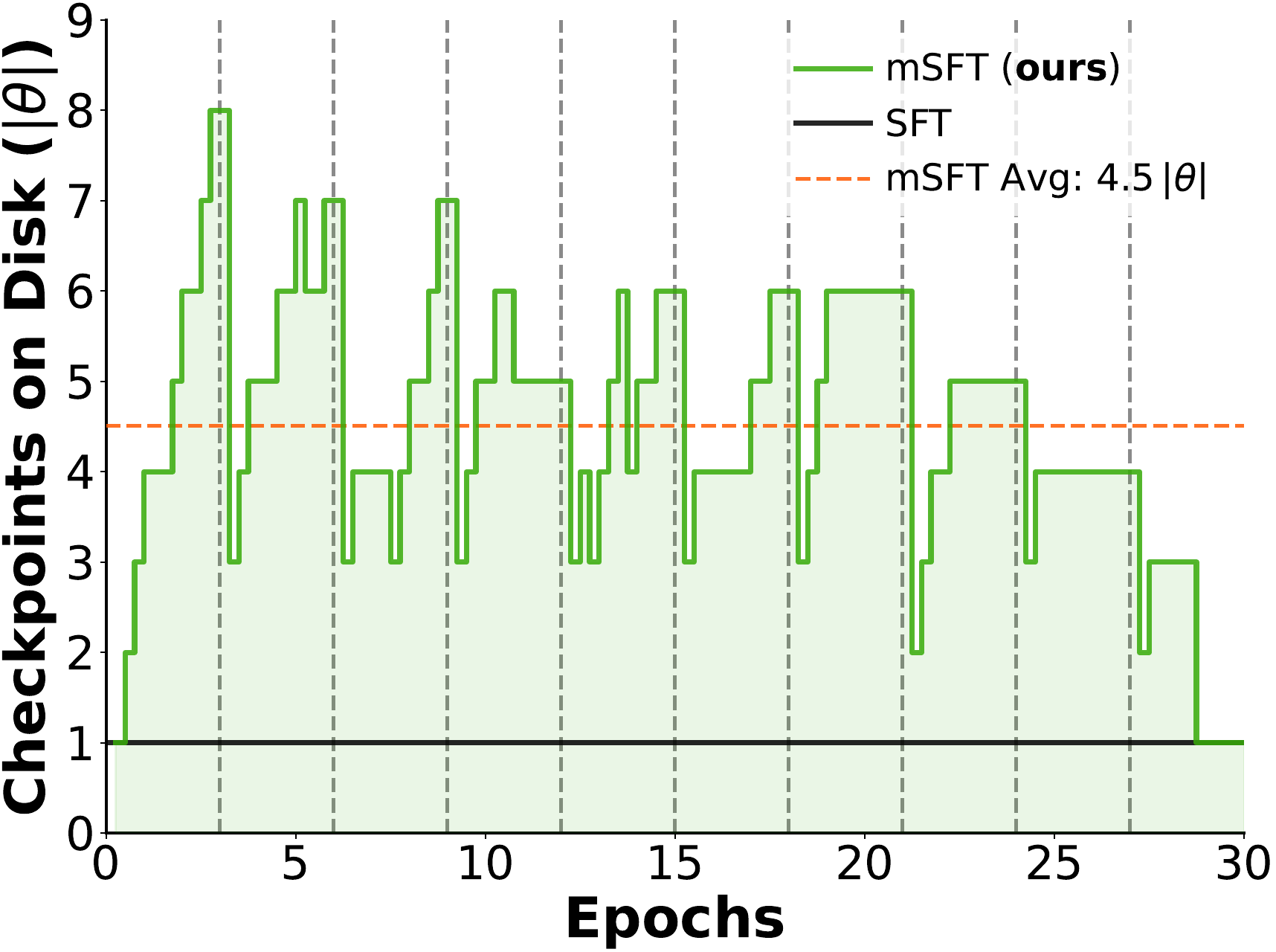} &
    \includegraphics[width=0.48\textwidth]{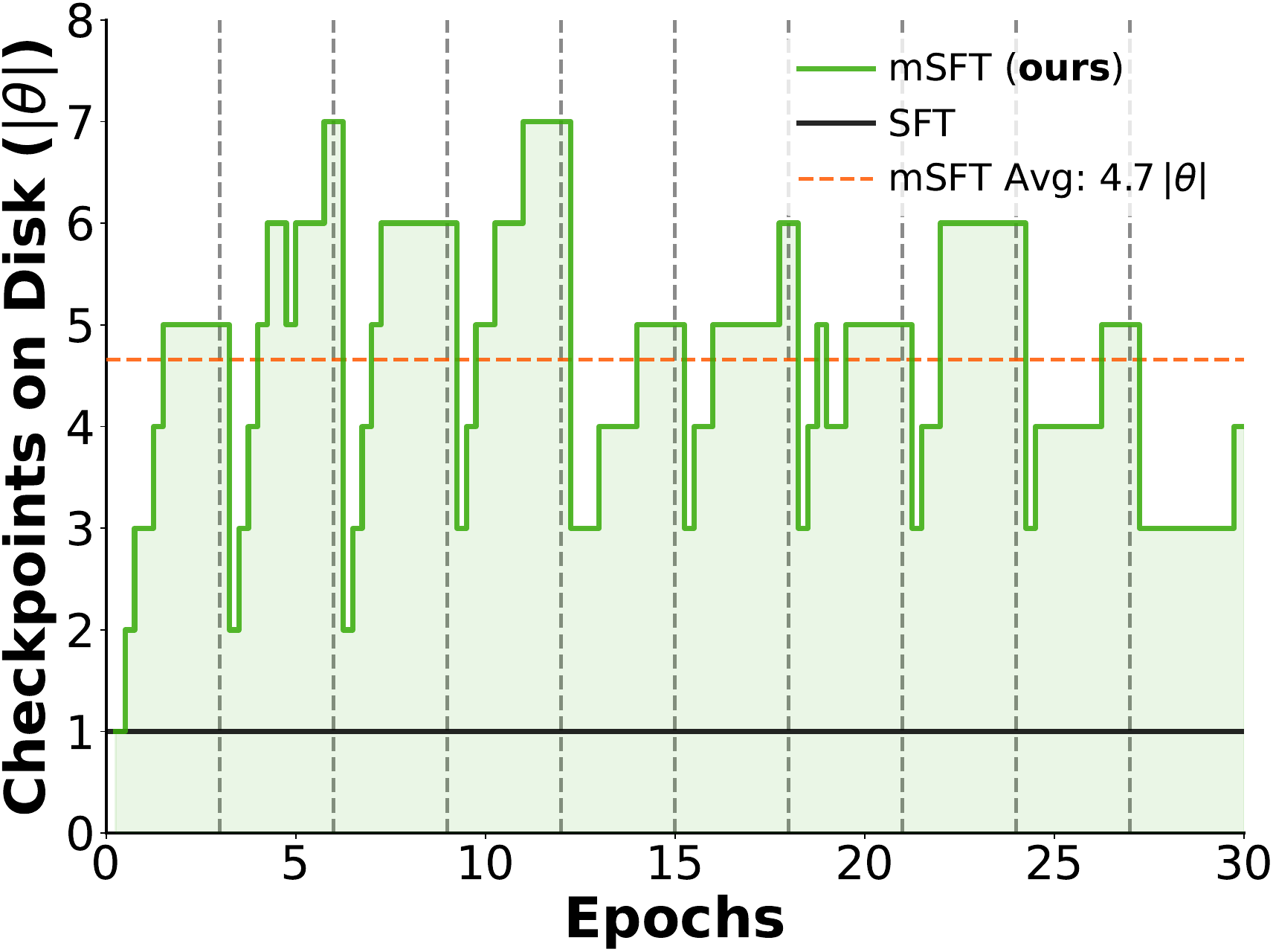} \\
    {\small (a) \ttt{Olmo 2 1B}, $C=3,\;N=10$} &
    {\small (b) \ttt{Qwen2.5 0.5B}, $C=3,\;N=10$} \\[1em]
    \includegraphics[width=0.48\textwidth]{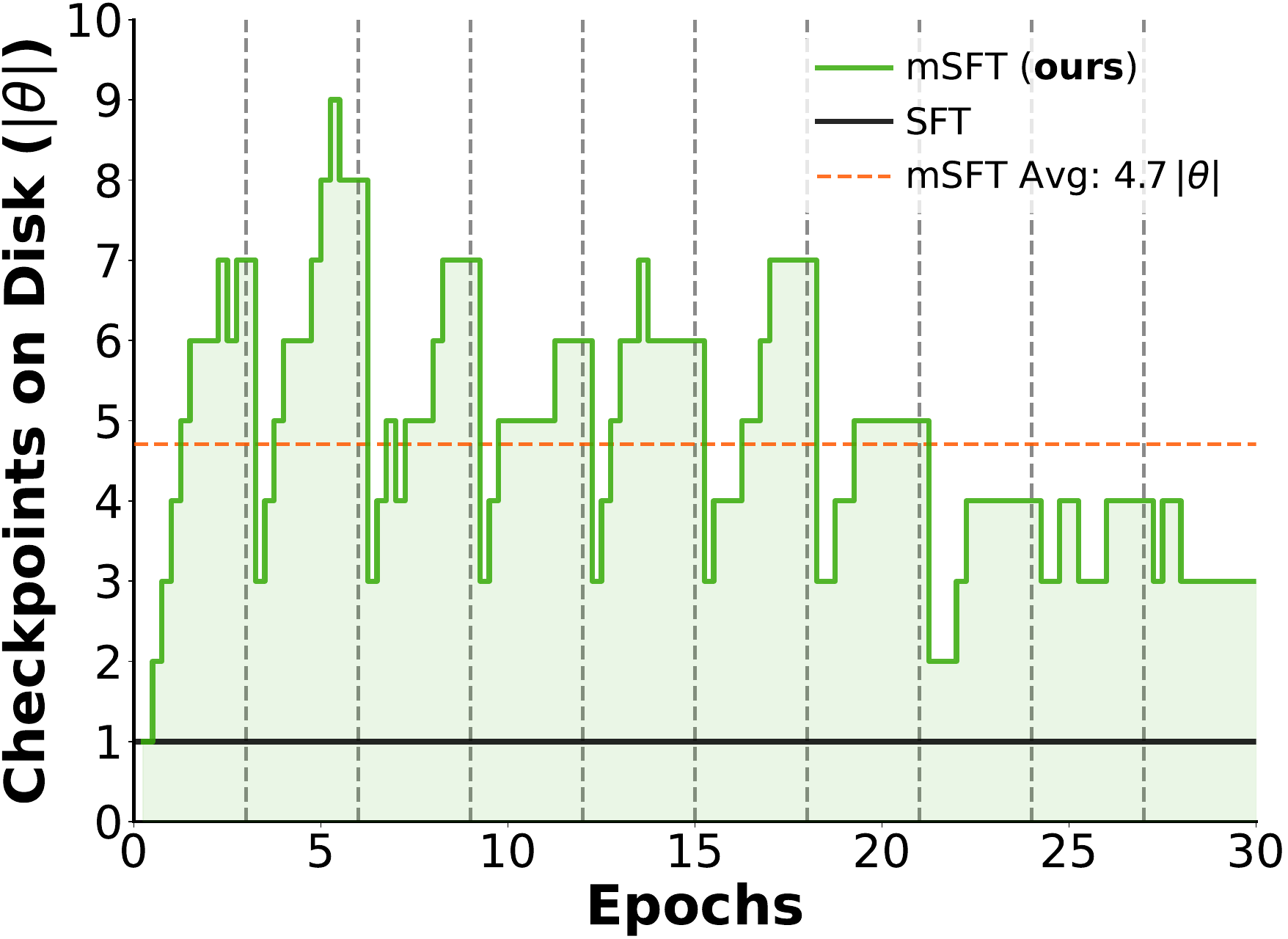} &
    \includegraphics[width=0.48\textwidth]{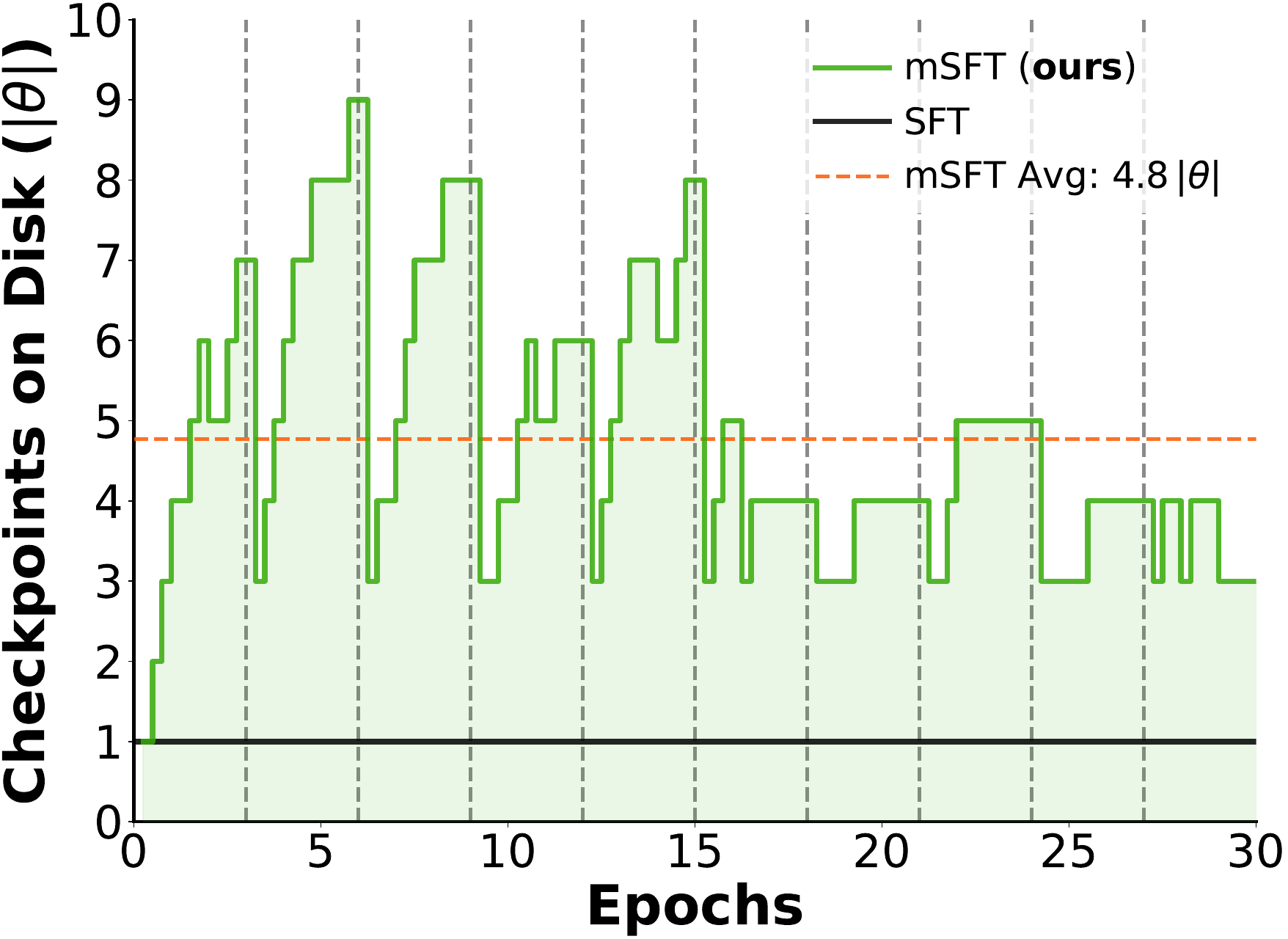} \\
    {\small (c) \ttt{Qwen2.5 1.5B}, $C=3,\;N=10$} &
    {\small (d) \ttt{Qwen2.5 3B}, $C=3,\;N=10$} \\[1em]
    \includegraphics[width=0.48\textwidth]{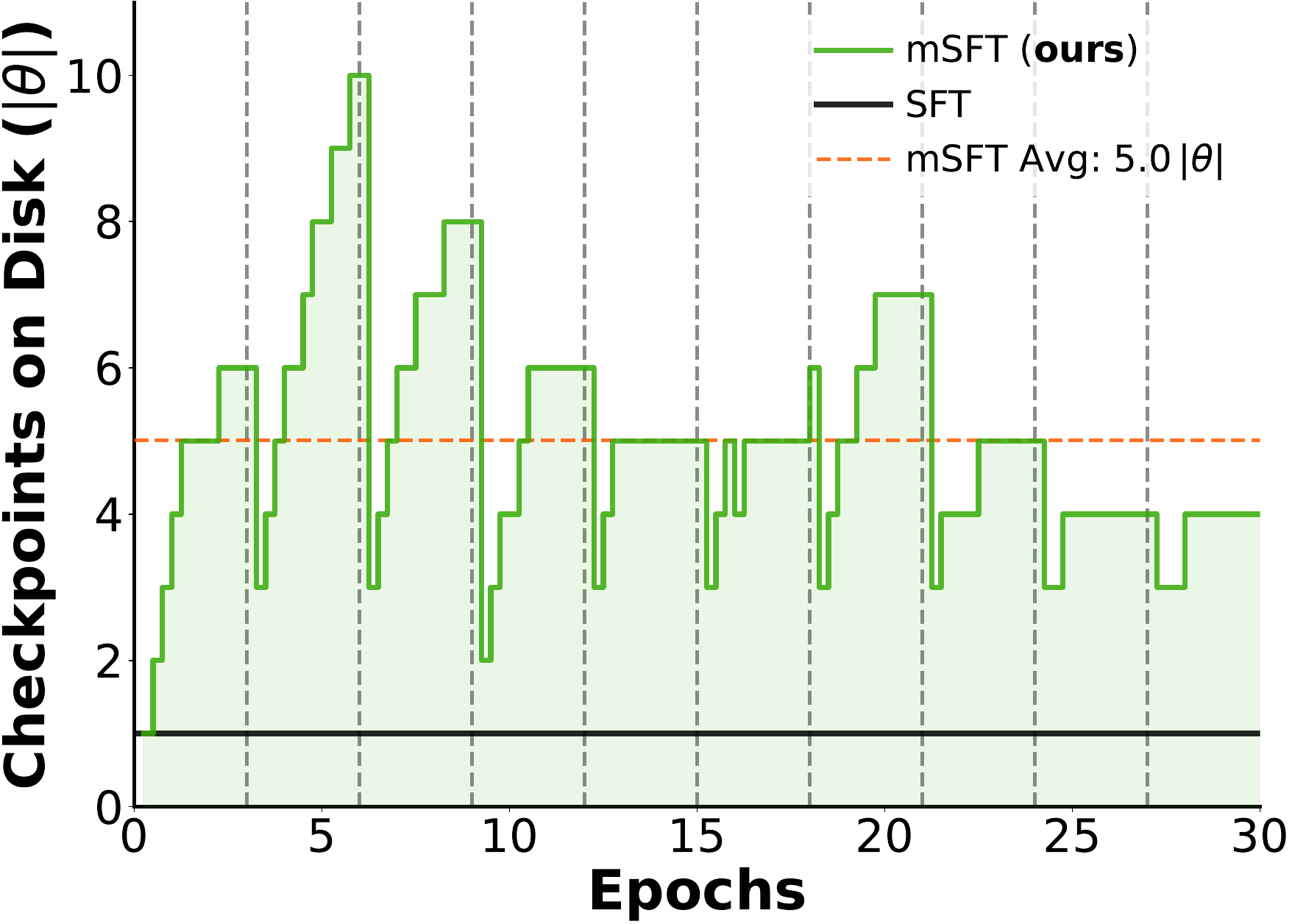} &
    \includegraphics[width=0.48\textwidth]{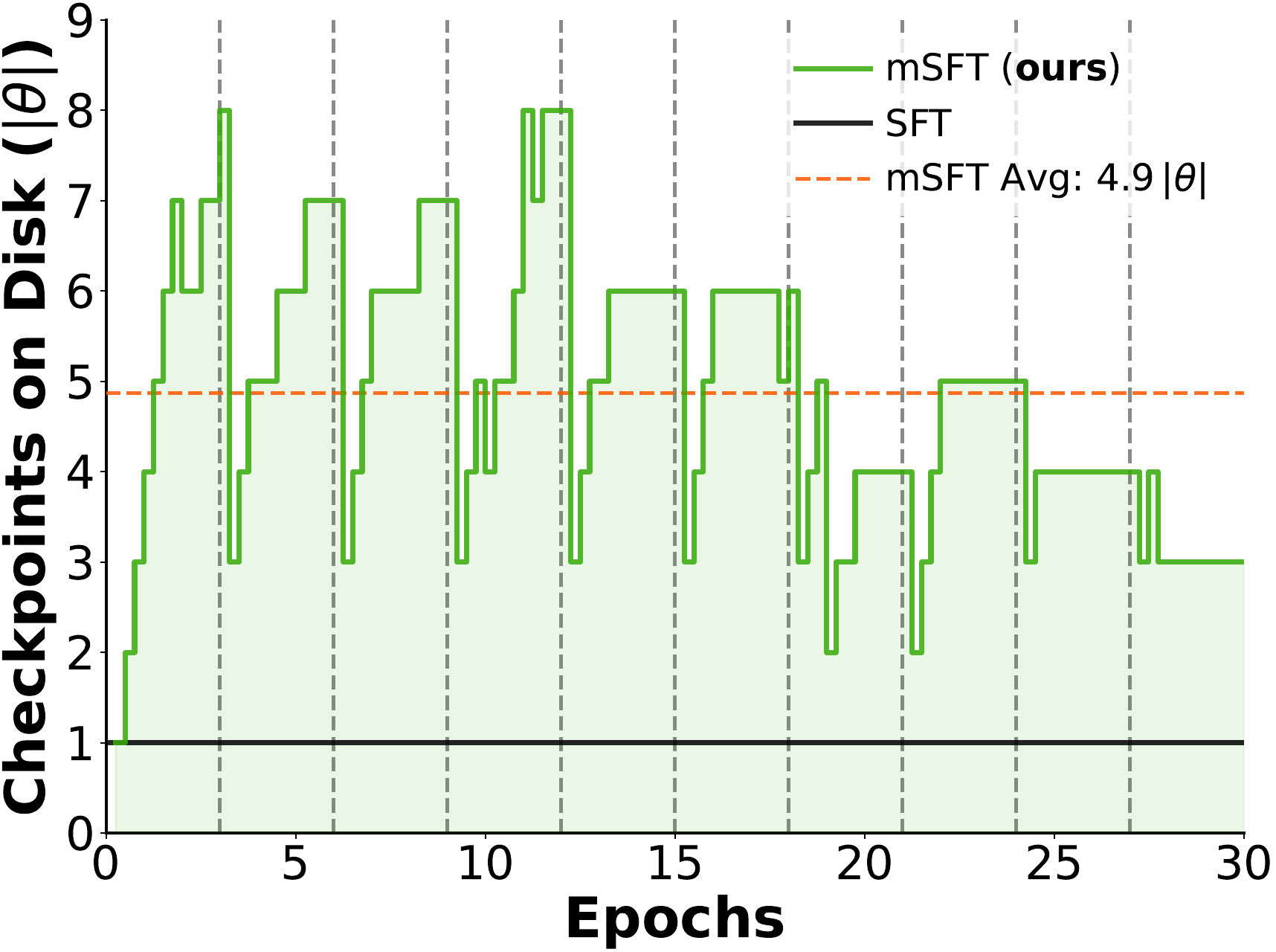} \\
    {\small (e) \ttt{Qwen2.5 7B}, $C=3,\;N=10$} &
    {\small (f) \ttt{Qwen3 8B}, $C=3,\;N=10$} \\
\end{tabular}
\begin{minipage}{\textwidth}
\captionof{figure}{\textbf{Disk utilization across \tsc{mSFT} iteration.} Each point denotes the number of checkpoints on disk at a given evaluation step, measured in multiples of model size $|\theta|$. Dashed vertical lines mark new roll-outs. The orange horizontal line indicates the average utilization across all evaluation steps.}
\label{fig:app_disk_1}
\end{minipage}
\end{center}

\clearpage

\begin{center}
\begin{tabular}{cc}
    \includegraphics[width=0.48\textwidth]{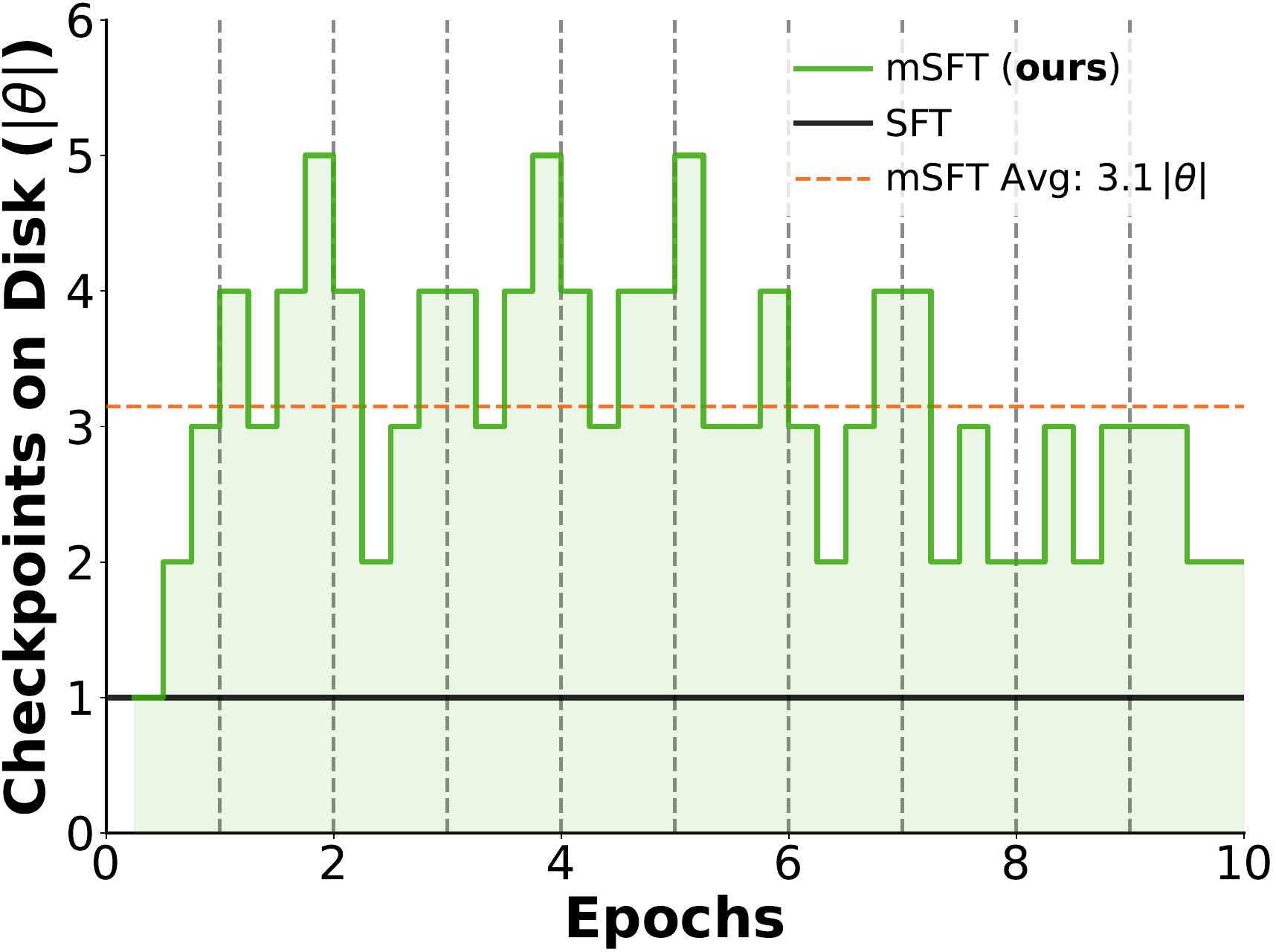} &
    \includegraphics[width=0.48\textwidth]{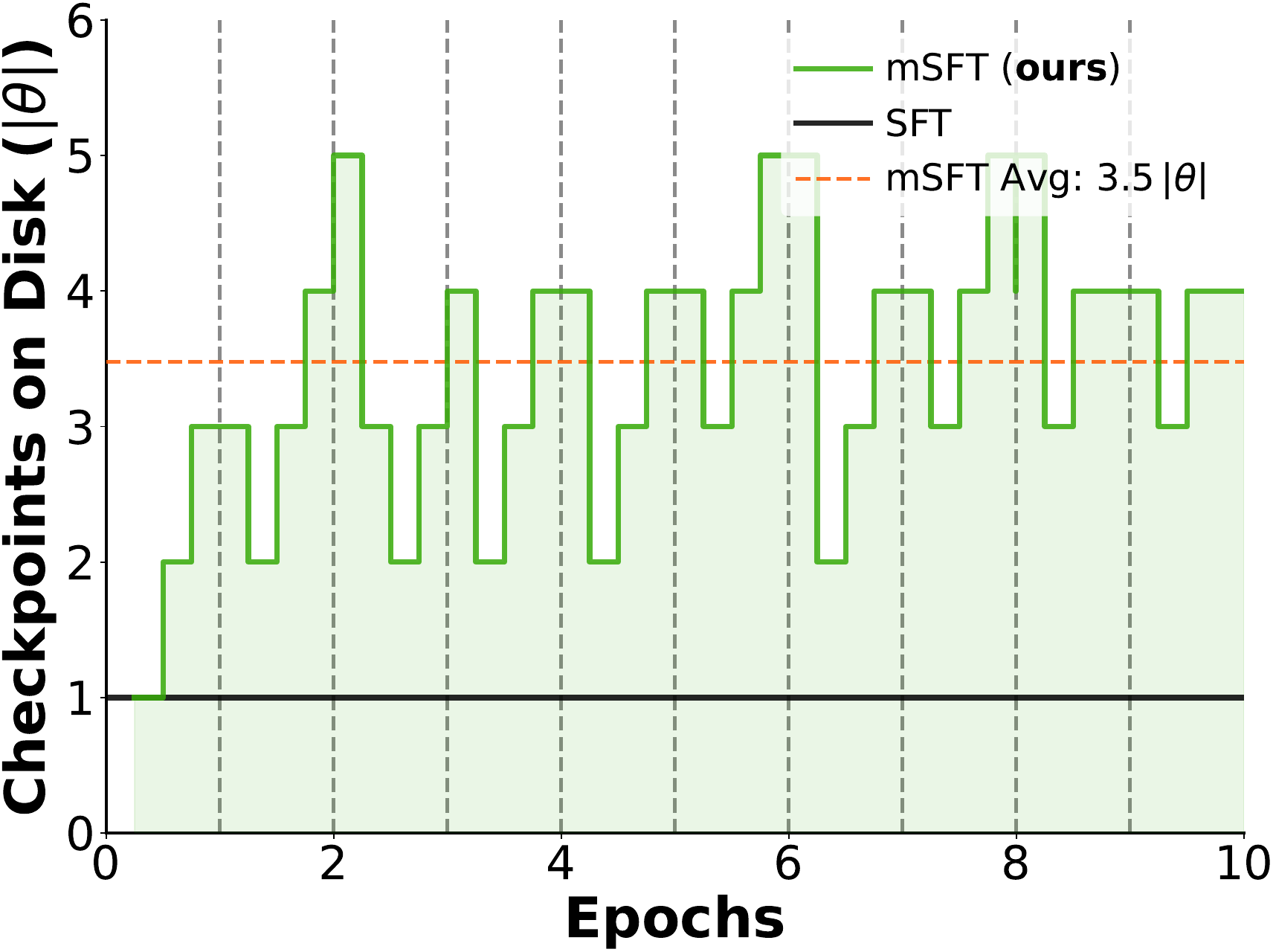} \\
    {\small (a) \ttt{Olmo 2 1B}, $C=1,\;N=10$} &
    {\small (b) \ttt{Qwen2.5 0.5B}, $C=1,\;N=10$} \\[1em]
    \includegraphics[width=0.48\textwidth]{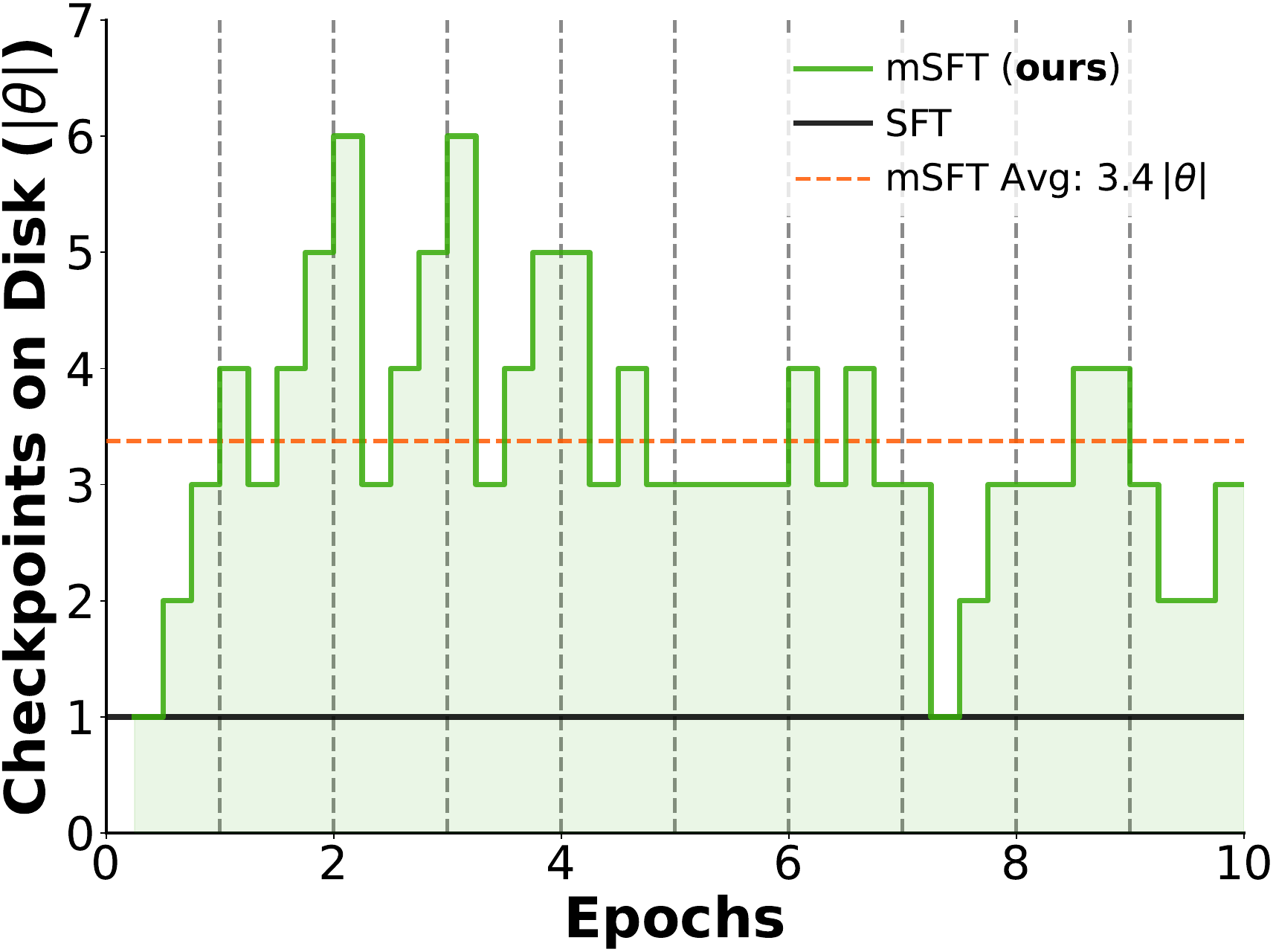} &
    \includegraphics[width=0.48\textwidth]{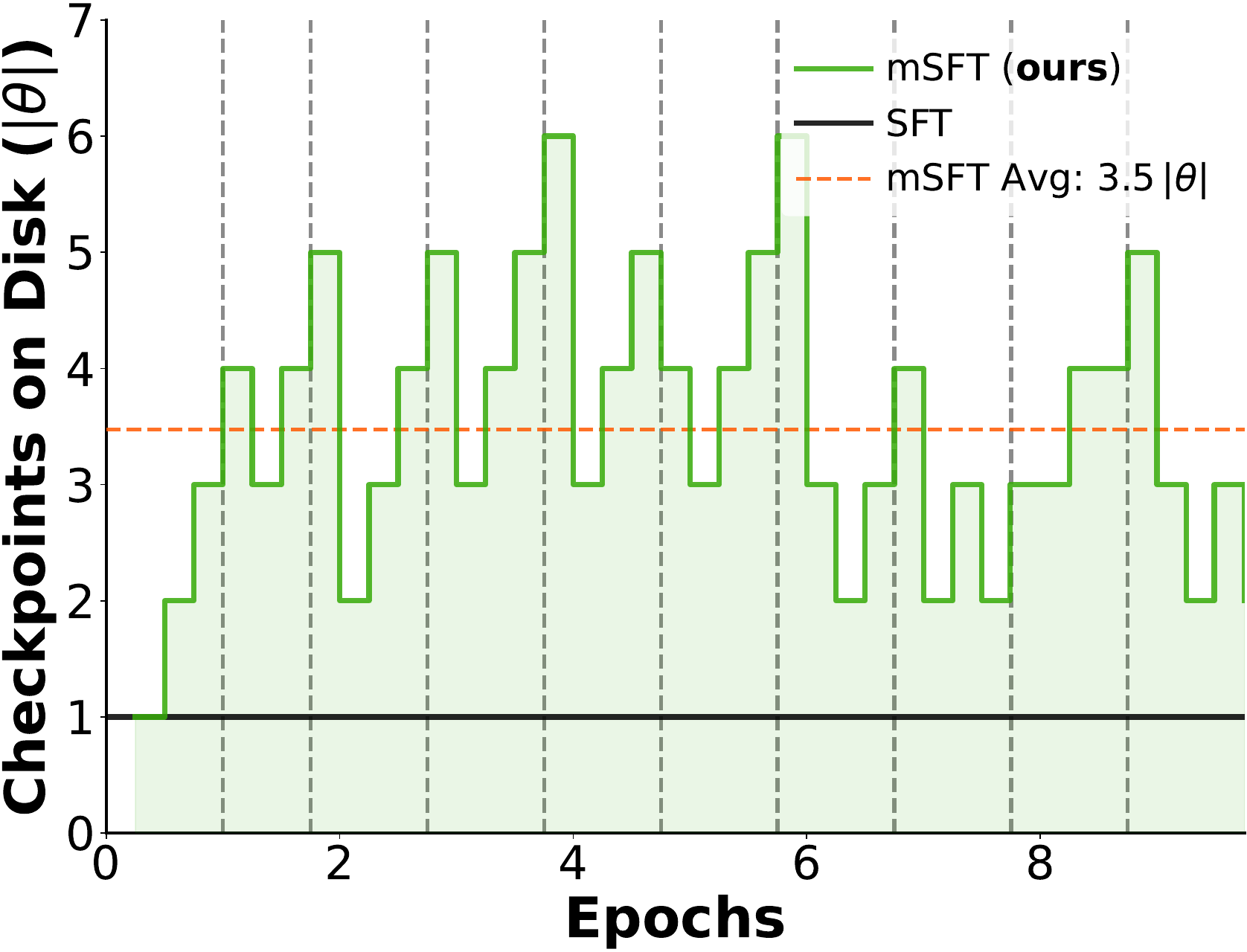} \\
    {\small (c) \ttt{Qwen2.5 1.5B}, $C=1,\;N=10$} &
    {\small (d) \ttt{Qwen2.5 3B}, $C=1,\;N=10$} \\
\end{tabular}
\begin{minipage}{\textwidth}
\captionof{figure}{\textbf{Disk utilization across \tsc{mSFT} iteration.} Each point denotes the number of checkpoints on disk at a given evaluation step, measured in multiples of model size $|\theta|$. Dashed vertical lines mark new roll-outs. The orange horizontal line indicates the average utilization across all evaluation steps.}
\label{fig:app_disk_2}
\end{minipage}
\end{center}

\clearpage

\begin{center}
\begin{tabular}{cc}
    \includegraphics[width=0.48\textwidth]{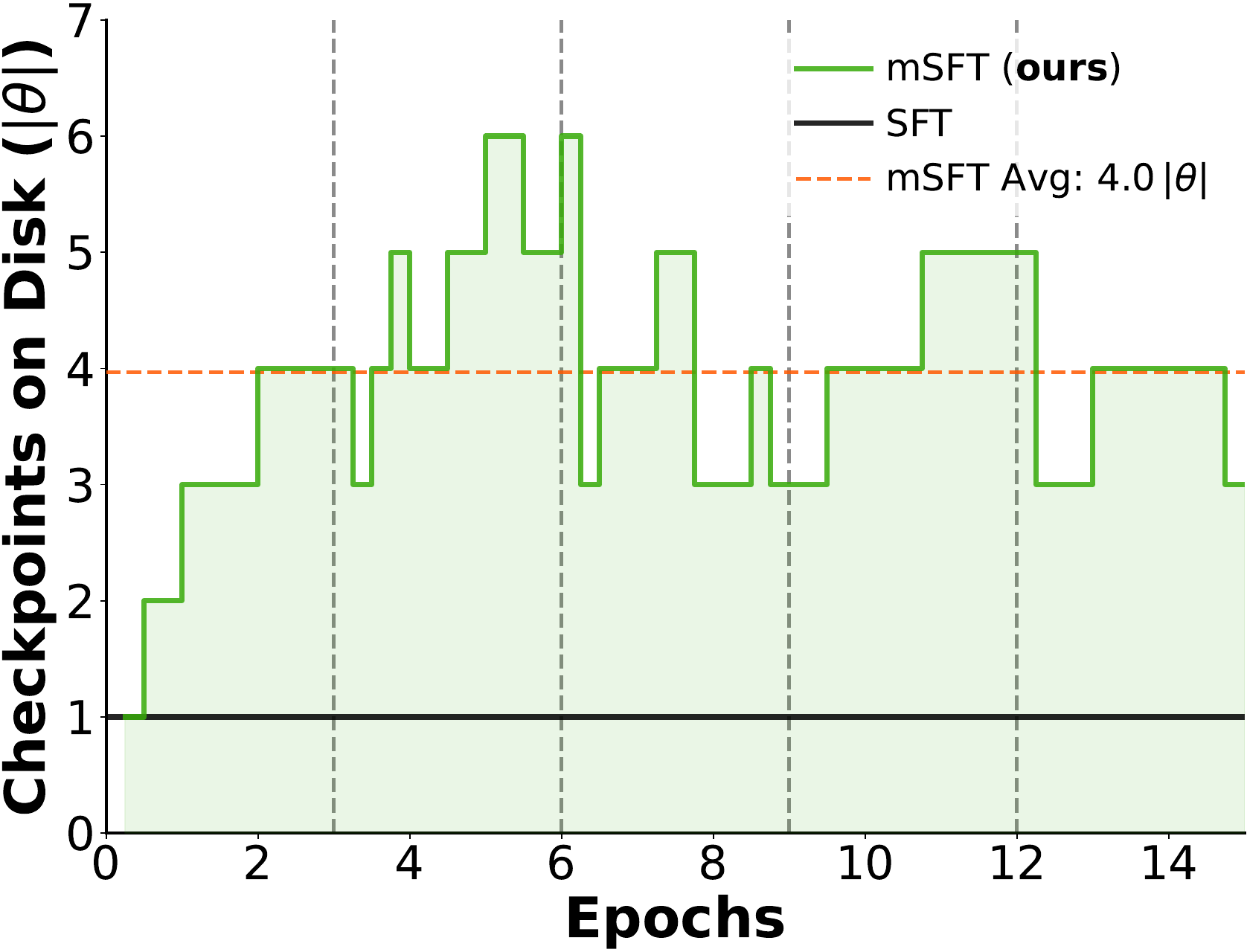} &
    \includegraphics[width=0.48\textwidth]{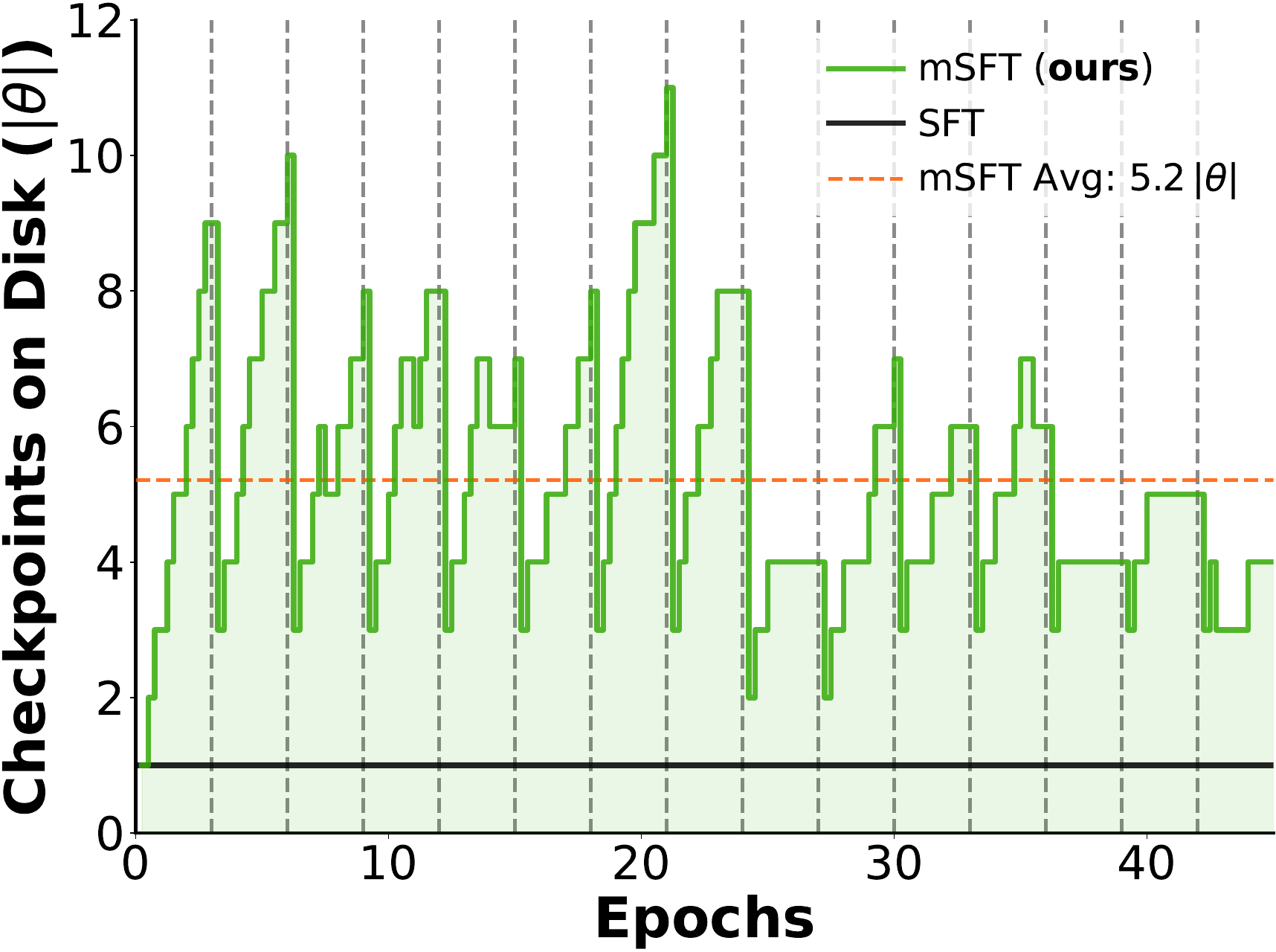} \\
    {\small (a) \ttt{Qwen2.5 3B}, $C=3,\;N=5$} &
    {\small (b) \ttt{Qwen2.5 3B}, $C=3,\;N=15$} \\[1em]
    \includegraphics[width=0.48\textwidth]{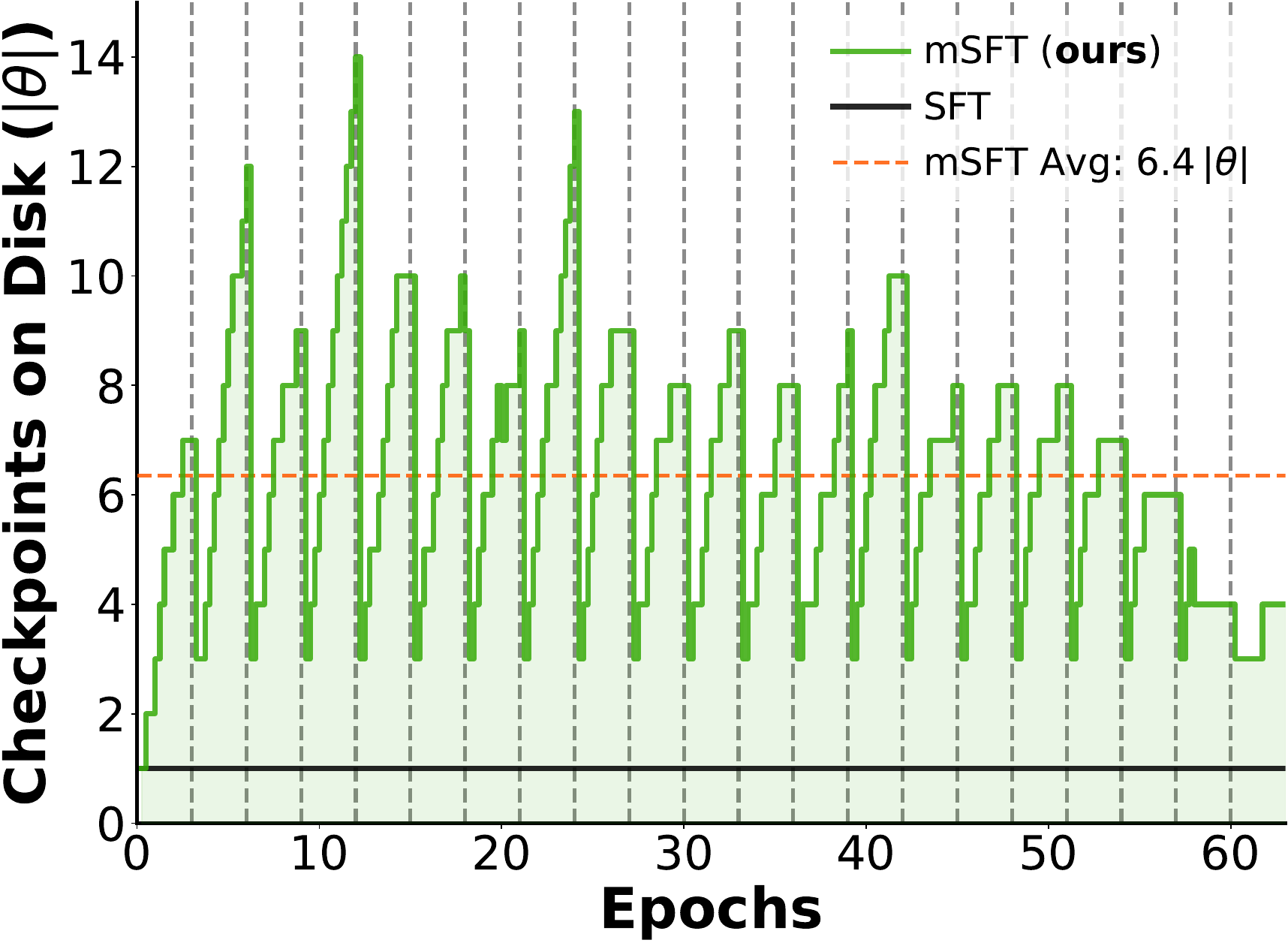} & \\
    {\small (c) \ttt{Qwen2.5 3B}, $C=3,\;N=21$} & \\
\end{tabular}
\begin{minipage}{\textwidth}
\captionof{figure}{\textbf{Disk utilization across \tsc{mSFT} iteration.} Each point denotes the number of checkpoints on disk at a given evaluation step, measured in multiples of model size $|\theta|$. Dashed vertical lines mark new roll-outs. The orange horizontal line indicates the average utilization across all evaluation steps.}
\label{fig:app_disk_3}
\end{minipage}
\end{center}

\section{LLM Usage}
As per ACL ARR policy, we report using LLMs for minor language polishing.

\end{document}